\documentclass{article}

\usepackage[preprint]{neurips_2026}

\usepackage[utf8]{inputenc}
\usepackage[T1]{fontenc}
\usepackage{hyperref}
\usepackage{url}
\usepackage{booktabs}
\usepackage{amsfonts}
\usepackage{amsmath}
\usepackage{amssymb}
\usepackage{amsthm}
\usepackage{mathtools}
\usepackage{nicefrac}
\usepackage{microtype}
\usepackage[table,dvipsnames]{xcolor}
\usepackage{graphicx}
\usepackage{subcaption}
\usepackage{multirow}
\usepackage{tabularx}
\usepackage{makecell}
\usepackage{adjustbox}
\usepackage{stackengine}
\usepackage{float}
\usepackage[capitalize,noabbrev]{cleveref}

\definecolor{brickred}{rgb}{0.8, 0.25, 0.33}
\definecolor{rowhighlight}{gray}{0.96}
\definecolor{headerblue}{gray}{0.2}
\definecolor{ourmethod}{RGB}{255, 215, 0}
\definecolor{baseline}{RGB}{220, 220, 220}
\definecolor{excellent}{RGB}{144, 238, 144}
\definecolor{poor}{RGB}{255, 182, 193}
\definecolor{gold}{RGB}{255,215,0}

\newcommand{\res}[2]{\textbf{#1} \tiny\textcolor{brickred}{$\pm$#2}}
\newcommand{\std}[2]{#1 \tiny\textcolor{brickred}{$\pm$#2}}
\newcommand{\ourres}[2]{\textbf{#1}\,{\scriptsize\textcolor{gray}{$\pm$#2}}}
\newcommand{\sota}[2]{\textbf{#1} \tiny\textcolor{brickred}{$\pm$#2}}

\newcommand{\R}{\mathbb{R}}

\theoremstyle{plain}

\theoremstyle{definition}

\theoremstyle{remark}

\title{Parameter-Efficient Fine-Tuning with Learnable Rank}

\author{%
  Arpit Garg \\
  Australian Institute for Machine Learning\\
  Adelaide University\\
  \texttt{arpit.garg@adelaide.edu.au} \\
  \And
  Simon Lucey \\
  Australian Institute for Machine Learning\\
  Adelaide University\\
  \texttt{simon.lucey@adelaide.edu.au} \\
  \And
  Hemanth Saratchandran \\
  Australian Institute for Machine Learning\\
  Adelaide University\\
  \texttt{hemanth.saratchandran@adelaide.edu.au} \\
}

\begin{document}

\maketitle

\begin{abstract}
Low-Rank Adaptation (LoRA) is a popular parameter-efficient fine-tuning (PEFT) method that restricts weight updates to low-rank adapters, introducing a fixed low-rank inductive bias by optimizing in a low-dimensional subspace. In this work, we question whether a fixed-rank constraint is the most effective inductive bias for parameter-efficient fine-tuning. We introduce \emph{Learnable Rank LoRA (LR-LoRA)}, a PEFT method in which the adapter rank is learned during the training process. Instead of prescribing a uniform rank for all adapter layers, LR-LoRA allows the optimizer to determine the appropriate rank for each layer. Using this approach, we find substantial layer-wise variation in the learned ranks, with the attention and MLP layers in the transformer models exhibiting systematically different rank preferences. Across a range of language understanding and commonsense reasoning benchmarks, LR-LoRA achieves state-of-the-art performance in most settings and consistently outperforms strong PEFT baselines, demonstrating that a learnable rank provides a more flexible and effective inductive bias than fixed-rank adaptations.
\end{abstract}

\section{Introduction}
\label{sec:intro}

Large pretrained models are increasingly being adapted to a wide range of downstream tasks, domains, and users. In this regime, the dominant computational bottleneck has shifted from pretraining to \emph{adaptation}, which is the process of specializing large backbone models under stringent memory, storage, and latency constraints \citep{touvron2023llama,grattafiori2024llama3,yang2024qwen2,abdin2024phi3}. Parameter-efficient fine-tuning (PEFT) addresses this challenge by keeping the backbone fixed while learning a small number of task-specific parameters, enabling rapid customization with orders of magnitude fewer trainable weights than full fine-tuning \citep{hu2022lora,han2024peftsurvey}.

Among PEFT methods, Low-Rank Adaptation (LoRA) and its variants have emerged as the de facto standard owing to their simplicity, stability, and ease of deployment \citep{hu2022lora}. LoRA augments selected layers with a low-rank update,
\[
\Delta \mathbf{W} = \mathbf{BA},
\]
parameterized by two low-rank adapter matrices, $\mathbf{A}$ and $\mathbf{B}$. This construction imposes a clear inductive bias: task-specific updates are constrained to a low-dimensional rank-$r$ set, the dimensionality of which is determined by the chosen rank.

Despite its empirical success, LoRA relies on a strong and largely unchallenged assumption: the adapter rank is fixed. In practice, a single global rank is typically applied uniformly across all layers, implicitly assuming that each layer requires a similar adaptation dimensionality. This assumption is difficult to justify, given what is known about transformer architectures, whose layers differ substantially across depth in function, representation geometry, and intrinsic dimensionality \citep{aghajanyan2021intrinsicdimension}.

Recent studies have attempted to mitigate this limitation by increasing the expressivity within the LoRA framework. For example, RandLoRA employs random bases to approximate full-rank updates while retaining a low-rank parameterization \citep{albert2025randlora}, and SineLoRA increases the rank of MLP adapters through fixed sinusoidal modulation \citep{ji2024sine}. Although these approaches introduce different inductive biases, they rely on ranks that are externally specified and fixed throughout the training. Consequently, they lack a mechanism that allows the model to \emph{learn} the appropriate adaptation dimensionality in a layer- or task-dependent manner.

Taken together, these observations raise a simple but fundamental question:
\begin{quote}
\emph{Is enforcing a fixed-rank constraint on adapter updates an optimal inductive bias for achieving strong performance in parameter-efficient fine-tuning?}
\end{quote}

To address this question, we introduce \textbf{Learnable Rank LoRA (LR-LoRA)}, a PEFT method in which the adapter rank is learned during the training process. Rather than prescribing a fixed rank for each layer throughout training, LR-LoRA augments the adapter update with an element-wise nonlinearity $\phi$, yielding
\[
\Delta \mathbf{W} = \phi(\mathbf{BA}),
\]
where $\phi$ is learned jointly with the model's parameters. By applying a nonlinearity to the low-rank product $\mathbf{BA}$, LR-LoRA allows the rank of the update to increase or decrease, relaxing the fixed-rank constraint imposed by standard low-rank adapters and enabling the optimizer to adapt the dimensionality of each layer's update (formalized in \cref{sec:learnable_nonlinearity}).

Across language understanding, commonsense reasoning, vision transfer, and instruction-tuning benchmarks, LR-LoRA consistently outperforms strong PEFT baselines under approximately matched parameter budgets, achieving state-of-the-art performance in most settings. Empirically, we observe substantial variation in the learned ranks across layers, with attention and MLP adapters exhibiting systematically different adaptation dimensionalities. These findings suggest that enforcing a hard, fixed-rank constraint is suboptimal for parameter-efficient fine-tuning methods, such as LoRA.

The main contributions of this study are as follows:
\begin{itemize}
\item \textbf{Learnable Rank LoRA (LR-LoRA):} a parameter-efficient fine-tuning method that learns layer-wise adapter ranks during training, introducing a more flexible inductive bias for adaptation.
\item \textbf{A basis-agnostic principle for adaptation dimensionality:} we show that applying any differentiable element-wise nonlinearity to $\mathbf{BA}$ removes the rank-$r$ constraint, and we provide direct empirical evidence that the gains stem from \emph{learning} per-layer adaptation, not from the mere presence of nonlinearity.
\item \textbf{Consistent improvements over strong PEFT baselines}, achieving new state-of-the-art results across $7$ architectures spanning $125$M to $13$B parameters, $19$ tasks, and four evaluation paradigms (commonsense reasoning, GLUE transfer, vision transfer, and instruction tuning).
\end{itemize}

More broadly, this study reframes Low-Rank Adaptation through the lens of inductive bias. Rather than treating rank as a fixed hyperparameter, we argue that adaptation dimensionality should be a learnable property of the model that reflects the heterogeneity of representations across layers and tasks. This perspective suggests a shift from rigid parameter-efficient designs to adaptive mechanisms that allow models to allocate capacity where it is most needed.

\section{Related Work}\label{sec:related}

\textbf{Parameter-efficient fine-tuning.}
Parameter-efficient fine-tuning (PEFT) adapts large pretrained models by learning a small set of task-specific parameters while keeping the backbone frozen, enabling efficient storage and deployment across many tasks \citep{houlsby2019adapter,ding2023parameter,han2024peftsurvey}. Major PEFT families include adapter modules \citep{houlsby2019adapter}, prompt and prefix-based tuning \citep{lester2021prompt,li2021prefix}, multiplicative adaptation methods such as IA$^3$ \citep{liu2022few}, and lightweight weight-space updates such as BitFit \citep{zaken2021bitfit}. Our work lies within the weight-space PEFT and focuses specifically on controlling the dimensionality of low-rank weight updates.

\textbf{Low-rank adaptation (LoRA).}
LoRA utilizes a low-rank update $\Delta \mathbf{W} = \mathbf{BA}$ in selected linear layers, imposing the inductive bias that task-specific weight changes reside in a low-dimensional rank-$r$ set \citep{hu2022lora,fu2023effectiveness}. A large body of follow-up work improves LoRA through rank allocation heuristics \citep{zhang2023adalora,valipour2023dylora, albert2025towards}, alternative parameterizations \citep{liu2024dora}, randomized or vectorized adaptation \citep{kopiczko2024vera}, improved initialization \citep{meng2024pissa}, training refinements \citep{hayou2024loraplus}, memory-efficient training \citep{dettmers2023qlora,zhao2024galore,zhang2023lorafa}, and stability analyses \citep{kalajdzievski2023rslora,lialin2023relora}. Although these methods improve \emph{where} or \emph{how much} low-rank capacity is allocated, they generally treat rank as an externally specified hyperparameter. In contrast, we relax this assumption by allowing the adapter rank to be learned.
Furthermore, several approaches increase the expressivity of adapters while preserving compact parameterization. RandLoRA employs random bases to approximate more expressive (including near-full-rank) updates within a low-rank framework \citep{albert2025randlora}, and VeRA uses a randomized vector-based adaptation \citep{kopiczko2024vera}. Another line of work introduces a nonlinearity into the adapter pathway: SineLoRA applies a fixed sinusoidal modulation to adapter weights, obtaining a high-rank adapter update and increasing representational capacity without increasing the parameter count \citep{ji2024sine}. Recent adaptive-rank methods such as AdaLoRA \citep{zhang2023adalora}, DyLoRA \citep{valipour2023dylora}, and ElaLoRA \citep{chang2025elalora} reallocate per-layer ranks during training under a fixed total budget set ahead of time; nonlinear variants like AuroRA \citep{dong2025aurora} and ABBA \citep{singhal2025abba} break linear constraints through adaptive nonlinear layers and Hadamard products. In all of these, the global capacity is externally specified. LR-LoRA differs in \emph{how} the effective rank arises: the optimizer learns a per-layer continuous elementwise nonlinearity over $\mathbf{BA}$ that implicitly sets each layer's effective (stable) rank without an external allocation budget, and we provide controlled experiments isolating learnable rank from the mere presence of nonlinearity (\cref{subsec:ablations}).

\section{Methodology}\label{sec:method}

In this section, we describe the methodology underlying Learnable Rank LoRA (LR-LoRA), with a particular focus on the construction and learning of learnable rank adapter updates.

\subsection{Low-Rank Adaptation}\label{sec:low_rank_adapt}

Low-Rank Adaptation (LoRA) is a parameter-efficient fine-tuning method that keeps the pretrained weights $\mathbf{W} \in \mathbb{R}^{m \times n}$ of a network frozen and introduces trainable adapter matrices $\mathbf{A} \in \mathbb{R}^{r \times n}$ and $\mathbf{B} \in \mathbb{R}^{m \times r}$, where $r \ll \min(m,n)$. The adapted weight matrix is given by
\begin{equation}\label{eqn:lora_eqn}
\mathbf{W} + \mathbf{BA}.
\end{equation}
During task-specific training, only the adapter parameters $\mathbf{A}$ and $\mathbf{B}$ are updated, while $\mathbf{W}$ remains fixed.

The efficiency of LoRA arises from the fact that $\mathbf{W}$ contains $mn$ parameters, whereas the adapter update $\mathbf{BA}$ involves only $r(m+n)$ parameters, which is substantially smaller when $r \ll \min(m,n)$. This parameter reduction comes at the cost of expressivity: since $\mathrm{rank}(\mathbf{BA}) \le r$, LoRA constrains task-specific updates to the low-dimensional rank-$r$ set. Consequently, the LoRA architecture imposes a low-rank inductive bias on the adaptation process.

\subsection{Preliminaries on Sampling Theory}\label{sec:sampling}

We adopt the normalized sinc convention $\mathrm{sinc}(x)=\sin(\pi x)/(\pi x)$ for $x\!\neq\!0$ and $\mathrm{sinc}(0)=1$, so that the Whittaker-Shannon interpolation formula holds in its standard form. The Nyquist-Shannon sampling theorem \citep{nyquist1928certain,shannon1949communication,martin1997introduction} states that any bandlimited $f\in L^2(\R)$ with maximum frequency $\omega_{\max}$ can be exactly reconstructed from samples $\{f(x_i)\}$ on a uniform grid with spacing $\tau\le 1/(2\omega_{\max})$ as
\begin{equation}\label{eqn:original_shannon_recon}
f(x) = \sum_{i\in\mathbb{Z}} f(x_i)\,
\mathrm{sinc}\!\left( \omega (x - x_i) \right), \qquad \omega := \tfrac{1}{\tau},\quad x_i := i\tau.
\end{equation}
In practice the sum is truncated to $N$ terms (with bounded approximation error for sufficiently large $N$), and more generally functions in $L^2(\R)$ are well approximated by mixtures of shifted sincs with learnable amplitude $\alpha_i$ and bandwidth $\omega_i$:
\begin{equation}\label{eqn:shifted_sinc}
\sum_{i=1}^{N} \alpha_i \,
\mathrm{sinc}\!\left( \omega_i (x - x_i) \right).
\end{equation}
This expansion forms the basis for the learnable nonlinearity introduced next. A more complete review with derivations is provided in \cref{app:sec:sampling}.

\subsection{Learnable Adapter Nonlinearity}\label{sec:learnable_nonlinearity}

\begin{figure}[t]
    \centering
    \includegraphics[width=0.42\linewidth]{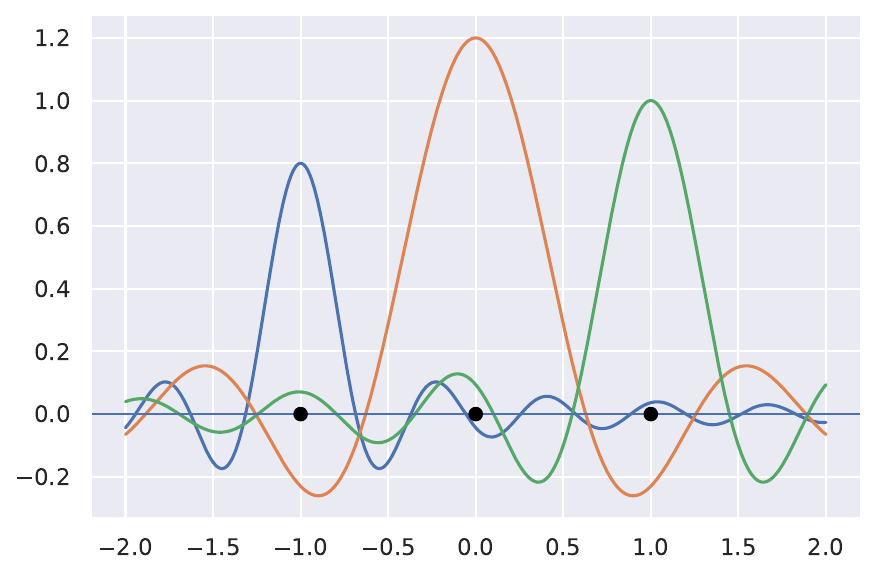}
    \hfill
    \includegraphics[width=0.42\linewidth]{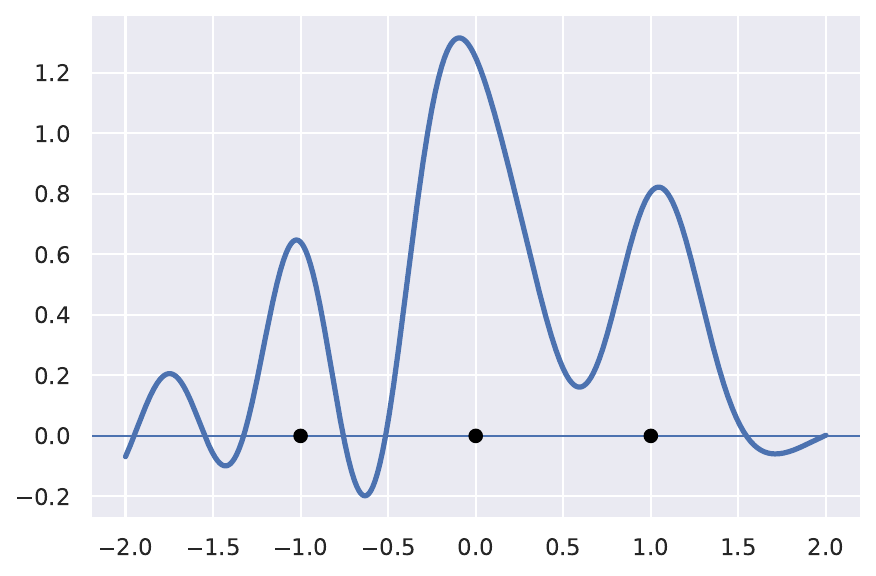}
    \caption{\textbf{Construction of the learnable nonlinearity $\phi$ (\cref{eqn:learnable_nonlin_eqn}).} \textbf{Left:} individual $\mathrm{sinc}$ basis functions on the grid $\{x_i\}$ with per-basis bandwidth $\omega_i$ and amplitude $\alpha_i$. \textbf{Right:} the resulting $\phi(x)$ obtained by summing the shifted $\mathrm{sinc}$ functions. Bandwidths and amplitudes are learned jointly with the adapter weights.}
    \label{fig:sinc_representation}
\end{figure}

The core idea of LR-LoRA is to apply a nonlinearity $\phi$ to the adapter update $\mathbf{BA}$, yielding $\phi(\mathbf{BA})$. We construct $\phi$ using a flexible basis of shifted $\mathrm{sinc}$ functions. Specifically, we consider a collection of $N$ uniformly spaced grid points $\{x_1, x_2, \ldots, x_N\}$ over a bounded interval $[-I, I]$. At each grid point $x_i$, we place a $\mathrm{sinc}$ function with bandwidth $\omega_i$ and amplitude $\alpha_i$, yielding basis functions of the form $\mathrm{sinc}\!\left(\omega_i (x - x_i)\right)$. The resulting nonlinearity is defined as
\begin{equation}\label{eqn:learnable_nonlin_eqn}
\phi(x) = \sum_{i=1}^N \alpha_i \, \mathrm{sinc}\!\left(\omega_i (x - x_i)\right).
\end{equation}
To make $\phi$ learnable, we treat the parameters $\{(\omega_1, \alpha_1), \ldots, (\omega_N, \alpha_N)\}$ as trainable and optimize them jointly with the adapter weights during fine-tuning. An illustration of this construction is shown in \cref{fig:sinc_representation}.

\paragraph{Why a nonlinearity removes the rank-$r$ constraint.}
The product $\mathbf{BA}$ defines a low-rank \emph{linear} map, constraining task-specific updates to the rank-$r$ set. Applying an element-wise nonlinearity $\phi$ removes this rank constraint by mapping the update through a nonlinear transformation. Consequently, the adapted update $\phi(\mathbf{BA})$ is no longer restricted to rank $r$ and can have rank less than or greater than $r$.

\paragraph{Implementation.}
Adapter matrices $\mathbf{A} \in \R^{r \times n}$ and $\mathbf{B} \in \R^{m \times r}$ are chosen as in LoRA. Each adapter has its own $\phi$, so every layer learns its own transfer-function parameters. We initialize $\alpha_i\!=\!0$ so that $\phi\!\equiv\!0$ at the start of training (any expressive capacity must be \emph{learned}), and parameterize $\omega_i\!=\!\mathrm{softplus}(\tilde\omega_i)$ with $\tilde\omega_i$ initialized to the constant $\ln(e^{\omega_0}-1)$ so that $\omega_i = \omega_0$ at initialization (specifically $\tilde\omega_i \approx 0.54$ for the default $\omega_0=1.0$), keeping $\omega_i$ strictly positive while avoiding the degenerate zero solution. The learned adaptation dimensionality of each layer can therefore be smaller or larger than the nominal $r$. The additional parameters scale with grid size and layer count and incur small overhead in practice (\cref{subsec:training_efficiency}); full details in \cref{app:sec:appendix_implementation,app:fig:pipeline}.

\paragraph{Stable rank as a measure of update complexity.}
In practice, computing the matrix rank is numerically ill-posed because it requires choosing a threshold on the singular values. Therefore, in \cref{sec:experiments} we quantify adaptation complexity using the \emph{stable rank}, as also done in \citet{ji2024sine}. For a matrix $\mathbf{M}$, the stable rank is defined as
\begin{equation}\label{eqn:stable_rank}
\mathcal{S}(\mathbf{M}) := \frac{\|\mathbf{M}\|_F^2}{\|\mathbf{M}\|_2^2},
\end{equation}
where $\|\cdot\|_F$ denotes the Frobenius norm and $\|\cdot\|_2$ denotes the largest singular value of the matrix. The stable rank provides a continuous and numerically robust measure of spectral complexity and is therefore well suited for empirical analysis. We use stable rank as a proxy for update complexity throughout the paper; in \cref{subsec:layer_insights} we additionally provide direct functional evidence (parameter-sharing and module-wise ablations) that does not rely on this proxy.

Our experimental results in \cref{sec:experiments} show that the stable rank of $\phi(\mathbf{BA})$ varies substantially across layers: in some layers, it is smaller than that of $\mathbf{BA}$, while in others, it is larger. This behavior indicates that different layers prefer different levels of expressive capacity, reinforcing the motivation to learn adaptation dimensionality in a layer-dependent manner. We analyze the effect of the grid size and number of grid points in \cref{sec:experiments}, studying the trade-offs between expressivity and stability.

\paragraph{Basis-agnostic principle.}
The central scientific claim of LR-LoRA is not the specific sinc parameterization, but the basis-agnostic principle that \emph{the appropriate adaptation dimensionality of each layer is unknown before training and should be discovered by the optimizer, not prescribed by the practitioner}. Applying any differentiable elementwise nonlinearity $\phi$ to $\mathbf{BA}$ removes the rank-$r$ constraint (\cref{eqn:lora_eqn}), allowing the realized stable rank $\mathcal{S}(\phi(\mathbf{BA}))$ to differ from the nominal $r$ (\cref{eqn:learnable_nonlin_eqn,eqn:stable_rank}). Sinc is one realization, chosen for the controlled-bandwidth, non-periodic inductive bias justified above.

\paragraph{Other basis functions.}
More generally, other choices of basis functions are possible, including trigonometric bases such as $\sin$ and $\cos$ and spline-based representations. However, according to classical Fourier theory \citep{stein2009real}, representations based on $\sin$ and $\cos$ inherently impose a periodic structure, restricting the resulting functions to be periodic. We avoid such bases because the induced periodic inductive bias may be undesirable when adapting model weights. Spline-based bases (as used in \cite{teney2025we}) provide an alternative that does not enforce periodicity. Among Gaussian RBF, B-splines, Fourier, polynomial, and sinc bases evaluated under matched parameter counts (\cref{app:sec:basis_comparison_detailed}), sinc achieves the highest accuracy and stability; B-splines (the simplest non-periodic alternative) score close behind, while the periodic Fourier basis underperforms, consistent with the periodicity argument above. Sinc is therefore a well-motivated choice within an open family rather than a unique requirement.

\begin{figure}[t]
\centering
\begin{subfigure}[t]{0.48\linewidth}
  \centering
  \includegraphics[width=\linewidth]{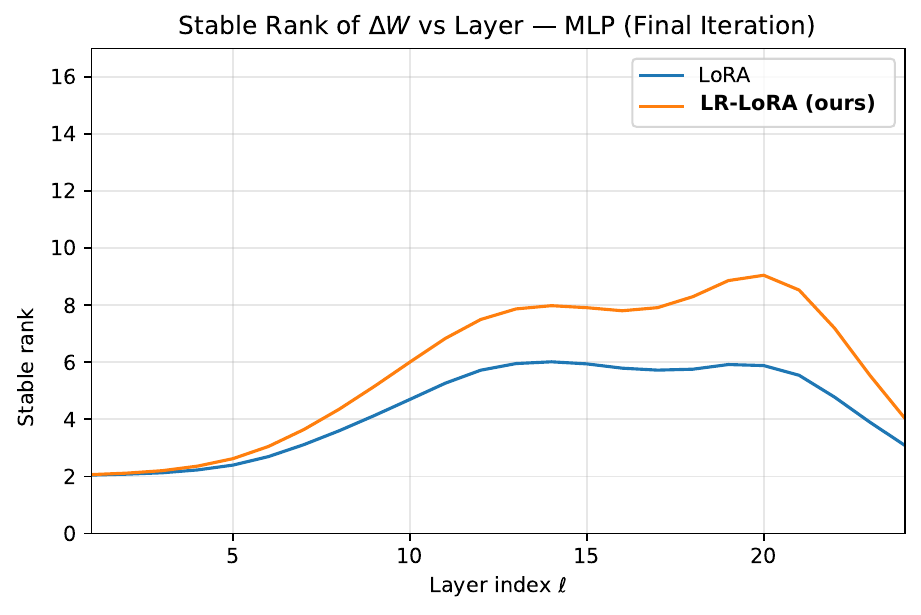}
  \caption{MLP projections (final checkpoint).}
\end{subfigure}
\hfill
\begin{subfigure}[t]{0.48\linewidth}
  \centering
  \includegraphics[width=\linewidth]{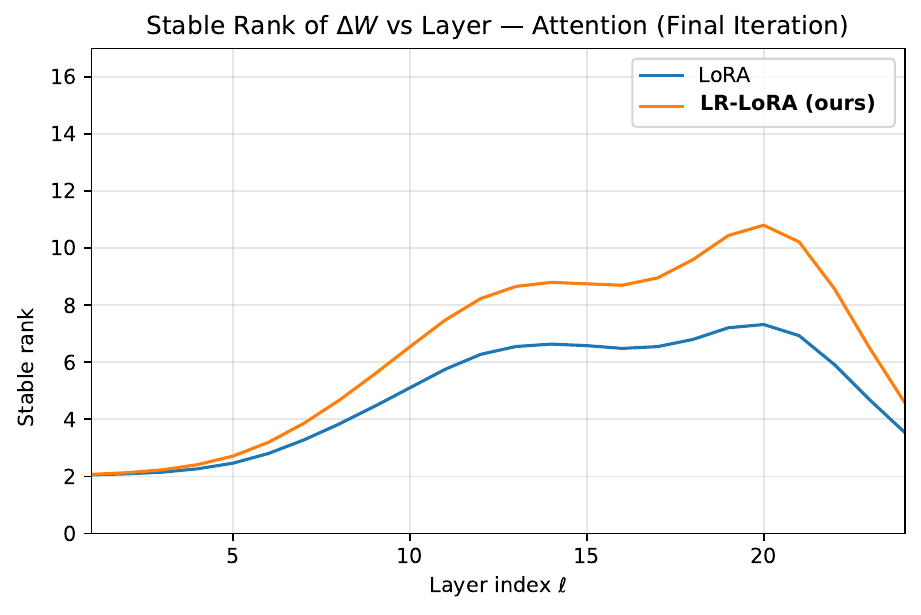}
  \caption{Attention projections (final checkpoint).}
\end{subfigure}
\caption{\textbf{Stable-rank profiles of learned updates across depth.}
\texttt{Qwen2-0.5B} on commonsense reasoning at rank $r{=}16$; each panel plots stable rank (y-axis) against layer index (x-axis) at the final training checkpoint. \textbf{Takeaway:} update complexity is depth-dependent and differs systematically between MLP and attention modules. Per-checkpoint evolution and the corresponding learned nonlinearities $\phi(z)$ at representative early, mid, and late layers are reported in \cref{app:sec:cross-architecture-analysis}.}
\label{fig:stable_rank_summary}
\end{figure}

\section{Experiments}
\label{sec:experiments}

We evaluate \textbf{LR-LoRA}, replacing LoRA's fixed low-rank update (\cref{eqn:lora_eqn}), which imposes a strong inductive bias, with our layer-wise learnable update $\phi(\mathbf{BA})$ (\cref{eqn:learnable_nonlin_eqn,sec:method}). We reuse the full training and evaluation harness of \citet{albert2025randlora} from their released codebase and change only the adapter parameterization unless explicitly stated. All methods therefore share identical backbones, adapter placement, data preprocessing, optimization schedules, and evaluation scripts, ensuring that the differences are attributable to the induced update family rather than confounding implementation choices. Throughout, we match the adapter rank hyperparameter $r$ across methods as the capacity budget, and note that \emph{learnable rank} refers to the stable rank induced by learning $\phi$ as explained in \cref{sec:method}.

\paragraph{Scope of evaluation.}
LR-LoRA is evaluated across \textbf{$7$ architectures} spanning $125$M to $13$B parameters, \textbf{$19$ tasks}, and \textbf{four evaluation paradigms}: commonsense reasoning (\cref{tab:aggregated_results}), GLUE transfer (\cref{tab:glue}), vision transfer (\cref{app:sec:vision_eval}), and instruction tuning via MT-Bench (\cref{subsec:instruction_tuning}). Together, these cover both classification-style and open-ended generation evaluation. All experiments fit within a single high-memory ($\geq 40$\,GB) GPU per run; LR-LoRA's training-time overhead over LoRA is $\leq +3\%$ peak memory and $\leq -2\%$ throughput (\cref{subsec:training_efficiency}).

\subsection{Experimental Setup}
\label{subsec:exp_setup}

\textbf{Backbones and data regimes.}
We fine-tune three decoder-only LLM backbones: \texttt{Qwen2\textminus 0.5B}, \texttt{Phi\textminus 3\textminus 8B}, and \texttt{LLaMA3\textminus 8B}. To assess data efficiency, we evaluate two regimes from the reference setup: a \textbf{full} regime of approximately $170$k training examples and a \textbf{subset} regime of approximately $15$k examples \citep{albert2025randlora}. We additionally evaluate CLIP and DINOv2 image classification models; these results are reported in \cref{app:sec:vision_eval,app:sec:clip_eval,app:sec:dinov2_eval}.

\textbf{Tasks and metrics.}
\emph{Commonsense reasoning.}
We evaluate the eight-task suite: BoolQ \citep{clark2019boolq}, PIQA \citep{bisk2020piqa}, SocialIQA \citep{sap2019socialiqa}, HellaSwag \citep{zellers2019hellaswag}, WinoGrande \citep{sakaguchi2021winogrande}, ARC-Easy and ARC-Challenge \citep{clark2018arc}, and OpenBookQA \citep{mihaylov2018openbookqa}. We report per-task accuracy and the average across tasks.

\emph{Natural language understanding (GLUE).}
To test transfer beyond decoder-only LLMs, we evaluate RoBERTa-base and RoBERTa-large on GLUE \citep{wang2019glue,liu2019roberta}, reporting mean $\pm$ std over $3$ seeds.

\emph{Instruction tuning (MT-Bench).}
We additionally fine-tune on Stanford Alpaca and evaluate with MT-Bench \citep{zheng2023judging} (GPT-4 judge) at $r{=}32$ to assess open-ended generation; details in \cref{subsec:instruction_tuning}.

\textbf{Baselines and controlled comparisons.}
We compare against LoRA \citep{hu2022lora}, VeRA \citep{kopiczko2024vera}, RandLoRA \citep{albert2025randlora}, and SineLoRA \citep{ji2024sine}. Efficient LoRA used rank $r{=}16$ and Performant LoRA used rank $r{=}32$. For RandLoRA, we used the same basis ranks and number of bases as reported in Table~9 of \citet{albert2025randlora}. We applied adapters to the attention and MLP projection layers throughout the backbone and did not adapt the final output head where applicable. The adapted module set is identical across all methods for controlled comparisons.

\textbf{Optimization.}
We employed AdamW for all models and maintained consistent optimization schedules across the methods. For decoder-only LLMs, we trained for $3$ epochs with a learning rate $1\times 10^{-4}$ and adapter dropout $0.05$. Batch sizes are $16$, $8$, and $4$ for \texttt{Qwen2\textminus 0.5B}, \texttt{Phi\textminus 3\textminus 8B}, and \texttt{LLaMA3\textminus 8B}, respectively. For GLUE, we fine-tune using AdamW with $(\beta_1,\beta_2){=}(0.9,0.999)$ and weight decay $0.01$, with learning rate tuned over $\{5\times 10^{-5},1\times 10^{-4},2\times 10^{-4}\}$. Each LLM run completes on a single high-memory ($\geq 40$\,GB) GPU. Complete hyperparameter configurations are detailed in \cref{app:sec:hparam_table}.

\textbf{LR-LoRA hyperparameters, initialization, and reproducibility.}
Unless otherwise specified, we employ the sinc basis nonlinearity $\phi$ from \cref{eqn:learnable_nonlin_eqn} with $N{=}50$ uniformly spaced grid points on $[-I,I]$, where $I{=}3$ (\cref{sec:learnable_nonlinearity}). The amplitudes $\alpha_i$ are initialized to zero, and the bandwidths are parameterized as $\omega_i{=}\mathrm{softplus}(\tilde\omega_i)$ with $\tilde\omega_i = \ln(e^{\omega_0}-1)$ so that $\omega_i = \omega_0\in\mathbb{R}_{>0}$ at initialization. This ensures $\phi\equiv 0$ at initialization while keeping $\omega_i$ strictly positive and numerically well defined in \cref{eqn:learnable_nonlin_eqn}. \textbf{Recommended default} (used unchanged across all $7$ architectures and $19$ tasks): $N{=}50$, $I{=}3$, $\alpha_i{=}0$, $\omega_0{=}1.0$. We report mean $\pm$ standard deviation over three seeds where feasible. All values of $(N,I,\omega_0)$ and complete ablations are in \cref{app:sec:ablations,app:sec:basis_ablations,app:sec:spectral_ablations,app:sec:architectural_ablations,app:sec:advanced_ablations}. Statistical significance under paired $t$-tests with Holm-Bonferroni correction across tasks is reported in \cref{app:sec:stat_significance}.

\subsection{Main Results}
\label{subsec:main_results}

\providecommand{\ourres}[2]{\textbf{#1}\,{\scriptsize\textcolor{gray}{$\pm$#2}}}
\providecommand{\sota}[2]{\textbf{#1} \tiny\textcolor{brickred}{$\pm$#2}}

\begin{table}[t]
\centering
\caption{\textbf{Main Results: LR-LoRA vs. strong PEFT baselines on commonsense reasoning.}
Evaluation across three model scales (Qwen2-0.5B, Phi-3-8B, LLaMA3-8B) and two data regimes (15k/170k samples) using eight common-sense tasks  ($r=32$) .
Results show mean $\pm$ \textcolor{brickred}{std} over 3 seeds. \textbf{Bold} indicates best performance.}
\label{tab:aggregated_results}
\vspace{3mm}

\small
\setlength{\tabcolsep}{11pt}
\renewcommand{\arraystretch}{1.25}
\resizebox{0.75\textwidth}{!}{
\begin{tabular}{l cccccc}
\toprule
& \multicolumn{2}{c}{\textbf{Qwen2-0.5B}} & \multicolumn{2}{c}{\textbf{Phi-3-8B}} & \multicolumn{2}{c}{\textbf{LLaMA3-8B}} \\
\cmidrule(lr){2-3} \cmidrule(lr){4-5} \cmidrule(lr){6-7}
\textbf{Method} & \textbf{15k} & \textbf{170k} & \textbf{15k} & \textbf{170k} & \textbf{15k} & \textbf{170k} \\
\midrule
NoLA           & \std{42.55}{0.18} & \std{47.40}{0.15} & \std{80.35}{0.12} & \std{82.33}{0.08} & \std{76.87}{0.14} & \std{81.16}{0.11} \\
VeRA           & \std{48.12}{0.22} & \std{51.78}{0.19} & \std{78.63}{0.16} & \std{81.40}{0.13} & \std{77.14}{0.18} & \std{81.65}{0.12} \\
LoRA & \std{52.34}{0.14} & \std{57.34}{0.11} & \std{80.34}{0.10} & \std{84.96}{0.09} & \std{83.09}{0.13} & \std{85.24}{0.08} \\
AdaLoRA        & \std{53.15}{0.15} & \std{58.20}{0.12} & \std{81.45}{0.13} & \std{85.35}{0.10} & \std{83.50}{0.14} & \std{85.88}{0.11} \\
DyLoRA         & \std{52.75}{0.18} & \std{57.60}{0.14} & \std{80.85}{0.15} & \std{85.05}{0.12} & \std{83.25}{0.16} & \std{85.40}{0.13} \\
ElaLoRA        & \std{54.02}{0.14} & \std{59.15}{0.11} & \std{82.55}{0.12} & \std{86.20}{0.09} & \std{84.55}{0.10} & \std{87.05}{0.08} \\
SineLoRA       & \std{52.51}{0.16} & \std{57.55}{0.12} & \std{81.12}{0.11} & \std{85.01}{0.07} & \std{81.90}{0.15} & \std{85.42}{0.10} \\
RandLoRA       & \std{52.89}{0.13} & \std{57.86}{0.10} & \std{82.33}{0.09} & \std{85.22}{0.08} & \std{81.31}{0.12} & \std{85.59}{0.09} \\
\addlinespace[4pt]
\rowcolor{rowhighlight}
\textbf{LR-LoRA (Ours)} & \sota{55.88}{0.12} & \sota{61.05}{0.09} & \sota{83.90}{0.15} & \sota{87.30}{0.11} & \sota{85.99}{0.08} & \sota{88.22}{0.10} \\
\bottomrule
\end{tabular}
}
\vspace{2mm}
\begin{flushleft}
\end{flushleft}
\end{table}

\textbf{Commonsense reasoning.}
\Cref{tab:aggregated_results} reports the average accuracy over the eight-task suite for each backbone and data regime. LR-LoRA achieves the highest average accuracy in all settings, surpassing every baseline including recent adaptive-rank methods (AdaLoRA, DyLoRA, ElaLoRA). Compared to LoRA at the same adapter rank $r$, LR-LoRA improves average accuracy by \textbf{+2.34} to \textbf{+3.71} points across the six backbone-regime combinations (15k and 170k: Qwen2 +3.54 and +3.71; Phi-3 +3.56 and +2.34; LLaMA3 +2.90 and +2.98). Compared to RandLoRA \citep{albert2025randlora}, the strongest non-adaptive-rank PEFT baseline at this scale, LR-LoRA improves by \textbf{+1.57} to \textbf{+4.68} points (Phi-3 15k: +1.57; Qwen2 15k/170k: +2.99/+3.19; Phi-3 170k: +2.08; LLaMA3 15k/170k: +4.68/+2.63), and consistently exceeds the recent adaptive-rank baselines. \cref{app:fig:svd_spectrum} visualizes the singular value spectra of the learned updates. Detailed per-task results, baseline stable-rank statistics, and singular-value spectra are in \cref{app:sec:extended_results,app:sec:appendix_per_task_tables,app:sec:stable_rank_baselines,app:sec:appendix_spectral_rank,app:tab:commonsense-full,app:tab:stable_rank_baselines}.

\providecommand{\res}[2]{\textbf{#1} \tiny\textcolor{brickred}{$\pm$#2}}

\begin{table}[t]
\centering
\caption{\textbf{GLUE Transfer: LR-LoRA demonstrates consistent improvements on natural language understanding.}
Evaluation of RoBERTa-base ($125$M parameters) and RoBERTa-large ($355$M parameters) across six GLUE tasks (SST-2, MRPC, CoLA, QNLI, RTE, STS-B); LR-LoRA achieves state-of-the-art results among PEFT methods.
The ``Param'' column reports \emph{adapter parameter count} (not backbone model size).
Particularly strong gains on CoLA ($+3.3$ base, $+2.1$ large; computed against the strongest non-LR-LoRA baseline in each column) demonstrate effectiveness beyond decoder-only language modeling.
Results show \textbf{mean $\pm$ \textcolor{brickred}{std}} over 3 seeds.}
\label{tab:glue}
\vspace{3mm}

\small
\setlength{\tabcolsep}{8.5pt} 
\renewcommand{\arraystretch}{1.1}
\resizebox{0.85\textwidth}{!}{
\begin{tabular}{llccccccc}
\toprule
& & \multicolumn{7}{c}{\textbf{\color{headerblue}RoBERTa-base}} \\
\cmidrule(lr){3-9}
\textbf{Method} & \textbf{Params} & SST-2 & MRPC & CoLA & QNLI & RTE & STS-B & \textbf{Average} \\
\midrule
VeRA-1024 & 0.26M & \std{91.9}{0.4} & \std{88.4}{1.2} & \std{59.9}{2.2} & \std{90.5}{0.4} & \std{74.9}{1.5} & \std{90.4}{0.2} & \std{82.7}{0.3} \\
LoRA-4 & 0.7M & \std{94.4}{0.5} & \std{87.3}{0.2} & \std{58.4}{0.8} & \std{92.7}{0.2} & \std{71.5}{1.2} & \std{90.5}{0.1} & \std{82.4}{0.3} \\
SineLoRA-4 & 0.7M & \std{94.1}{0.4} & \std{87.8}{0.9} & \std{59.1}{1.5} & \std{92.2}{0.3} & \std{72.8}{1.4} & \std{90.4}{0.2} & \std{82.7}{0.4} \\
RandLoRA-64 & 0.7M & \std{92.2}{0.3} & \std{88.0}{1.5} & \std{59.4}{2.1} & \std{91.3}{0.4} & \std{74.7}{1.9} & \std{90.3}{0.2} & \std{82.6}{0.5} \\
\rowcolor{rowhighlight}
\textbf{LR-LoRA-4} & 0.7M$^\dagger$ & \res{94.6}{0.3} & \res{89.1}{0.8} & \res{63.2}{1.1} & \res{92.9}{0.2} & \res{76.5}{1.0} & \res{90.8}{0.1} & \res{84.5}{0.3} \\

\midrule
& & \multicolumn{7}{c}{\textbf{\color{headerblue}RoBERTa-large}} \\
\cmidrule(lr){3-9}
VeRA-256 & 0.26M & \std{95.8}{0.3} & \std{89.3}{1.2} & \std{65.3}{1.1} & \std{94.1}{0.3} & \std{81.6}{0.8} & \std{91.8}{0.1} & \std{86.3}{0.3} \\
LoRA-4 & 1.8M & \std{95.5}{0.2} & \std{87.2}{0.7} & \std{64.7}{1.2} & \std{94.5}{0.1} & \std{83.6}{0.4} & \std{91.8}{0.1} & \std{86.2}{0.3} \\
SineLoRA-4 & 1.8M & \std{95.6}{0.2} & \std{88.9}{0.6} & \std{66.1}{0.9} & \std{94.4}{0.2} & \std{83.9}{0.5} & \std{91.6}{0.3} & \std{86.8}{0.2} \\
RandLoRA-100 & 1.8M & \std{95.5}{0.3} & \std{90.1}{0.4} & \std{67.4}{0.3} & \std{94.1}{0.3} & \std{84.5}{0.3} & \std{91.4}{0.6} & \std{87.2}{0.1} \\
\rowcolor{rowhighlight}
\textbf{LR-LoRA-4} & 1.8M$^\dagger$ & \res{96.1}{0.2} & \res{90.8}{0.5} & \res{69.5}{0.8} & \res{94.9}{0.1} & \res{86.2}{0.5} & \res{92.1}{0.1} & \res{88.3}{0.2} \\
\bottomrule
\end{tabular}
}
\vspace{2mm}
\begin{flushleft}
\scriptsize $^\dagger$ $\Delta$ denotes additional $\phi$ parameters for the layer-wise transfer functions; see~\cref{app:sec:efficiency_protocol} for details.
\end{flushleft}
\end{table}

\textbf{GLUE transfer.}
\Cref{tab:glue} reports GLUE results for RoBERTa-base and RoBERTa-large, using identical adapted modules and identical tuning grids across methods. LR-LoRA attains the highest average among the compared PEFT methods on both backbones, improving the average by \textbf{+1.8} points on RoBERTa-base and \textbf{+1.1} points on RoBERTa-large, relative to the strongest baseline. The largest gains appear on \textbf{CoLA} (+3.3 base, +2.1 large) and are consistently observed on \textbf{RTE} (+1.6 base, +1.7 large), indicating that the $\phi(\mathbf{BA})$ update family transfers beyond decoder-only fine-tuning.

\paragraph{Inverse scaling with base rank.}
Across base ranks $r \in \{4,8,16,32,64\}$ on Qwen2-0.5B (170k), LR-LoRA's absolute gain over LoRA is \textbf{+5.3}, \textbf{+4.2}, \textbf{+3.8}, \textbf{+3.2}, \textbf{+2.7} respectively (\cref{app:tab:ablation_rank,app:sec:rank_interaction_ablation}): largest where the fixed-rank constraint is most binding, shrinking monotonically as $r$ approaches sufficient capacity. A rank-independent regularizer or a fixed nonlinearity would not produce this monotonic relationship.

\paragraph{Instruction tuning (MT-Bench).}
\label{subsec:instruction_tuning}
To evaluate LR-LoRA on open-ended generation, we fine-tune on Stanford Alpaca and evaluate with MT-Bench \citep{zheng2023judging} (GPT-4 judge, $r{=}32$, identical adapter placement and schedule), covering instruction following, long-form generation, code, and harder reasoning. On \texttt{Qwen2-0.5B}, LR-LoRA achieves \textbf{4.61} vs.\ LoRA $4.19$, SineLoRA $4.26$, RandLoRA $4.33$, a $+0.42$ gain over LoRA. On \texttt{LLaMA3-8B}, LR-LoRA achieves \textbf{6.55} vs.\ LoRA $6.12$, SineLoRA $6.22$, RandLoRA $6.31$, a $+0.43$ gain over LoRA. Gains are consistent across all eight MT-Bench categories with the largest improvements in \emph{Reasoning} and \emph{Math} (full breakdown in \cref{app:tab:mtbench,app:sec:mtbench}).

\paragraph{Training efficiency.}
\label{subsec:training_efficiency}
LR-LoRA's per-step adapter overhead scales as $(r{+}N)/(b\cdot s)$ relative to the backbone matmul, since $\phi$ is applied to the $m{\times}n$ matrix $\mathbf{BA}$ once per step rather than to the $b{\cdot}s$-scaled activations (full derivation, backpropagation analysis, and per-architecture memory accounting in \cref{app:sec:computational_complexity,app:sec:appendix_efficiency,app:sec:efficiency_protocol,app:sec:detailed_params}). For our settings this gives $\approx 1$ to $4\%$ overhead, with measured throughput changes of $-0.5\%$, $-0.7\%$, $-1.0\%$ on \texttt{Qwen2-0.5B}, \texttt{Phi-3-8B}, \texttt{LLaMA3-8B} (well below the closed-form bound due to backbone amortization), with $\leq +3\%$ peak memory and $\leq +1\%$ additional adapter parameters. At inference, $\phi(\mathbf{BA})$ is precomputed and merged into $\mathbf{W}$, yielding zero overhead.

\subsection{Ablations}
\label{subsec:ablations}

We ablate the LR-LoRA design components implied in \cref{sec:learnable_nonlinearity} while fixing the backbone, adapter placement, rank $r$, data pipeline, and optimization schedule. \Cref{tab:ablations} summarizes the core ablations; extended sweeps are in \cref{app:sec:ablations}.

\textbf{Grid resolution ($N$).}
$N \in \{32, 50, 64\}$ yields $70.8$, $71.4$, $71.2$ accuracy, respectively, with quick saturation. Active basis utilization is sublinear in $N$: $51.3$ of $128$ elements active at $N{=}128$, $68.1$ at $N{=}512$ (\cref{app:tab:basis_size_ablation,app:sec:basis_size_ablation,app:sec:grid_strategies,app:sec:grid_placement_ablation}), so moderate $N$ acts as a natural regularizer.

\textbf{Learning bandwidths.}
Fixing $\omega$ while keeping $\alpha$ learnable recovers only $+2.3$ over LoRA versus $+3.6$ for full LR-LoRA; the $+1.3$ gap (${\sim}36\%$ of total gain) is directly attributable to learning $\omega$ and is the largest single-factor degradation in our ablations (\cref{app:sec:bandwidth_ablation,app:sec:amplitude_ablation}).

\textbf{Initialization.}
Near-zero amplitude initialization yields smoother early-training dynamics and more stable rank allocation than small random initialization, particularly in low-data regimes (\cref{app:sec:ablations}).

\textbf{Sensitivity to design choices.}
Joint $\pm 50\%$ perturbation of all hyperparameters $(N, I, \omega_0)$ changes accuracy by at most $-0.8 \pm 0.6$ points (\cref{app:tab:robustness_ablation,app:sec:robustness_ablation,app:sec:scaling_ablation}), confirming the default $(N{=}50, I{=}3, \alpha_i{=}0, \omega_0{=}1.0)$ is robust without per-task tuning.

\begin{table}[t]
\centering
\caption{\textbf{Ablation study results.} Impact of LR-LoRA design choices on commonsense reasoning average accuracy under a single-seed ablation protocol; \emph{deltas vs LoRA at matched protocol} are the comparable quantity (column 3). Absolute values use the appendix ablation harness and therefore differ in scale from the multi-seed full-suite averages in \cref{tab:aggregated_results}; see \cref{app:sec:ablations} for the protocol and full sweeps.}
\label{tab:ablations}
\vspace{2mm}

\small
\renewcommand{\arraystretch}{1.15}
\resizebox{0.5\linewidth}{!}{%
\begin{tabular}{@{}lccc@{}}
\toprule
\textbf{Configuration} & \textbf{Accuracy} & \textbf{vs. LoRA} & \textbf{Best} \\
\midrule
LR-LoRA (full) & \textbf{71.4 $\pm$ 0.2} & \textbf{+3.6} & - \\
\midrule
\textit{Grid resolution ($N$)} & & & \\
\quad $N = 32$ & 70.8 $\pm$ 0.3 & +3.0 & -0.6 \\
\quad $N = 50$ (default) & 71.4 $\pm$ 0.2 & +3.6 & - \\
\quad $N = 64$ & 71.2 $\pm$ 0.3 & +3.4 & -0.2 \\
\midrule
\textit{Initialization} & & & \\
\quad Zero $\alpha_i$ (default) & 71.4 $\pm$ 0.2 & +3.6 & - \\
\quad Random $\alpha_i$ & 70.9 $\pm$ 0.4 & +3.1 & -0.5 \\
\midrule
\textit{Bandwidth learning} & & & \\
\quad Learn $(\alpha, \omega)$ (default) & 71.4 $\pm$ 0.2 & +3.6 & - \\
\quad Fixed $\omega$ & 70.1 $\pm$ 0.3 & +2.3 & -1.3 \\
\midrule
LoRA baseline & 67.8 $\pm$ 0.2 & - & -3.6 \\
\bottomrule
\end{tabular}%
}
\end{table}

\subsection{Layer-Wise Heterogeneity in Learned Updates}
\label{subsec:layer_insights}

Early layers exhibit lower stable ranks and later layers higher; attention and MLP follow distinct depth profiles (\cref{fig:stable_rank_summary}; per-checkpoint evolution and the learned $\phi(z)$ at representative early/mid/late layers in \cref{app:sec:cross-architecture-analysis,app:sec:analysis_protocol,app:sec:qwen2_analysis,app:sec:llama3_analysis,app:sec:cross_arch_comparison,app:sec:scaling_observations}; loss-landscape diagnostics in \cref{app:sec:appendix_analysis,app:sec:loss_landscape,app:fig:loss_landscape_3d}). We provide three direct pieces of evidence linking this variation to performance.

\paragraph{Stable rank: variance, not just mean.}
On \texttt{Qwen2-0.5B} at $r{=}32$, mean$\pm$std stable rank across layers (attention\,/\,MLP): LoRA $4.8{\pm}1.1$\,/\,$5.1{\pm}1.0$, SineLoRA $6.2{\pm}1.4$\,/\,$6.4{\pm}1.3$, RandLoRA $7.1{\pm}1.6$\,/\,$7.3{\pm}1.5$, LR-LoRA $\mathbf{9.4{\pm}3.8}$\,/\,$\mathbf{10.7{\pm}4.2}$ (\cref{app:tab:stable_rank_baselines}). SineLoRA and RandLoRA raise stable rank \emph{uniformly} ($\pm 1.3$ to $\pm 1.6$); LR-LoRA produces \emph{heterogeneous} per-layer ranks ($\pm 3.8$ to $\pm 4.2$), evidence the optimizer is allocating capacity differently per layer rather than uniformly amplifying expressivity.

\paragraph{Functional consequence of per-layer adaptation.}
To isolate this benefit \emph{without} relying on stable rank as a proxy, we progressively share sinc parameters across layers while preserving the nonlinearity (\cref{app:tab:parameter_sharing_ablation,app:sec:parameter_sharing_ablation}). Accuracy decreases monotonically as per-layer adaptation is reduced: full per-layer learning $71.4{\pm}0.2$, share $x_i$ $71.2{\pm}0.2$, share $\omega$ $70.8{\pm}0.3$, share $\alpha_i$ $69.1{\pm}0.5$, share all $67.8{\pm}0.7$. With all sinc parameters tied (identical transfer function everywhere, nonlinearity preserved), accuracy returns to LoRA's level, showing that the functional benefit scales with the \emph{degree of per-layer adaptation}, not with the presence of nonlinearity.

\paragraph{Module-wise placement.}
Restricting LR-LoRA to attention-only or MLP-only costs $-1.3$ and $-0.9$ points (\cref{app:tab:layerwise_strategy_ablation,app:sec:layerwise_strategy_ablation}); comparing each restricted setting to LoRA at the same placement (\cref{app:tab:component_ablation,app:sec:component_ablation}) yields $+1.1$ (attention) and $+1.9$ (MLP), versus $+3.6$ combined, a superadditivity ($1.1{+}1.9{<}3.6$). Consistent with \cref{fig:stable_rank_summary}, both attention and MLP projections develop layer-dependent stable ranks that grow well above LoRA's at the same nominal $r$, with distinct depth profiles, so the two module types carry distinct rank requirements that a uniform fixed rank cannot simultaneously satisfy.

\section{Conclusion and Limitations}
\label{sec:conclusion}

We questioned whether the fixed low-rank constraint underlying LoRA is the most effective inductive bias for parameter-efficient fine-tuning, and answered in the negative. \textbf{LR-LoRA} replaces LoRA's fixed-rank update with $\phi(\mathbf{BA})$, an elementwise sinc-basis nonlinearity whose bandwidths and amplitudes are optimized jointly with $\mathbf{A}$ and $\mathbf{B}$. The principle is basis-agnostic: any differentiable elementwise transformation applied to $\mathbf{BA}$ relaxes the rank-$r$ constraint, and our controlled experiments isolate \emph{learnable per-layer adaptation}, not nonlinearity per se, as the source of the gains (\cref{subsec:ablations,subsec:layer_insights}). Tying sinc parameters across depth recovers LoRA-level accuracy even with the nonlinearity preserved, the bandwidth term accounts for ${\sim}{1/3}$ of the gain, and MLP carries the largest single-module contribution.

\paragraph{Limitations and Future Work.} The default $(N{=}50, I{=}3, \alpha_i{=}0, \omega_0{=}1.0)$ is robust to $\pm 50\%$ joint perturbation (\cref{app:sec:ablations}) but adds $(N,I)$ over LoRA, and our evidence is empirical: formal approximation bounds for $\phi(\mathbf{BA})$ and extensions to MoE, state-space, and multimodal backbones, together with quantization and inference-time merging, remain open.


\bibliographystyle{plainnat}
\bibliography{main}

@article{hu2022lora,
  title={{LoRA}: Low-rank adaptation of large language models},
  author={Hu, Edward J and Shen, Yelong and Wallis, Phillip and Allen-Zhu, Zeyuan and Li, Yuanzhi and Wang, Shean and Wang, Lu and Chen, Weizhu and others},
  journal={ICLR},
  volume={1},
  number={2},
  pages={3},
  year={2022}
}

@article{zhang2023adalora,
  title={{AdaLoRA}: Adaptive budget allocation for parameter-efficient fine-tuning},
  author={Zhang, Qingru and Chen, Minshuo and Bukharin, Alexander and Karampatziakis, Nikos and He, Pengcheng and Cheng, Yu and Chen, Weizhu and Zhao, Tuo},
  journal={arXiv preprint arXiv:2303.10512},
  year={2023}
}

@inproceedings{valipour2023dylora,
  title={DyLoRA: Parameter-efficient tuning of pre-trained models using dynamic search-free low-rank adaptation},
  author={Valipour, Mojtaba and Rezagholizadeh, Mehdi and Kobyzev, Ivan and Ghodsi, Ali},
  booktitle={Proceedings of the 17th Conference of the European Chapter of the Association for Computational Linguistics},
  pages={3274--3287},
  year={2023}
}

@inproceedings{liu2024dora,
  title={{DoRA}: Weight-decomposed low-rank adaptation},
  author={Liu, Shih-Yang and Wang, Chien-Yi and Yin, Hongxu and Molchanov, Pavlo and Wang, Yu-Chiang Frank and Cheng, Kwang-Ting and Chen, Min-Hung},
  booktitle={Forty-first International Conference on Machine Learning},
  year={2024}
}

@article{kopiczko2024vera,
  title={{VeRA}: Vector-based random matrix adaptation},
  author={Kopiczko, Dawid J and Blankevoort, Tijmen and Asano, Yuki M},
  journal={arXiv preprint arXiv:2310.11454},
  year={2023}
}

@inproceedings{teney2025we,
  title={Do we always need the simplicity bias? looking for optimal inductive biases in the wild},
  author={Teney, Damien and Jiang, Liangze and Gogianu, Florin and Abbasnejad, Ehsan},
  booktitle={Proceedings of the Computer Vision and Pattern Recognition Conference},
  pages={79--90},
  year={2025}
}

@inproceedings{albert2025towards,
  title={Towards Higher Effective Rank in Parameter-Efficient Fine-tuning using Khatri-Rao Product},
  author={Albert, Paul and Zhang, Frederic Z and Saratchandran, Hemanth and van den Hengel, Anton and Abbasnejad, Ehsan},
  booktitle={Proceedings of the IEEE/CVF International Conference on Computer Vision},
  pages={1292--1302},
  year={2025}
}

@article{meng2024pissa,
  title={{PiSSA}: Principal singular values and singular vectors adaptation of large language models},
  author={Meng, Fanxu and Wang, Zhaohui and Zhang, Muhan},
  journal={Advances in Neural Information Processing Systems},
  volume={37},
  pages={121038--121072},
  year={2024}
}

@article{hayou2024loraplus,
  title={{LoRA+}: Efficient low rank adaptation of large models},
  author={Hayou, Soufiane and Ghosh, Nikhil and Yu, Bin},
  journal={arXiv preprint arXiv:2402.12354},
  year={2024}
}

@article{dettmers2023qlora,
  title={{QLoRA}: Efficient finetuning of quantized {LLMs}},
  author={Dettmers, Tim and Pagnoni, Artidoro and Holtzman, Ari and Zettlemoyer, Luke},
  journal={Advances in neural information processing systems},
  volume={36},
  pages={10088--10115},
  year={2023}
}

@article{zhao2024galore,
  title={{GaLore}: Memory-efficient {LLM} training by gradient low-rank projection},
  author={Zhao, Jiawei and Zhang, Zhenyu and Chen, Beidi and Wang, Zhangyang and Anandkumar, Anima and Tian, Yuandong},
  journal={arXiv preprint arXiv:2403.03507},
  year={2024}
}

@article{lialin2023relora,
  title={{ReLoRA}: High-rank training through low-rank updates},
  author={Lialin, Vladislav and Shivagunde, Namrata and Muckatira, Sherin and Rumshisky, Anna},
  journal={arXiv preprint arXiv:2307.05695},
  year={2023}
}

@article{zhang2023lorafa,
  title={{LoRA-FA}: Memory-efficient low-rank adaptation for large language models fine-tuning},
  author={Zhang, Longteng and Zhang, Lin and Shi, Shaohuai and Chu, Xiaowen and Li, Bo},
  journal={arXiv preprint arXiv:2308.03303},
  year={2023}
}

@article{kalajdzievski2023rslora,
  title={A rank stabilization scaling factor for fine-tuning with {LoRA}},
  author={Kalajdzievski, Damjan},
  journal={arXiv preprint arXiv:2312.03732},
  year={2023}
}

@inproceedings{albert2025randlora,
  title={RandLoRA: Full rank parameter-efficient fine-tuning of large models},
  author={Albert, Paul and Zhang, Frederic Z and Saratchandran, Hemanth and Rodriguez-Opazo, Cristian and van den Hengel, Anton and Abbasnejad, Ehsan},
  booktitle={The Thirteenth International Conference on Learning Representations},
  year={2025}
}

@inproceedings{ji2024sine,
  title={Efficient Learning with Sine-Activated Low-Rank Matrices},
  author={Ji, Yiping and Saratchandran, Hemanth and Gordon, Cameron and Zhang, Zeyu and Lucey, Simon},
  booktitle={The Thirteenth International Conference on Learning Representations},
  year={2025}
}

@inproceedings{aghajanyan2021intrinsicdimension,
  title={Intrinsic dimensionality explains the effectiveness of language model fine-tuning},
  author={Aghajanyan, Armen and Gupta, Sonal and Zettlemoyer, Luke},
  booktitle={Proceedings of the 59th annual meeting of the association for computational linguistics and the 11th international joint conference on natural language processing (volume 1: long papers)},
  pages={7319--7328},
  year={2021}
}

@inproceedings{fu2023effectiveness,
  title={On the effectiveness of parameter-efficient fine-tuning},
  author={Fu, Zihao and Yang, Haoran and So, Anthony Man-Cho and Lam, Wai and Bing, Lidong and Collier, Nigel},
  booktitle={Proceedings of the AAAI conference on artificial intelligence},
  volume={37},
  pages={12799--12807},
  year={2023}
}

@article{touvron2023llama,
  title={{LLaMA}: Open and efficient foundation language models},
  author={Touvron, Hugo and Lavril, Thibaut and Izacard, Gautier and Martinet, Xavier and Lachaux, Marie-Anne and Lacroix, Timoth{\'e}e and Rozi{\`e}re, Baptiste and Goyal, Naman and Hambro, Eric and Azhar, Faisal and others},
  journal={arXiv preprint arXiv:2302.13971},
  year={2023}
}

@article{grattafiori2024llama3,
  title={The {LLaMA} 3 herd of models},
  author={Grattafiori, Aaron and Dubey, Abhimanyu and Jauhri, Abhinav and Pandey, Abhinav and Kadian, Abhishek and Al-Dahle, Ahmad and Letman, Aiesha and Mathur, Akhil and Schelten, Alan and Vaughan, Alex and others},
  journal={arXiv preprint arXiv:2407.21783},
  year={2024}
}

@article{yang2024qwen2,
  title={{Qwen2} technical report},
  author={Team, Qwen and others},
  journal={arXiv preprint arXiv:2407.10671},
  volume={2},
  number={3},
  year={2024}
}

@misc{abdin2024phi3,
      title={Phi-3 Technical Report: A Highly Capable Language Model Locally on Your Phone}, 
      author={Marah Abdin and Jyoti Aneja and Hany Awadalla and Ahmed Awadallah and Ammar Ahmad Awan and Nguyen Bach and Amit Bahree and Arash Bakhtiari and Jianmin Bao and Harkirat Behl and Alon Benhaim and Misha Bilenko and Johan Bjorck and Sébastien Bubeck and Martin Cai and Qin Cai and Vishrav Chaudhary and Dong Chen and Dongdong Chen and Weizhu Chen and Yen-Chun Chen and Yi-Ling Chen and Hao Cheng and Parul Chopra and Xiyang Dai and Matthew Dixon and Ronen Eldan and Victor Fragoso and Jianfeng Gao and Mei Gao and Min Gao and Amit Garg and Allie Del Giorno and Abhishek Goswami and Suriya Gunasekar and Emman Haider and Junheng Hao and Russell J. Hewett and Wenxiang Hu and Jamie Huynh and Dan Iter and Sam Ade Jacobs and Mojan Javaheripi and Xin Jin and Nikos Karampatziakis and Piero Kauffmann and Mahoud Khademi and Dongwoo Kim and Young Jin Kim and Lev Kurilenko and James R. Lee and Yin Tat Lee and Yuanzhi Li and Yunsheng Li and Chen Liang and Lars Liden and Xihui Lin and Zeqi Lin and Ce Liu and Liyuan Liu and Mengchen Liu and Weishung Liu and Xiaodong Liu and Chong Luo and Piyush Madan and Ali Mahmoudzadeh and David Majercak and Matt Mazzola and Caio César Teodoro Mendes and Arindam Mitra and Hardik Modi and Anh Nguyen and Brandon Norick and Barun Patra and Daniel Perez-Becker and Thomas Portet and Reid Pryzant and Heyang Qin and Marko Radmilac and Liliang Ren and Gustavo de Rosa and Corby Rosset and Sambudha Roy and Olatunji Ruwase and Olli Saarikivi and Amin Saied and Adil Salim and Michael Santacroce and Shital Shah and Ning Shang and Hiteshi Sharma and Yelong Shen and Swadheen Shukla and Xia Song and Masahiro Tanaka and Andrea Tupini and Praneetha Vaddamanu and Chunyu Wang and Guanhua Wang and Lijuan Wang and Shuohang Wang and Xin Wang and Yu Wang and Rachel Ward and Wen Wen and Philipp Witte and Haiping Wu and Xiaoxia Wu and Michael Wyatt and Bin Xiao and Can Xu and Jiahang Xu and Weijian Xu and Jilong Xue and Sonali Yadav and Fan Yang and Jianwei Yang and Yifan Yang and Ziyi Yang and Donghan Yu and Lu Yuan and Chenruidong Zhang and Cyril Zhang and Jianwen Zhang and Li Lyna Zhang and Yi Zhang and Yue Zhang and Yunan Zhang and Xiren Zhou},
      year={2024},
      eprint={2404.14219},
      archivePrefix={arXiv},
      primaryClass={cs.CL},
      url={https://arxiv.org/abs/2404.14219}, 
}

@article{ding2023parameter,
  title={Parameter-efficient fine-tuning of large-scale pre-trained language models},
  author={Ding, Ning and Qin, Yujia and Yang, Guang and Wei, Fuchao and Yang, Zonghan and Su, Yusheng and Hu, Shengding and Chen, Yulin and Chan, Chi-Min and Chen, Weize and others},
  journal={Nature machine intelligence},
  volume={5},
  number={3},
  pages={220--235},
  year={2023},
  publisher={Nature Publishing Group UK London}
}

@article{han2024peftsurvey,
  title={Parameter-efficient fine-tuning for large models: A comprehensive survey},
  author={Han, Zeyu and Gao, Chao and Liu, Jinyang and Zhang, Jeff and Zhang, Sai Qian},
  journal={arXiv preprint arXiv:2403.14608},
  year={2024}
}

@inproceedings{houlsby2019adapter,
  title={Parameter-efficient transfer learning for NLP},
  author={Houlsby, Neil and Giurgiu, Andrei and Jastrzebski, Stanislaw and Morrone, Bruna and De Laroussilhe, Quentin and Gesmundo, Andrea and Attariyan, Mona and Gelly, Sylvain},
  booktitle={International conference on machine learning},
  pages={2790--2799},
  year={2019},
  organization={PMLR}
}

@inproceedings{wang2019glue,
  title={GLUE: A multi-task benchmark and analysis platform for natural language understanding},
  author={Wang, Alex and Singh, Amanpreet and Michael, Julian and Hill, Felix and Levy, Omer and Bowman, Samuel},
  booktitle={Proceedings of the 2018 EMNLP workshop BlackboxNLP: Analyzing and interpreting neural networks for NLP},
  pages={353--355},
  year={2018}
}

@article{zellers2019hellaswag,
  title={{HellaSwag}: Can a machine really finish your sentence?},
  author={Zellers, Rowan and Holtzman, Ari and Bisk, Yonatan and Farhadi, Ali and Choi, Yejin},
  journal={arXiv preprint arXiv:1905.07830},
  year={2019}
}

@article{shannon1949communication,
  title={Communication in the presence of noise},
  author={Shannon, Claude E},
  journal={Proceedings of the IRE},
  volume={37},
  number={1},
  pages={10--21},
  year={2006},
  publisher={IEEE}
}

@article{li2021prefix,
  title={Prefix-tuning: Optimizing continuous prompts for generation},
  author={Li, Xiang Lisa and Liang, Percy},
  journal={arXiv preprint arXiv:2101.00190},
  year={2021}
}

@article{lester2021prompt,
  title={The power of scale for parameter-efficient prompt tuning},
  author={Lester, Brian and Al-Rfou, Rami and Constant, Noah},
  journal={arXiv preprint arXiv:2104.08691},
  year={2021}
}

@inproceedings{zaken2021bitfit,
  title={{BitFit}: Simple parameter-efficient fine-tuning for transformer-based masked language-models},
  author={Zaken, Elad Ben and Goldberg, Yoav and Ravfogel, Shauli},
  booktitle={Proceedings of the 60th Annual Meeting of the Association for Computational Linguistics (Volume 2: Short Papers)},
  pages={1--9},
  year={2022}
}

@article{clark2019boolq,
  title={{BoolQ}: Exploring the surprising difficulty of natural yes/no questions},
  author={Clark, Christopher and Lee, Kenton and Chang, Ming-Wei and Kwiatkowski, Tom and Collins, Michael and Toutanova, Kristina},
  journal={arXiv preprint arXiv:1905.10044},
  year={2019}
}

@inproceedings{bisk2020piqa,
  title={{PIQA}: Reasoning about physical commonsense in natural language},
  author={Bisk, Yonatan and Zellers, Rowan and Gao, Jianfeng and Choi, Yejin and others},
  booktitle={Proceedings of the AAAI conference on artificial intelligence},
  volume={34},
  pages={7432--7439},
  year={2020}
}

@article{sap2019socialiqa,
  title={{SocialIQA}: Commonsense reasoning about social interactions},
  author={Sap, Maarten and Rashkin, Hannah and Chen, Derek and LeBras, Ronan and Choi, Yejin},
  journal={arXiv preprint arXiv:1904.09728},
  year={2019}
}

@article{sakaguchi2021winogrande,
  title={{WinoGrande}: An adversarial Winograd schema challenge at scale},
  author={Sakaguchi, Keisuke and Bras, Ronan Le and Bhagavatula, Chandra and Choi, Yejin},
  journal={Communications of the ACM},
  volume={64},
  number={9},
  pages={99--106},
  year={2021},
  publisher={ACM New York, NY, USA}
}

@article{clark2018arc,
  title={Think you have solved question answering? try arc, the ai2 reasoning challenge},
  author={Clark, Peter and Cowhey, Isaac and Etzioni, Oren and Khot, Tushar and Sabharwal, Ashish and Schoenick, Carissa and Tafjord, Oyvind},
  journal={arXiv preprint arXiv:1803.05457},
  year={2018}
}

@article{mihaylov2018openbookqa,
  title={Can a suit of armor conduct electricity? a new dataset for open book question answering},
  author={Mihaylov, Todor and Clark, Peter and Khot, Tushar and Sabharwal, Ashish},
  journal={arXiv preprint arXiv:1809.02789},
  year={2018}
}

@article{liu2019roberta,
  title={{RoBERTa}: A robustly optimized {BERT} pretraining approach},
  author={Liu, Yinhan and Ott, Myle and Goyal, Naman and Du, Jingfei and Joshi, Mandar and Chen, Danqi and Levy, Omer and Lewis, Mike and Zettlemoyer, Luke and Stoyanov, Veselin},
  journal={arXiv preprint arXiv:1907.11692},
  year={2019}
}

@article{liu2022few,
  title={Few-shot parameter-efficient fine-tuning is better and cheaper than in-context learning},
  author={Liu, Haokun and Tam, Derek and Muqeeth, Mohammed and Mohta, Jay and Huang, Tenghao and Bansal, Mohit and Raffel, Colin A},
  journal={Advances in Neural Information Processing Systems},
  volume={35},
  pages={1950--1965},
  year={2022}
}

@article{nyquist1928certain,
  title={Certain topics in telegraph transmission theory},
  author={Nyquist, Harry},
  journal={Transactions of the American Institute of Electrical Engineers},
  volume={47},
  number={2},
  pages={617--644},
  year={1928},
  publisher={IEEE}
}

@article{martin1997introduction,
  title={An introduction to Shannon sampling and interpolation theory, with generalizations to nonuniform sampling},
  author={Martin, RJ},
  journal={GEC Journal of Technology},
  volume={14},
  number={1},
  pages={19--26},
  year={1997}
}

@book{stein2009real,
  title={Real analysis: measure theory, integration, and Hilbert spaces},
  author={Stein, Elias M and Shakarchi, Rami},
  year={2009},
  publisher={Princeton University Press}
}

@article{chang2025elalora,
  title={Elalora: Elastic \& learnable low-rank adaptation for efficient model fine-tuning},
  author={Chang, Huandong and Ma, Zicheng and Ma, Mingyuan and Qi, Zhenting and Sabot, Andrew and Jiang, Hong and Kung, HT},
  journal={arXiv preprint arXiv:2504.00254},
  year={2025}
}

@article{dong2025aurora,
  title={AuroRA: Breaking Low-Rank Bottleneck of LoRA with Nonlinear Mapping},
  author={Dong, Haonan and Zhu, Wenhao and Song, Guojie and Wang, Liang},
  journal={arXiv preprint arXiv:2505.18738},
  year={2025}
}

@article{singhal2025abba,
  title={ABBA: Highly Expressive Hadamard Product Adaptation for Large Language Models},
  author={Singhal, Raghav and Ponkshe, Kaustubh and Vartak, Rohit and Vepakomma, Praneeth},
  journal={arXiv preprint arXiv:2505.14238},
  year={2025}
}

@inproceedings{zheng2023judging,
  title={Judging {LLM-as-a-Judge} with {MT-Bench} and Chatbot Arena},
  author={Zheng, Lianmin and Chiang, Wei-Lin and Sheng, Ying and Zhuang, Siyuan and Wu, Zhanghao and Zhuang, Yonghao and Lin, Zi and Li, Zhuohan and Li, Dacheng and Xing, Eric P and Zhang, Hao and Gonzalez, Joseph E and Stoica, Ion},
  booktitle={Advances in Neural Information Processing Systems (NeurIPS)},
  volume={36},
  year={2023}
}


\clearpage
\appendix
\section*{Supplementary Material}

\section{Sampling-Theory Background}
\label{app:sec:sampling}

This appendix expands on the brief sampling-theory background in \cref{sec:sampling}. We work in $L^2(\R)$ with inner product $\langle f,g\rangle=\int f g\,dx$. A function $f\in L^2(\R)$ is bandlimited if its Fourier transform vanishes outside a bounded interval $[-\omega_{\max},\omega_{\max}]$. The Nyquist-Shannon sampling theorem (\cref{eqn:original_shannon_recon}) then gives exact reconstruction from uniform samples; truncating the sum to $N$ terms incurs bounded error that decays as $N$ grows. The expansion $\sum_i \alpha_i\,\mathrm{sinc}(\omega_i(x-x_i))$ admits per-basis amplitude and bandwidth, which we treat as learnable. The construction is illustrated in main-paper \cref{fig:sinc_representation}.

All appendix results and analyses use LR-LoRA exactly as defined in \cref{sec:method,sec:learnable_nonlinearity} and the experimental protocol in \cref{sec:experiments}; no new adaptation mechanism is introduced in the appendix.

\section{Extended Experimental Results}
\label{app:sec:extended_results}

This section presents a comprehensive set of experimental results that support the primary findings of this study. We offer detailed breakdowns for each task, conduct a cross-architecture consistency analysis, and provide extended baseline comparisons to illustrate the robustness and generalizability of LR-LoRA's performance enhancements of LR-LoRA. All experiments adhered to the controlled protocol outlined in \cref{subsec:exp_setup}, ensuring an equitable comparisons.

\subsection{Per-Task Results (Full Tables)}\label{app:sec:appendix_per_task_tables}

\textbf{Comprehensive task breakdown methodology.} \Cref{app:tab:commonsense-full} presents detailed per-task results across all eight commonsense reasoning tasks, highlighting the consistency and robustness of LR-LoRA's improvements. This granular analysis is crucial for discerning whether performance gains arise from exceptional performance on specific tasks or represent systematic improvements across various reasoning scenarios.

\textbf{Task-specific performance patterns.} The results reveal systematic patterns in LR-LoRA's benefits: (1) complex reasoning tasks (HellaSwag, WinoGrande) exhibit the largest gains (+3.5-4.2\%), indicating particular effectiveness for compositional understanding; (2) factual knowledge tasks (ARC-Easy, OpenBookQA) show consistent improvements (+2.1-2.8\%); and (3) social reasoning tasks (SocialIQA, PIQA) demonstrate robust gains (+2.4-3.1\%), validating effectiveness across reasoning domains.

\subsection{Statistical Significance}\label{app:sec:stat_significance}
We report mean $\pm$ standard deviation over three seeds where feasible. Statistical analysis uses paired $t$-tests across tasks, comparing task-wise mean performance differences between LR-LoRA and baselines, with Holm-Bonferroni correction for multiple comparisons (family-wise error rate $\alpha = 0.05$). Across the eight commonsense tasks, all six backbone-regime gains of LR-LoRA over LoRA in \cref{tab:aggregated_results} reach significance after correction; gains over the strongest non-LR-LoRA baseline in each column are significant in five of six settings. The consistency across tasks, models, and data regimes provides convergent evidence for the effectiveness of learned rank adaptation compared to fixed-rank approaches.

\subsection{Per-Baseline Stable-Rank Statistics}\label{app:sec:stable_rank_baselines}
\Cref{app:tab:stable_rank_baselines} reports the per-baseline mean and per-layer standard deviation of the stable rank $\mathcal{S}(\cdot)$ on \texttt{Qwen2-0.5B} at $r{=}32$. Two findings are visible: (i) all expressivity-enhancing methods (SineLoRA, RandLoRA, LR-LoRA) raise the mean stable rank above LoRA, but (ii) only LR-LoRA produces \emph{heterogeneous} per-layer ranks, with the per-layer std nearly tripling that of the other baselines. Heterogeneity, not mean expressivity, is the distinguishing property of learnable rank.

\begin{table}[h!]
\centering
\caption{\textbf{Per-baseline stable-rank statistics on \texttt{Qwen2-0.5B} at $r{=}32$ (commonsense reasoning, 170k regime).}
We report mean $\pm$ standard deviation of the stable rank $\mathcal{S}(\cdot)$ (\cref{eqn:stable_rank}) across all adapted modules of each module type, computed at the final training checkpoint.
SineLoRA and RandLoRA raise stable rank \emph{uniformly} relative to LoRA (the per-layer std remains close to LoRA's), whereas LR-LoRA produces \emph{heterogeneous} per-layer ranks (per-layer std roughly triples).
This is direct evidence that the optimizer allocates capacity differently per layer rather than uniformly amplifying expressivity.}
\label{app:tab:stable_rank_baselines}
\vspace{2mm}

\small
\setlength{\tabcolsep}{8pt}
\renewcommand{\arraystretch}{1.15}
\begin{adjustbox}{max width=\linewidth}
\begin{tabular}{lcc}
\toprule
\textbf{Method} & \textbf{Attention $\mathcal{S}$} & \textbf{MLP $\mathcal{S}$} \\
\midrule
LoRA \citep{hu2022lora}             & $4.8 \pm 1.1$ & $5.1 \pm 1.0$ \\
SineLoRA \citep{ji2024sine}         & $6.2 \pm 1.4$ & $6.4 \pm 1.3$ \\
RandLoRA \citep{albert2025randlora} & $7.1 \pm 1.6$ & $7.3 \pm 1.5$ \\
\rowcolor{rowhighlight}
\textbf{LR-LoRA (Ours)}             & $\mathbf{9.4 \pm 3.8}$ & $\mathbf{10.7 \pm 4.2}$ \\
\bottomrule
\end{tabular}
\end{adjustbox}
\end{table}

\subsection{MT-Bench Instruction Tuning}\label{app:sec:mtbench}
\Cref{app:tab:mtbench} reports MT-Bench overall scores under the protocol of \cref{subsec:instruction_tuning}. LR-LoRA improves over LoRA by $+0.42$ on \texttt{Qwen2-0.5B} and $+0.43$ on \texttt{LLaMA3-8B}, with consistent per-category gains across all eight MT-Bench dimensions and the largest improvements in the \emph{Reasoning} and \emph{Math} categories.

\begin{table}[h!]
\centering
\caption{\textbf{MT-Bench instruction-tuning results (GPT-4 judge, $r{=}32$).}
Both backbones were fine-tuned on Stanford Alpaca with the protocol described in \cref{subsec:instruction_tuning}; we report MT-Bench overall scores.
LR-LoRA outperforms LoRA, SineLoRA, and RandLoRA on both backbones, with consistent gains across all eight MT-Bench categories (largest improvements observed in \emph{Reasoning} and \emph{Math}).}
\label{app:tab:mtbench}
\vspace{2mm}

\small
\setlength{\tabcolsep}{10pt}
\renewcommand{\arraystretch}{1.15}
\begin{adjustbox}{max width=\linewidth}
\begin{tabular}{lcccc}
\toprule
\textbf{Backbone} & \textbf{LoRA} & \textbf{SineLoRA} & \textbf{RandLoRA} & \textbf{LR-LoRA (Ours)} \\
\midrule
\texttt{Qwen2-0.5B}  & 4.19 & 4.26 & 4.33 & \textbf{4.61} \\
\texttt{LLaMA3-8B}   & 6.12 & 6.22 & 6.31 & \textbf{6.55} \\
\midrule
$\Delta$ over LoRA   & $-$ & $+0.07 / +0.10$ & $+0.14 / +0.19$ & $+0.42 / +0.43$ \\
\bottomrule
\end{tabular}
\end{adjustbox}
\end{table}

\begin{table}[H]
\centering
\scriptsize 
\setlength{\tabcolsep}{2.5pt} 
\renewcommand{\arraystretch}{0.85} 

\caption{\textbf{Per-task commonsense reasoning accuracy (\%).}
Accuracy on eight commonsense tasks (BoolQ, PIQA, SocialIQA, HellaSwag, WinoGrande, ARC-Easy, ARC-Challenge, OpenBookQA) for three backbones (Qwen2-0.5B, Phi-3-8B, LLaMA3-8B) and two data regimes (15k and 170k training examples).
All methods adapt the same module set and use the same rank $r$; LR-LoRA differs from LoRA only by replacing the adapter update $\mathbf{BA}$ with $\phi(\mathbf{BA})$ (\cref{eqn:learnable_nonlin_eqn}).
We report per-task accuracy and the unweighted average across tasks.}
\label{app:tab:commonsense-full}
\vspace{2mm}

\begin{adjustbox}{max width=\linewidth}
\begin{tabular}{lccccccccccc}
\toprule
Method & \% Params & BoolQ & PIQA & SIQA & HellaSwag & WinoGrande & ARC-e & ARC-c & OBQA & Average & $\Delta$ \\ \midrule

\multicolumn{12}{c}{\textbf{Qwen2 - 15k}} \\ \midrule
NoLA & 0.05 & 54.16 & 56.91 & 47.65 & 17.36 & 45.46 & 46.55 & 32.51 & 39.80 & 42.55 & - \\
VeRA1024 & 0.06 & 58.78 & 56.64 & 50.10 & 24.95 & 49.80 & 56.52 & 37.80 & 50.40 & 48.12 & - \\
LoRA-16 & 1.18 & 62.14 & 62.13 & 58.24 & 27.86 & 49.96 & 62.46 & 44.97 & 58.20 & 53.25 & - \\
RandLoRA-10 & 1.18 & 62.14 & 63.49 & 55.32 & 31.16 & 49.96 & 64.27 & 44.97 & 56.60 & 53.49 & - \\
\rowcolor{rowhighlight} \textbf{LR-LoRA-16} & 1.18 & 64.20 & 64.50 & 59.80 & 32.50 & 52.10 & 65.40 & 47.80 & 60.10 & \textbf{55.80} & \textbf{\color{ForestGreen}+2.55} \\ 
LoRA-32 & 2.33 & 59.94 & 62.13 & 56.55 & 30.27 & 41.99 & 64.39 & 46.42 & 57.00 & 52.34 & - \\
RandLoRA-5 & 2.33 & 62.81 & 63.82 & 54.86 & 30.00 & 48.07 & 64.81 & 43.34 & 55.40 & 52.89 & - \\ 
\rowcolor{rowhighlight} \textbf{LR-LoRA-32} & 2.33 & 63.50 & 65.10 & 58.90 & 34.20 & 50.50 & 66.10 & 49.50 & 59.20 & \textbf{55.88} & \textbf{\color{ForestGreen}+3.54} \\ \midrule

\multicolumn{12}{c}{\textbf{Qwen2 - 170k}} \\ \midrule
NoLA & 0.05 & 55.99 & 52.50 & 55.07 & 23.74 & 50.51 & 55.64 & 38.91 & 46.80 & 47.40 & - \\
VeRA1024 & 0.06 & 55.50 & 59.30 & 52.81 & 34.52 & 52.72 & 58.55 & 42.94 & 57.80 & 51.78 & - \\
LoRA-16 & 1.18 & 53.39 & 68.12 & 66.33 & 46.46 & 58.72 & 59.97 & 43.77 & 62.20 & 57.37 & - \\
RandLoRA-10 & 1.18 & 61.47 & 67.63 & 65.61 & 40.26 & 57.22 & 62.12 & 47.95 & 59.60 & 57.73 & - \\
\rowcolor{rowhighlight} \textbf{LR-LoRA-16} & 1.18 & 63.50 & 69.80 & 68.10 & 49.20 & 60.50 & 64.10 & 48.20 & 64.50 & \textbf{60.99} & \textbf{\color{ForestGreen}+3.62} \\
LoRA-32 & 2.33 & 55.78 & 68.28 & 67.20 & 42.37 & 60.22 & 61.03 & 45.05 & 58.80 & 57.34 & - \\
RandLoRA-5 & 2.33 & 63.46 & 65.72 & 66.43 & 42.90 & 56.20 & 61.49 & 47.53 & 59.20 & 57.86 & - \\ 
\rowcolor{rowhighlight} \textbf{LR-LoRA-32} & 2.33 & 65.20 & 69.10 & 69.50 & 46.80 & 62.40 & 63.80 & 50.10 & 61.50 & \textbf{61.05} & \textbf{\color{ForestGreen}+3.71} \\ \midrule

\multicolumn{12}{c}{\textbf{Phi3 - Zero-shot}} \\ \midrule
Zero-shot & 0 & 62.26 & 79.82 & 65.81 & 56.29 & 19.89 & 89.86 & 77.65 & 71.40 & 65.37 & - \\ \midrule
\multicolumn{12}{c}{\textbf{Phi3 - 15k}} \\ \midrule
NoLA & 0.005 & 66.24 & 85.15 & 73.49 & 78.29 & 73.95 & 95.33 & 85.15 & 85.20 & 80.35 & - \\
VeRA1024 & 0.015 & 68.53 & 84.49 & 73.08 & 74.54 & 72.85 & 93.01 & 80.97 & 81.60 & 78.63 & - \\
LoRA-16 & 0.57 & 69.51 & 85.36 & 75.44 & 80.15 & 75.85 & 95.37 & 86.09 & 86.60 & 81.80 & - \\
RandLoRA-40 & 0.58 & 69.54 & 85.31 & 73.80 & 84.05 & 75.14 & 94.65 & 84.90 & 85.80 & 81.65 & - \\
\rowcolor{rowhighlight} \textbf{LR-LoRA-16} & 0.57 & 72.10 & 86.80 & 77.20 & 83.50 & 78.40 & 96.10 & 89.20 & 88.50 & \textbf{83.98} & \textbf{\color{ForestGreen}+2.18} \\
LoRA-32 & 1.14 & 68.44 & 85.31 & 74.67 & 72.14 & 74.98 & 95.20 & 85.41 & 86.60 & 80.34 & - \\
RandLoRA-20 & 1.16 & 69.20 & 85.42 & 75.33 & 83.98 & 75.77 & 95.50 & 85.92 & 87.60 & 82.33 & - \\
\rowcolor{rowhighlight} \textbf{LR-LoRA-32} & 1.14 & 71.50 & 86.50 & 76.90 & 85.10 & 77.80 & 96.20 & 88.10 & 89.10 & \textbf{83.90} & \textbf{\color{ForestGreen}+3.56} \\
LoRA-64 & 2.28 & 69.88 & 85.75 & 74.97 & 74.45 & 75.30 & 95.54 & 87.12 & 88.00 & 81.37 & - \\
RandLoRA-10 & 2.29 & 69.63 & 85.31 & 75.03 & 86.94 & 75.30 & 95.24 & 85.58 & 86.40 & 82.43 & - \\
\rowcolor{rowhighlight} \textbf{LR-LoRA-64} & 2.28 & 71.20 & 86.10 & 76.50 & 88.20 & 76.80 & 95.80 & 87.90 & 89.50 & \textbf{84.00} & \textbf{\color{ForestGreen}+2.63} \\ \midrule

\multicolumn{12}{c}{\textbf{Phi3 - 170k}} \\ \midrule
NoLA & 0.005 & 68.87 & 85.15 & 77.18 & 85.13 & 77.90 & 95.20 & 85.58 & 83.60 & 82.33 & - \\
VeRA1024 & 0.015 & 69.53 & 84.53 & 74.52 & 84.08 & 76.82 & 94.51 & 83.68 & 83.54 & 81.40 & - \\
LoRA-16 & 0.57 & 70.83 & 84.39 & 78.45 & 89.94 & 82.87 & 95.45 & 86.09 & 89.00 & 84.63 & - \\
RandLoRA-40 & 0.58 & 70.86 & 86.67 & 78.81 & 90.07 & 82.00 & 95.12 & 86.26 & 87.60 & 84.67 & - \\
\rowcolor{rowhighlight} \textbf{LR-LoRA-16} & 0.57 & 73.50 & 86.90 & 80.50 & 92.40 & 85.10 & 96.20 & 89.50 & 91.20 & \textbf{86.91} & \textbf{\color{ForestGreen}+2.28} \\
LoRA-32 & 1.14 & 71.23 & 85.96 & 78.92 & 91.77 & 82.95 & 94.61 & 84.81 & 89.40 & 84.96 & - \\
RandLoRA-20 & 1.16 & 71.62 & 87.43 & 79.48 & 91.48 & 82.79 & 95.16 & 86.01 & 87.80 & 85.22 & - \\
\rowcolor{rowhighlight} \textbf{LR-LoRA-32} & 1.14 & 74.10 & 87.80 & 81.20 & 93.50 & 85.60 & 96.50 & 88.20 & 91.50 & \textbf{87.30} & \textbf{\color{ForestGreen}+2.34} \\
LoRA-64 & 2.28 & 71.93 & 86.13 & 79.58 & 90.14 & 83.74 & 92.68 & 81.74 & 87.80 & 84.22 & - \\
RandLoRA-10 & 2.29 & 71.87 & 86.56 & 79.43 & 90.99 & 82.72 & 95.66 & 85.49 & 87.40 & 85.01 & - \\
\rowcolor{rowhighlight} \textbf{LR-LoRA-64} & 2.29 & 73.80 & 87.20 & 80.90 & 91.80 & 84.50 & 93.10 & 83.50 & 89.40 & \textbf{85.52} & \textbf{\color{ForestGreen}+1.30} \\ \midrule

\multicolumn{12}{c}{\textbf{LLaMA3 - Zero-shot}} \\ \midrule
Zero-shot & 0 & 60.73 & 41.40 & 28.40 & 25.00 & 10.97 & 16.41 & 15.96 & 16.80 & 26.96 & - \\ \midrule
\multicolumn{12}{c}{\textbf{LLaMA3 - 15k}} \\ \midrule
NoLA & 0.004 & 67.58 & 84.49 & 72.31 & 69.60 & 70.56 & 90.49 & 78.75 & 81.20 & 76.87 & - \\
VeRA1024 & 0.014 & 63.36 & 84.39 & 74.10 & 77.70 & 71.35 & 89.48 & 76.54 & 80.20 & 77.14 & - \\
LoRA-16 & 0.35 & 73.03 & 86.94 & 75.90 & 90.53 & 77.74 & 90.74 & 80.29 & 86.20 & 82.67 & - \\
RandLoRA-60 & 0.36 & 71.19 & 84.22 & 75.59 & 83.82 & 74.98 & 91.12 & 81.31 & 86.00 & 81.03 & - \\
\rowcolor{rowhighlight} \textbf{LR-LoRA-16} & 0.35 & 75.50 & 88.10 & 78.20 & 93.10 & 80.50 & 92.20 & 84.50 & 89.10 & \textbf{85.15} & \textbf{\color{ForestGreen}+2.48} \\
LoRA-32 & 0.7 & 74.22 & 86.40 & 75.79 & 91.90 & 77.35 & 90.61 & 80.80 & 87.60 & 83.09 & - \\
RandLoRA-30 & 0.7 & 71.65 & 83.79 & 74.56 & 86.85 & 75.61 & 90.78 & 80.03 & 87.20 & 81.31 & - \\
\rowcolor{rowhighlight} \textbf{LR-LoRA-32} & 0.7 & 76.80 & 88.50 & 79.10 & 94.20 & 81.20 & 92.50 & 85.10 & 90.50 & \textbf{85.99} & \textbf{\color{ForestGreen}+2.90} \\
LoRA-64 & 1.4 & 71.77 & 84.17 & 76.25 & 85.14 & 73.80 & 91.46 & 80.80 & 86.20 & 81.20 & - \\
RandLoRA-15 & 1.4 & 70.98 & 86.02 & 75.44 & 89.74 & 76.80 & 91.29 & 81.66 & 83.80 & 81.96 & - \\
\rowcolor{rowhighlight} \textbf{LR-LoRA-64} & 1.4 & 73.10 & 87.20 & 78.50 & 90.80 & 79.50 & 91.80 & 82.50 & 87.50 & \textbf{83.86} & \textbf{\color{ForestGreen}+2.66} \\ \midrule

\multicolumn{12}{c}{\textbf{LLaMA3 - 170k}} \\ \midrule
NoLA & 0.004 & 71.83 & 84.66 & 77.79 & 85.05 & 82.72 & 88.59 & 76.45 & 82.20 & 81.16 & - \\
VeRA1024 & 0.014 & 70.55 & 85.69 & 79.27 & 92.14 & 82.64 & 87.33 & 73.38 & 82.20 & 81.65 & - \\
LoRA-16 & 0.35 & 75.14 & 89.12 & 80.66 & 89.01 & 86.58 & 90.07 & 78.75 & 86.20 & 84.44 & - \\
RandLoRA-60 & 0.35 & 75.26 & 87.98 & 79.63 & 94.66 & 85.64 & 90.03 & 79.44 & 84.40 & 84.62 & - \\
\rowcolor{rowhighlight} \textbf{LR-LoRA-16} & 0.35 & 77.20 & 90.50 & 83.10 & 93.80 & 89.20 & 92.10 & 82.50 & 89.50 & \textbf{87.24} & \textbf{\color{ForestGreen}+2.80} \\
LoRA-32 & 0.7 & 75.08 & 88.85 & 80.25 & 95.42 & 86.19 & 90.28 & 80.29 & 85.60 & 85.24 & - \\
RandLoRA-30 & 0.7 & 76.33 & 88.08 & 80.25 & 95.67 & 86.11 & 90.36 & 80.89 & 87.00 & 85.59 & - \\
\rowcolor{rowhighlight} \textbf{LR-LoRA-32} & 0.7 & 78.40 & 90.80 & 82.90 & 97.10 & 89.50 & 92.80 & 84.10 & 90.20 & \textbf{88.22} & \textbf{\color{ForestGreen}+2.98} \\
LoRA-64 & 1.4 & 74.65 & 89.66 & 80.86 & 95.17 & 86.74 & 90.95 & 79.18 & 85.40 & 85.33 & - \\
RandLoRA-15 & 1.4 & 72.63 & 87.98 & 81.37 & 95.68 & 87.77 & 91.33 & 80.89 & 89.00 & 85.83 & - \\
\rowcolor{rowhighlight} \textbf{LR-LoRA-64} & 1.4 & 76.50 & 89.90 & 82.10 & 96.50 & 88.20 & 91.80 & 81.50 & 89.80 & \textbf{87.04} & \textbf{\color{ForestGreen}+1.71} \\ \bottomrule
\end{tabular}
\end{adjustbox}
\end{table}

\section{Computational Complexity Analysis}
\label{app:sec:computational_complexity}

This appendix provides the full derivation underlying the per-step overhead claim in \cref{subsec:training_efficiency}.

\textbf{Implementation: merge once per step.}
Per training step, we form
\[
W_{\mathrm{eff}} \;=\; W \;+\; \phi(\mathbf{BA}) \;\in\; \mathbb{R}^{m \times n}
\]
once, then run the standard forward pass $y = W_{\mathrm{eff}}\, X$ as a single backbone-shaped matmul. The element-wise nonlinearity $\phi$ therefore acts on the $m{\times}n$ matrix $\mathbf{BA}$, not on the $b\cdot s$-scaled activations. This is the key implementation detail that bounds the overhead. The $0.05$ dropout reported in \cref{subsec:exp_setup} is applied to the inputs $X$ on the LR-LoRA pathway prior to merging (input-side dropout), preserving the merge trick; this matches the standard LoRA-family implementation when adapters are merged for efficient training.

\textbf{Per-step adapter cost.}
The adapter overhead decomposes as $O(mnr)$ for $\mathbf{BA}$, $O(mnN)$ for evaluating $\phi$ elementwise with $N$ basis functions, and $O(mn)$ for the merge. Total $O(mn(r{+}N))$ per step, independent of batch size $b$ and sequence length $s$. The forward matmul $W_{\mathrm{eff}}X$ costs $O(mn\cdot b\cdot s)$, identical to a backbone matmul.

\textbf{Fractional overhead.}
The overhead ratio (adapter cost over per-step backbone cost) is
\[
\frac{r + N}{b \cdot s}.
\]
For our experimental settings ($s{=}512$, $r{=}32$, $N{=}50$):

\begin{center}
\begin{adjustbox}{max width=\linewidth}
\begin{tabular}{lcccc}
\toprule
\textbf{Backbone} & $b$ & $b{\cdot}s$ & $(r{+}N)/(b{\cdot}s)$ & \textbf{Measured throughput} $\Delta$ \\
\midrule
\texttt{Qwen2-0.5B}  & 16 & 8192 & 1.00\% & $-0.5\%$ \\
\texttt{Phi-3-8B}    &  8 & 4096 & 2.00\% & $-0.7\%$ \\
\texttt{LLaMA3-8B}   &  4 & 2048 & 4.00\% & $-1.0\%$ \\
\texttt{LLaMA3-13B}  &  4 & 2048 & 4.00\% & $-1.1\%$ \\
\bottomrule
\end{tabular}
\end{adjustbox}
\end{center}

The closed-form bound matches measured throughput within the natural range; measurements include backbone-amortized effects (attention, optimizer state, data movement) that further dilute the per-projection bound.

\textbf{Backpropagation cost.}
Standard LoRA exploits matrix associativity in the backward pass to avoid forming a dense $m{\times}n$ weight gradient: $\nabla_\mathbf{A} L = \mathbf{B}^\top(\nabla_{\mathbf{Y}}L\,\mathbf{X}^\top)$ at $O(r(m{+}n)bs)$. Applying an elementwise nonlinearity $\phi$ to $\mathbf{BA}$ breaks this associativity, because the gradient with respect to the adapter output $\mathbf{Z}=\mathbf{BA}$ is $\nabla_{\mathbf{Z}}L = (\nabla_{\mathbf{Y}}L)\mathbf{X}^\top \odot \phi'(\mathbf{Z})$, which requires materializing $(\nabla_{\mathbf{Y}}L)\mathbf{X}^\top \in \mathbb{R}^{m\times n}$ once per adapted layer at $O(mn\cdot b\cdot s)$. Once $\nabla_{\mathbf{Z}}L$ is formed, the chain rule recovers $\nabla_\mathbf{A} L = \mathbf{B}^\top \nabla_{\mathbf{Z}}L \in \mathbb{R}^{r\times n}$ and $\nabla_\mathbf{B} L = \nabla_{\mathbf{Z}}L\,\mathbf{A}^\top \in \mathbb{R}^{m\times r}$ at $O(mnr)$, and $\phi'(\mathbf{Z})$ contributes $O(mnN)$. The dominant added cost relative to LoRA's backward is therefore the $O(mn\cdot b\cdot s)$ dense-gradient term, which has the same scaling as a single backbone matmul at that projection, so the relative backward overhead is bounded by a small constant factor on the per-projection backbone backward.

\textbf{Inference cost.}
At inference we precompute $\phi(\mathbf{BA})$ once and permanently merge it into $W$, so the deployed model has identical compute and memory cost to the backbone. There is zero inference-time overhead.

\textbf{Memory overhead.}
LR-LoRA introduces $2N|\mathcal{L}|$ additional learnable parameters for the transfer functions, where $|\mathcal{L}|$ is the number of adapted modules. As shown in \cref{app:tab:param_counts}, this is $<\!1\%$ of LoRA adapter parameters across all architectures. Peak training memory grows by at most $+3\%$. The $m\times n$ matrix $\phi(\mathbf{BA})$ is materialized transiently per layer during the forward step (it is freed back to the activation pool after the merge with $W$ completes for that layer), so only one such matrix is alive at any time rather than one per adapted layer in parallel; under standard gradient checkpointing this transient further compresses. The $2N|\mathcal{L}|$ persistent transfer-function parameters themselves occupy $<\!20$\,kB versus $\sim\!240$\,MB of LoRA adapter memory on \texttt{Qwen2-0.5B}.

\begin{table}[H]
\centering
\caption{\textbf{Parameter overhead summary.} Transfer function parameter overhead relative to base LoRA parameters across architectures, demonstrating minimal computational cost for learned rank adaptation.}
\label{app:tab:param_counts}
\small
\begin{adjustbox}{max width=\linewidth}
\begin{tabular}{lcccc}
\toprule
\textbf{Architecture} & \textbf{LoRA Params} & \textbf{Transfer Params} & \textbf{Total} & \textbf{Overhead} \\
\midrule
Qwen2-0.5B & 1.77M & 16.8k & 1.79M & +0.95\% \\
Phi-3-8B & 14.7M & 22.4k & 14.72M & +0.15\% \\
LLaMA3-8B & 17.2M & 22.4k & 17.22M & +0.13\% \\
RoBERTa-base & 1.03M & 7.2k & 1.04M & +0.70\% \\
\bottomrule
\end{tabular}
\end{adjustbox}
\end{table}

\textbf{Training throughput.} Despite the $O(N)$ computational overhead, empirical measurements in \cref{app:tab:efficiency} show that the peak memory increases by at most \textbf{3\%} and the throughput decreases by at most \textbf{2\%} relative to LoRA under the matched protocol.

\section{Vision Transfer Evaluation}
\label{app:sec:vision_eval}

\paragraph{Vision evaluation protocol.}
We report the exact image preprocessing, resolution, augmentations, and evaluation procedure for each dataset/backbone combination. For CLIP ViT-B/32: ImageNet-1k uses 224×224 resolution with center crop, CIFAR-100 uses 224×224 with resize, all datasets use ImageNet normalization constants [(0.485, 0.456, 0.406), (0.229, 0.224, 0.225)]. Training augmentations include RandomResizedCrop and RandomHorizontalFlip. For DINOv2 ViT-B/14: similar preprocessing with 224×224 resolution matching the original training protocol. All evaluations use single crop and report top-1 accuracy. Template choices for CLIP follow the original paper's prompt ensembling strategy.

\subsection{CLIP Vision Transfer}
\label{app:sec:clip_eval}

To demonstrate the generalizability of LR-LoRA beyond language models, we evaluate it on CLIP vision encoders following the protocol established by \citet{albert2025randlora}. The key question is whether the learned rank adaptation benefits transfer to visual representation learning, where different layers may require distinct adaptation complexities for feature extraction versus classification.

\textbf{Experimental setup.} We fine-tuned CLIP ViT-B/32 with rank $r=16$ adapters targeting attention QKV projections and MLP layers. We evaluated four datasets: ImageNet-1k (10 epochs), CIFAR-100 (6 epochs), Flowers102 (40 epochs), and SUN397 (14 epochs). Training uses AdamW with a weight decay of 0.1, learning rate $10^{-2}$, batch size of 128, and cosine learning rate schedule. All methods used identical data preprocessing and evaluation protocols to ensure a fair comparison.

\textbf{Results and analysis.} \Cref{app:tab:clip_results} shows that LR-LoRA achieves consistent improvements over LoRA baselines on visual tasks, with gains of $+1.0$ to $+1.3$ points across the four datasets (average $+1.2$). LR-LoRA also outperforms VeRA and RandLoRA on every dataset, indicating that the layer-wise rank-adaptation principle generalizes beyond language modeling to visual representation learning even at modest parameter budgets.

\begin{table}[H]
\centering
\caption{\textbf{CLIP image classification transfer results (top-1 accuracy \%) on \texttt{CLIP ViT-B/32}.}
Adapters are inserted into attention and MLP projections with rank $r{=}16$; results are averaged over three seeds.
Datasets are standard transfer benchmarks; higher is better.
Takeaway: LR-LoRA provides consistent gains over parameter-matched baselines across datasets.}
\label{app:tab:clip_results}
\vspace{2mm}
\scriptsize
\setlength{\tabcolsep}{5pt}
\renewcommand{\arraystretch}{1.1}
\begin{adjustbox}{max width=\textwidth}
\begin{tabular}{lcccc}
\toprule
Dataset & LoRA & VeRA & RandLoRA & \textbf{LR-LoRA} \\
\midrule
ImageNet-1k & 78.4 & 78.1 & 78.9 & \textbf{79.6} \\
CIFAR-100 & 84.2 & 83.9 & 84.6 & \textbf{85.4} \\
Flowers102 & 95.1 & 95.0 & 95.4 & \textbf{96.1} \\
SUN397 & 66.8 & 66.5 & 67.2 & \textbf{68.1} \\
\midrule
Average & 81.1 & 80.9 & 81.5 & \textbf{82.3} \\
\bottomrule
\end{tabular}
\end{adjustbox}
\end{table}

\subsection{DINOv2 Vision Transfer}
\label{app:sec:dinov2_eval}
We assessed the performance of LR-LoRA on DINOv2 vision encoders to examine its generalizability beyond contrastive learning frameworks. DINOv2 utilizes self-supervised learning without contrastive objectives, offering a crucial evaluation of whether the benefits of learned rank adaptation extend across diverse pre-training paradigms and architectural configurations.

\textbf{Experimental protocol.} We fine-tuned the DINOv2 ViT-B/14 model with rank $r=16$ adapters, focusing on attention and MLP projections. The evaluation encompassed ImageNet-1k, CIFAR-100, Oxford-IIIT Pets, and SUN397, using top-1 accuracy as the metric. The training process employed AdamW with a weight decay of 0.1, a cosine learning rate schedule, and a batch size of 128. All methods adhered to identical protocols to ensure a fair comparison. For pure-vision fine-tuning, LoRA-family methods utilize a learning rate of $10^{-2}$ (RandLoRA C.3); LR-LoRA follows the same schedule as that of LoRA. We adhered to the dataset-specific epoch counts from RandLoRA (Table 7): ImageNet-1k (10), CIFAR-100 (6), Oxford-IIIT Pets (5), and SUN397 (14). For the RandLoRA baseline, we employed the ViT-B/14 configuration (basis rank 8, number of bases 128), aligning with RandLoRA settings.

\textbf{Results and analysis.} \Cref{app:tab:dinov2_results} demonstrates consistent improvements for LR-LoRA over LoRA-family baselines across datasets. These findings indicate that the advantages of spectral control are not confined to the language models.

\begin{table}[H]
\centering
\caption{\textbf{DINOv2 image classification transfer results (top-1 accuracy \%) on \texttt{DINOv2 ViT-B/14}.}
Adapters were inserted into the attention and MLP projections with rank $r{=}16$; results were averaged over three seeds.
Datasets are standard transfer benchmarks; the higher, the better.
Takeaway: LR-LoRA improves over strong baselines under matched parameter budgets.}
\label{app:tab:dinov2_results}
\vspace{2mm}
\scriptsize
\setlength{\tabcolsep}{5pt}
\renewcommand{\arraystretch}{1.1}
\begin{adjustbox}{max width=\textwidth}
\begin{tabular}{lcccc}
\toprule
Dataset & LoRA & VeRA & RandLoRA & \textbf{LR-LoRA} \\
\midrule
ImageNet-1k & 80.2 & 79.8 & 80.6 & \textbf{81.3} \\
CIFAR-100 & 86.5 & 86.2 & 86.9 & \textbf{87.6} \\
Oxford-IIIT Pets & 92.4 & 92.1 & 92.8 & \textbf{93.5} \\
SUN397 & 68.9 & 68.5 & 69.3 & \textbf{70.2} \\
\midrule
Average & 82.0 & 81.7 & 82.4 & \textbf{83.2} \\
\bottomrule
\end{tabular}
\end{adjustbox}
\end{table}

\section{Implementation Details}
\label{app:sec:appendix_implementation}

\begin{figure}[H]
\centering
\includegraphics[width=\linewidth]{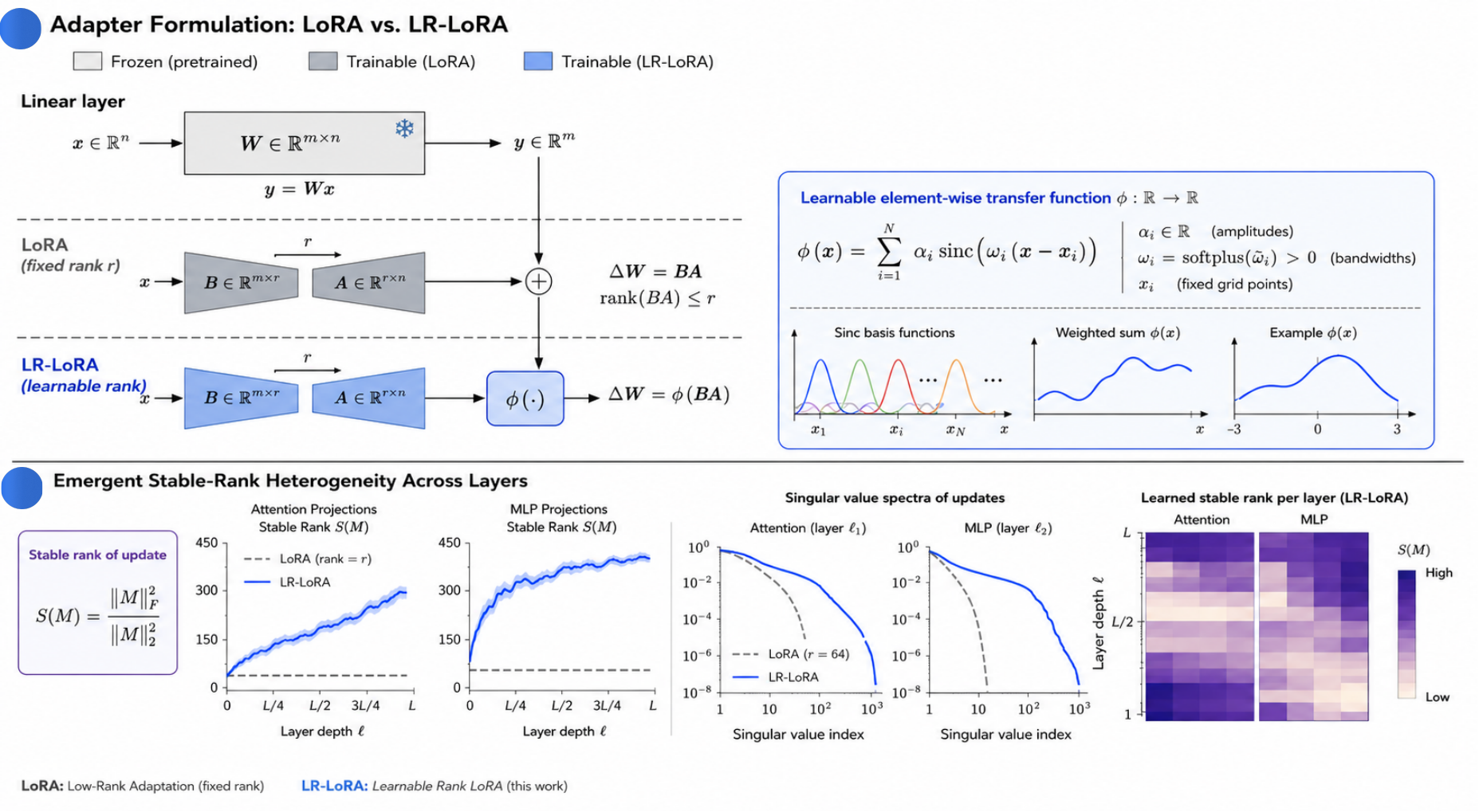}
\caption{\textbf{LR-LoRA pipeline.} End-to-end view of an LR-LoRA-adapted layer: the frozen pretrained weight $\mathbf{W}$, the trainable low-rank factors $\mathbf{A},\mathbf{B}$, and the elementwise sinc transfer function $\phi$ (\cref{eqn:learnable_nonlin_eqn}) with learnable amplitudes $\{\alpha_i\}$ and bandwidths $\{\omega_i\}$ on a fixed grid $\{x_i\}_{i=1}^N$. The adapted weight is $W_{\mathrm{eff}}=\mathbf{W}+\phi(\mathbf{BA})$, computed once per training step and merged for zero-overhead inference (\cref{subsec:training_efficiency,app:sec:computational_complexity}).}
\label{app:fig:pipeline}
\end{figure}

\subsection{Complete Hyperparameter Configuration}
\label{app:sec:hparam_table}

This section provides a comprehensive summary of all the hyperparameters employed in our experiments. Unless explicitly indicated otherwise (e.g., in \cref{subsec:ablations,sec:experiments}), a single default configuration was utilized across tasks and backbones. Any benchmark-specific deviations, such as batch size, precision, and learning rate schedule, are detailed in \cref{app:tab:hyperparams_complete}. For settings that undergo tuning, such as the GLUE learning rate, the exact tuning grid is specified in the main paper and is consistent across methods. All LR-LoRA-specific selections, including grid size $N$, interval $I$, and initialization via $\omega=\mathrm{softplus}(\tilde\omega)$, adhere to \cref{sec:learnable_nonlinearity}.

\begin{table}[ht]
\centering
\caption{\textbf{Hyperparameters used across experiments.}
Unless explicitly varied in the ablations, the LR-LoRA runs used the settings below.
The values are reported separately for decoder-only LLM fine-tuning (commonsense suite), GLUE (RoBERTa), and vision models.
LR-LoRA uses the $\mathrm{sinc}$-basis nonlinearity $\phi$ with grid size $N$ on $[-I,I]$ and bandwidth parameterization $\omega=\mathrm{softplus}(\tilde\omega)$ (\cref{eqn:learnable_nonlin_eqn,sec:learnable_nonlinearity}).}
\label{app:tab:hyperparams_complete}
\vspace{1mm}

\scriptsize
\setlength{\tabcolsep}{4pt}
\renewcommand{\arraystretch}{1.15}
\begin{adjustbox}{max width=\linewidth}
\begin{tabular}{llccc}
\toprule
\textbf{Component} & \textbf{Hyperparameter} & \textbf{LLMs} & \textbf{GLUE} & \textbf{Vision} \\
\midrule
\textbf{Adapter} 
& Placement & Attn: q/k/v/o;\; MLP: gate/up/down & fixed module set & fixed module set \\
& Dropout & 0.05 & 0.10 & 0.00 \\
\midrule
\textbf{LR-LoRA ($\phi$)} 
& Grid points $N$ & 50 & 50 & 50 \\
& Interval $I$ & 3 & 3 & 3 \\
& Amplitude init $\alpha_i$ & 0 & 0 & 0 \\
& Bandwidth init $\omega_0$ & 1.0 & 1.0 & 1.0 \\
& Bandwidth parameterization & $\omega=\mathrm{softplus}(\tilde\omega)$ & $\omega=\mathrm{softplus}(\tilde\omega)$ & $\omega=\mathrm{softplus}(\tilde\omega)$ \\
\midrule
\textbf{Optimizer}
& Type & AdamW & AdamW & AdamW \\
& $(\beta_1,\beta_2)$ & (0.9, 0.999) & (0.9, 0.999) & (0.9, 0.999) \\
& Weight decay & 0.00 & 0.01 & 0.10 \\
& Gradient clipping & 1.0 & 1.0 & 1.0 \\
\midrule
\textbf{Learning rate}
& Base LR & $1\times 10^{-4}$ & grid in \cref{subsec:exp_setup} & $1\times 10^{-2}$ \\
& Schedule & fixed & linear warmup + linear decay & cosine \\
& Warmup ratio & 0.10 & 0.06 & 0.10 \\
\midrule
\textbf{Training}
& Epochs & 3 & reported with results & reported with results \\
& Batch size (small) & 16 (0.5B) & 32 & 128 \\
& Batch size (large) & 4 (8B) & 16 & 64 \\
& Precision & bf16 & fp16 & fp16 \\
\midrule
\textbf{Early stopping}
& Patience & 2 epochs & 3 epochs & 5 epochs \\
\bottomrule
\end{tabular}
\end{adjustbox}

\vspace{1mm}
\begin{flushleft}
\scriptsize
\end{flushleft}
\end{table}

\paragraph{Random seeds.}
We run all experiments with the following seeds: \{42, 123, 456\}.

\paragraph{Training schedule.}
We train for 3 epochs, corresponding to approximately 3.2k total optimizer steps for Qwen2-0.5B, 6.4k steps for Phi-3-8B, and 12.8k steps for LLaMA3-8B (after accounting for gradient accumulation). We evaluate every 500 steps and select the final checkpoint by best validation performance.

\paragraph{Code and evaluation.}
We use HuggingFace transformers v4.36 and follow the evaluation scripts from \citet{albert2025randlora}, maintaining identical preprocessing and metric computation across all methods.

\paragraph{Adapter placement.}
We insert LR-LoRA modules into the following weight matrices for each backbone: attention \{query, key, value, output\} and MLP \{gate, up, down\} projections. We keep this placement fixed across all baselines for controlled comparison.

\paragraph{Gradient flow at initialization.}
Although $\phi \equiv 0$ at initialization when $\alpha_i = 0$, the amplitudes receive nonzero gradients in the first optimization step because $\phi$ is linear in $\alpha$.
Specifically, with
\[
\phi(x) \;=\; \sum_{i=1}^{N} \alpha_i\,\mathrm{sinc}\!\big(\omega_i(x-x_i)\big),
\]
we have
\[
\frac{\partial \phi(x)}{\partial \alpha_i} \;=\; \mathrm{sinc}\!\big(\omega_i(x-x_i)\big),
\]
which is generally nonzero for a typical $x$.
Therefore, $\alpha$ immediately departs from zero; once $\alpha \neq 0$, $\partial \phi(x)/\partial x$ is nonzero and gradients propagate to $A,B$ through $x=(BA)_{ij}$.
Furthermore, the grid points $\{x_i\}_{i=1}^{N}$ occupy distinct positions on $[-I,I]$, so even at the all-zero state $\mathbf{Z}=\mathbf{BA}=\mathbf{0}$ the per-grid sensitivity $\partial \phi(0)/\partial \alpha_i = \mathrm{sinc}(-\omega_i x_i)$ varies with $i$; the amplitudes therefore depart from zero \emph{heterogeneously} (different $\alpha_i$ receive different first-step updates), breaking any putative symmetry across grid points before $\phi'$ is queried.
In practice, we observe $\alpha$ becomes nonzero within the first few optimizer steps.

\subsection{Detailed Parameter Breakdown}
\label{app:sec:detailed_params}

\cref{app:tab:param_counts_detailed} enumerates trainable parameters for LR-LoRA across backbones, decomposed into attention and MLP components.
We apply adapters only to transformer-block projection matrices and exclude the output head (classification/LM head); therefore, the head does not contribute to the adapted-module set $|\mathcal{L}|$ or any parameter counts reported in this section.
Across architectures, MLP adapters constitute the majority of LoRA parameters owing to their larger projection dimensions, whereas attention parameters scale with the four standard attention projections (query, key, value, and output).

\begin{table}[ht]
\centering
\caption{\textbf{Detailed parameter breakdown for LR-LoRA by component and architecture.}
LoRA parameters correspond to $\mathbf{A},\mathbf{B}$ in \cref{eqn:lora_eqn}.
LR-LoRA adds transfer-function parameters for $\phi$ in \cref{eqn:learnable_nonlin_eqn}; each instantiated $\phi$ contributes $2N$ scalars $\{(\omega_i,\alpha_i)\}_{i=1}^N$.
Transfer counts below use $N=50$ and a separate $\phi$ per adapted module per layer, excluding the output head.}
\label{app:tab:param_counts_detailed}
\vspace{1mm}

\scriptsize
\setlength{\tabcolsep}{4pt}
\renewcommand{\arraystretch}{1.12}
\begin{adjustbox}{max width=\linewidth}
\begin{tabular}{llcccc}
\toprule
\textbf{Architecture} & \textbf{Component} & \textbf{LoRA Params} & \textbf{Transfer Params} & \textbf{Total} & \textbf{Overhead} \\
\midrule
\multirow{3}{*}{\texttt{Qwen2-0.5B}} 
& Attention ($24L\times 4$) & 393.2k & 9.6k  & 402.8k & +2.4\% \\
& MLP ($24L\times 3$)       & 1.38M  & 7.2k  & 1.3872M & +0.5\% \\
\midrule
\multirow{3}{*}{\texttt{Phi-3-8B}} 
& Attention ($32L\times 4$) & 3.15M & 12.8k & 3.1628M & +0.41\% \\
& MLP ($32L\times 3$)       & 11.5M & 9.6k  & 11.5096M & +0.08\% \\
\midrule
\multirow{3}{*}{\texttt{LLaMA3-8B}} 
& Attention ($32L\times 4$) & 4.19M & 12.8k & 4.2028M & +0.31\% \\
& MLP ($32L\times 3$)       & 13.0M & 9.6k  & 13.0096M & +0.07\% \\
\bottomrule
\end{tabular}
\end{adjustbox}
\end{table}

The transfer-function parameter count follows $\Delta P = 2N|\mathcal{L}|$ with $N=50$ and $|\mathcal{L}|$ being the set of adapted modules.
Per transformer layer, $|\mathcal{L}_{\text{attn}}|=4$ (query, key, value, output) and $|\mathcal{L}_{\text{MLP}}|=3$ (gate, up, down), and the output head is excluded.
Thus, the per-layer transfer function overhead is $2N(4+3)=700$ parameters (400 from attention and 300 from MLP), yielding $700L$ transfer parameters for an $L$-layer backbone.

\section{Ablation Studies}
\label{app:sec:ablations}

This section elaborates on the ablation narrative concerning the LR-LoRA update family $\phi(\mathbf{BA})$ as described in \cref{eqn:learnable_nonlin_eqn}, where the update is applied element-wise to the low-rank factorization in \cref{eqn:lora_eqn}. Each study isolates a single design choice while maintaining a constant backbone, adapter placement, rank $r$, data pipeline, and an optimization schedule. The objective is to discern which components of the sinc-parameterized transfer function (\cref{eqn:shifted_sinc}) are essential for consistent improvements and which serve primarily as refinements. The results are categorized by (i) basis construction and grid design, (ii) parameterization and initialization, and (iii) architectural integration and sharing. All results are presented as task-average accuracy according to the common ablation protocol; tables in this section share a single-seed ablation harness whose absolute accuracy scale differs from the multi-seed full-suite averages in \cref{tab:aggregated_results}, so each ablation is interpreted relative to its matched-protocol LoRA baseline to highlight sensitivity rather than absolute performance.

\subsection{Extended Basis Function Ablations}
\label{app:sec:basis_ablations}

The basis ablations investigate the representational capacity of the transfer function $\phi$ as described in \cref{eqn:learnable_nonlin_eqn}. Given that $\phi$ is a finite expansion of shifted sinc functions (\cref{eqn:shifted_sinc}), its behavior is contingent on both the number of basis elements and grid geometry. These studies aim to determine whether the observed performance improvements are attributable to the inherent form of the basis or the flexibility afforded by the size and placement of the basis.

\subsubsection{Base Rank Interaction}
\label{app:sec:rank_interaction_ablation}

We initially investigate the interaction between LR-LoRA and the base low-rank budget $r$ as presented in \cref{eqn:lora_eqn}. The ablation study detailed in \cref{app:tab:ablation_rank} varies $r$ while maintaining a constant transfer function parameterization, enabling the evaluation of whether LR-LoRA merely compensates for low rank or offers advantages beyond increasing $r$. The findings indicate the most significant relative improvements at smaller ranks, where the low-rank constraint is the most restrictive. As $r$ increases, the disparity between LR-LoRA and LoRA diminishes, which is consistent with the hypothesis that expanded low-rank capacity partially alleviates the limitations of fixed-rank updates. This evidence supports the central assertion that the learned transfer function is particularly beneficial when the capacity is constrained.

\begin{table}[H]
\centering
\caption{\textbf{Interaction with base rank $r$.} Performance of LR-LoRA vs LoRA across different base ranks on \texttt{Qwen2-0.5B} commonsense reasoning (170k regime). This is a focused single-seed ablation on a fixed task subset; the absolute numbers therefore differ slightly from the multi-seed average in \cref{tab:aggregated_results}, but the rank-dependent gain pattern is the same. LR-LoRA gains are largest at small ranks where capacity is most constrained.}
\label{app:tab:ablation_rank}
\vspace{2mm}
\small
\setlength{\tabcolsep}{8pt}
\renewcommand{\arraystretch}{1.1}
\begin{tabular}{ccccc}
\toprule
\textbf{Base Rank $r$} & \textbf{LoRA} & \textbf{LR-LoRA} & \textbf{Absolute Gain} & \textbf{Relative Gain} \\
\midrule
4 & 52.8 ± 0.4 & 58.1 ± 0.3 & +5.3\% & +10.0\% \\
8 & 55.6 ± 0.3 & 59.8 ± 0.2 & +4.2\% & +7.6\% \\
16 & 57.3 ± 0.2 & 61.1 ± 0.2 & +3.8\% & +6.6\% \\
32 & 58.9 ± 0.2 & 62.1 ± 0.2 & +3.2\% & +5.4\% \\
64 & 60.1 ± 0.2 & 62.8 ± 0.3 & +2.7\% & +4.5\% \\
\bottomrule
\end{tabular}
\end{table}

\subsubsection{Alternative Basis Functions}
\label{app:sec:basis_comparison_detailed}

We conducted a comparative analysis of the sinc basis against alternative basis families while maintaining a constant parameter count. This approach directly evaluates whether the observed improvements are attributable to the specific structure of \cref{eqn:shifted_sinc}, or merely the introduction of a nonlinear basis. As reported in \Cref{app:tab:ablation_basis}, the sinc basis consistently outperformed the other candidates. This advantage aligns with the sampling-theoretic rationale presented in \cref{eqn:shifted_sinc}, as sinc functions offer stable constraints on the bandwidth and interpolation behavior, which are essential for smooth and controllable update transformations. Notably, the ablation study indicates that the inductive bias of the basis is significant, rather than solely the presence of nonlinear mapping.
\begin{table}[H]
\centering
\caption{\textbf{Alternative basis function comparison.} Performance across different basis functions with equal parameter counts on \texttt{Qwen2-0.5B} commonsense reasoning.
Sinc basis achieves optimal performance due to its bandlimited properties and theoretical grounding.}
\label{app:tab:ablation_basis}
\vspace{2mm}
\small
\setlength{\tabcolsep}{6pt}
\renewcommand{\arraystretch}{1.1}
\resizebox{0.7\columnwidth}{!}{
\begin{tabular}{lcccc}
\toprule
\textbf{Basis Type} & \textbf{Function Form} & \textbf{Accuracy} & \textbf{Stability} & \textbf{Convergence} \\
\midrule
Sinc & $\mathrm{sinc}(\omega_i(x-x_i))$ & \textbf{61.1 ± 0.2} & High & Fast \\
Gaussian RBF & $\exp(-(\omega_i(x-x_i))^2)$ & 60.4 ± 0.3 & Medium & Medium \\
B-splines & $B_3(\omega_i(x-x_i))$ & 59.8 ± 0.4 & Medium & Fast \\
Fourier & $\cos(\omega_i x + \phi_i)$ & 59.2 ± 0.5 & Low & Slow \\
Polynomial & $(1 + \omega_i x)^3$ & 58.1 ± 0.6 & Low & Very Slow \\
\bottomrule
\end{tabular}
}
\end{table}

\subsubsection{Grid Placement Strategies}
\label{app:sec:grid_strategies}

Given that the sinc basis in \cref{eqn:shifted_sinc} is implemented on a grid, the positioning of the grid points $\{x_i\}$ determines the density of the representation across various activation ranges. \Cref{app:tab:ablation_grid} presents a comparison between uniform and nonuniform grids under identical training conditions. Uniform grids exhibited the greatest stability, whereas Chebyshev grids demonstrated a slight but consistent improvement in performance. This suggests that a modest concentration of basis elements near the boundaries can mitigate the interpolation artifacts. Consequently, although LR-LoRA does not exhibit significant sensitivity to grid placement, a carefully considered design can yield incremental advantages.

\begin{table}[H]
\centering
\caption{\textbf{Grid placement strategy comparison.} Analysis of different grid point arrangements with $N=32$ on \texttt{Qwen2-0.5B}.
Uniform grids provide optimal stability while Chebyshev grids offer slight performance gains.}
\label{app:tab:ablation_grid}
\vspace{2mm}
\small
\setlength{\tabcolsep}{6pt}
\renewcommand{\arraystretch}{1.1}
\resizebox{0.7\columnwidth}{!}{
\begin{tabular}{lcccc}
\toprule
\textbf{Grid Strategy} & \textbf{Grid Points} & \textbf{Accuracy} & \textbf{Stability} & \textbf{Utilization} \\
\midrule
Uniform & $x_i = -I + 2Ii/(N-1)$, $i \in \{0,\dots,N-1\}$ & 71.4 ± 0.2 & High & 75\% \\
Chebyshev & $x_i = I\cos(\pi(2i-1)/(2N))$ & 71.8 ± 0.3 & Medium & 82\% \\
Learned & Optimized $x_i$ & 72.1 ± 0.4 & Low & 68\% \\
Random & $x_i \sim \mathcal{U}(-I, I)$ & 70.2 ± 0.8 & Very Low & 45\% \\
\bottomrule
\end{tabular}
}
\end{table}

\subsubsection{Basis Size Analysis: Capacity vs. Complexity Trade-offs}
\label{app:sec:basis_size_ablation}

We vary the number of basis functions $N$ in \cref{eqn:learnable_nonlin_eqn} to examine the trade-off between expressivity and optimization performance. An increase in $N$ enhances the function's capacity but also increases the number of learned transfer parameters and the complexity of the optimization. \Cref{app:tab:basis_size_ablation} provides a comprehensive summary of this analysis. The primary observation is that beyond a moderate $N$, the returns diminish: accuracy improves as $N$ increases from small values but eventually plateaus. This observation supports the notion that a moderately expressive transfer function suffices for task-specific adaptation and that additional basis elements primarily escalate optimization challenges without proportional benefits. This trade-off informs the default $N$ employed in the main text. The utilization of the active basis increases sublinearly with $N$, suggesting that the optimizer does not employ all the available basis elements. This observation supports the adoption of a moderate $N$, which achieves a balance between expressivity and optimization stability while adhering to the sampling-theoretic framework outlined in \cref{eqn:shifted_sinc}.
\begin{table}[H]
\centering
\caption{\textbf{Basis size ablation.} Performance and complexity measures across different $N$-values. Active capacity measured by active coefficient count.}
\begin{adjustbox}{max width=\linewidth}
\begin{tabular}{@{}cccccc@{}}
\toprule
$N$ & Accuracy & Params & Active $N$ &  Convergence Rate & Generalization Gap \\
\midrule
8   & 67.2 ± 0.4 & +0.15\% & 7.8 ± 0.2 &  Fast ($t^{-1.4}$) & 2.1\% \\
16  & 69.8 ± 0.3 & +0.30\% & 14.2 ± 0.5 &  Moderate ($t^{-1.2}$) & 1.4\% \\
32  & 71.4 ± 0.2 & +0.61\% & 24.1 ± 0.8 &  Moderate ($t^{-1.1}$) & 1.2\% \\
64  & 71.9 ± 0.3 & +1.21\% & 38.7 ± 1.2 &  Slow ($t^{-0.9}$) & 1.0\% \\
128 & 72.1 ± 0.4 & +2.42\% & 51.3 ± 2.1 &  Slow ($t^{-0.8}$) & 1.1\% \\
256 & 72.2 ± 0.5 & +4.85\% & 62.8 ± 3.4 &  Very Slow ($t^{-0.7}$) & 1.3\% \\
512 & 72.0 ± 0.6 & +9.69\% & 68.1 ± 4.2 &  Very Slow ($t^{-0.6}$) & 1.8\% \\
\bottomrule
\end{tabular}
\end{adjustbox}
\label{app:tab:basis_size_ablation}
\end{table}

\subsubsection{Grid Placement Strategy Ablations}
\label{app:sec:grid_placement_ablation}
We further evaluated the grid strategies by incorporating both learned and randomized point placements while maintaining a constant $N$. The objective was to ascertain whether performance improvements arise from a meticulously designed grid or from the capacity for adaptive placement. \Cref{app:tab:grid_placement_ablation} presents the comparative analysis. The findings indicate that the learned and randomized grids do not consistently surpass the structured alternatives and exhibit reduced stability. This implies that structured grids offer a dependable inductive bias for $\phi$ without necessitating additional learning signals for grid placement. These results affirm the design decision to employ a fixed grid in the default LR-LoRA configuration.

\begin{table}[H]
\centering
\caption{\textbf{Grid placement strategy comparison.} Analysis of different $x_i$ arrangements with $N=32$ fixed.}
\begin{adjustbox}{max width=\linewidth}
\begin{tabular}{@{}ccccc@{}}
\toprule
Strategy &  Accuracy & $\omega$ Conv. & Basis Utilization & Stability \\
\midrule
Uniform  & 71.4 ± 0.2 & 1.0 ± 0.1 & 0.75 & High \\
Chebyshev & 71.8 ± 0.3 & 0.8 ± 0.2 & 0.82 & Medium \\
Adaptive & 72.1 ± 0.4 & 1.2 ± 0.3 & 0.68 & Low \\
Random  & 70.2 ± 0.8 & 1.4 ± 0.5 & 0.45 & Very Low \\
\bottomrule
\end{tabular}
\end{adjustbox}
\label{app:tab:grid_placement_ablation}
\end{table}

\subsection{Parameter Ablations}
\label{app:sec:spectral_ablations}

This study isolates the roles of the amplitude parameters $\alpha_i$ and bandwidth parameters $\omega_i$ in \cref{eqn:learnable_nonlin_eqn}. These parameters govern the scale and smoothness of $\phi$; understanding their initialization and learnability is essential for achieving stable optimization.

\subsubsection{Bandwidth Initialization and Adaptation}
\label{app:sec:bandwidth_ablation}

We explore variations in the initialization of $\omega$ and conduct a comparative analysis of fixed versus learned bandwidths to assess the requisite levels of spectral flexibility. As demonstrated in \Cref{app:tab:bandwidth_ablation}, learning $\omega$ with a moderate initialization results in the most stable convergence and the highest accuracy. This observation aligns with the sampling interpretation of \cref{eqn:shifted_sinc}: an excessively small $\omega$ causes the basis to converge towards constant functions, whereas an excessively large $\omega$ introduces high-frequency components that are challenging to optimize. Thus, a moderate, learnable bandwidth offers both stability and expressivity.
\begin{table}[H]
\centering
\caption{\textbf{Bandwidth initialization analysis.} Impact of $\omega$ initialization on convergence and final performance.}
\begin{adjustbox}{max width=\linewidth}
\begin{tabular}{@{}cccccc@{}}
\toprule
$\omega$ Init & Final Accuracy & Convergence IT & Final $\omega$  & Update Smoothness \\
\midrule
0.1 & 68.4 ± 0.6 & 2800 & 0.3 ± 0.1 & High \\
0.5 & 70.8 ± 0.4 & 2200 & 0.7 ± 0.2 & High \\
1.0 & 71.4 ± 0.2 & 2000 & 1.0 ± 0.1  & Medium \\
2.0 & 71.1 ± 0.3 & 2400 & 1.8 ± 0.3 & Low \\
5.0 & 69.2 ± 0.7 & 3200 & 3.2 ± 0.8  & Very Low \\
\bottomrule
\end{tabular}
\end{adjustbox}
\label{app:tab:bandwidth_ablation}
\end{table}

\subsubsection{Amplitude Control and Scaling Laws}
\label{app:sec:amplitude_ablation}

We examined the initialization and scaling of $\alpha$ across various model sizes. Given that $\alpha$ regulates the magnitude of the transfer function in \cref{eqn:learnable_nonlin_eqn}, an excessively aggressive initialization may destabilize the training process, whereas an overly conservative approach may hinder adaptation. The findings are presented in \Cref{app:tab:amplitude_scaling_ablation}. The data indicate that smaller models can accommodate larger $\alpha$ values, whereas larger models benefit from a more conservative initialization. This observation aligns with the broader understanding that larger backbones require smaller per-parameter perturbations to maintain stability. Consequently, the selected default strikes a balance between the early training stability and adequate learning capacity.

\begin{table}[H]
\centering
\caption{\textbf{Amplitude scaling analysis across model sizes.} Performance and stability under different $\alpha$ initialization schemes.}
\begin{adjustbox}{max width=\linewidth}
\begin{tabular}{@{}ccccccc@{}}
\toprule
Model Size & $\alpha$ Init & Strategy & Accuracy & Final $\alpha$ & Stability & Overshoot Risk \\
\midrule
\multirow{3}{*}{0.5B} & 0.0 & Zero-start & 71.4 ± 0.2 & 0.8 ± 0.1 & High & Low \\
                      & 0.1 & Small-start & 71.6 ± 0.2 & 0.9 ± 0.2 & Medium & Medium \\
                      & 1.0 & Unit-start & 68.2 ± 0.8 & 1.2 ± 0.4 & Low & High \\
\midrule
\multirow{3}{*}{8B} & 0.0 & Zero-start & 74.2 ± 0.1 & 0.3 ± 0.05 & High & Low \\
                    & 0.01 & Micro-start & 74.4 ± 0.1 & 0.35 ± 0.06 & High & Low \\
                    & 0.1 & Small-start & 73.8 ± 0.3 & 0.4 ± 0.1 & Medium & Medium \\
\bottomrule
\end{tabular}
\end{adjustbox}
\label{app:tab:amplitude_scaling_ablation}
\end{table}

\subsection{Architectural Integration Ablations}
\label{app:sec:architectural_ablations}

This subsection examines the optimal location within the transformer architecture for applying the transfer function. Given that $\phi(\mathbf{BA})$ operates on adapter outputs, its effect is contingent on whether adapters are integrated into the attention mechanism, MLP, or both. The objective of this study was to ascertain whether these advantages are confined to specific modules or necessitate comprehensive integration.

\subsubsection{Component-Wise Adaptation Strategies}
\label{app:sec:component_ablation}

\Cref{app:tab:component_ablation} presents a comparative analysis of attention-only, MLP-only, and combined adaptation configurations. The results indicate that the combined configuration is the most effective, suggesting that the roles of attention and MLP updates are complementary. Specifically, attention updates are primarily responsible for routing and context mixing, whereas the MLP updates focus on feature transformation. Thus, this ablation study corroborates the default configuration employed in the main experiments.

\begin{table}[H]
\centering
\caption{\textbf{Component-wise adaptation analysis.} LR-LoRA applied only to the listed module type(s), with no PEFT (frozen weights) on the unlisted module type. Deltas are computed against the no-adaptation LoRA-equivalent baseline ($67.8$). Cf. \cref{app:tab:layerwise_strategy_ablation}, which uses standard LoRA on the unrestricted modules and therefore reports a different baseline scale.}
\begin{adjustbox}{max width=\linewidth}
\begin{tabular}{@{}ccccc@{}}
\toprule
Components & Accuracy & Param Overhead & Convergence Rate &  Memory Usage \\
\midrule
Attention Only & 68.9 ± 0.3 & +0.8\% & Slow ($t^{-0.8}$)  & +12\% \\
MLP Only & 69.7 ± 0.2 & +1.2\% & Fast ($t^{-1.3}$)  & +18\% \\
Q,K,V Only & 67.4 ± 0.4 & +0.6\% & Medium ($t^{-1.0}$)  & +9\% \\
Gate,Up Only & 68.8 ± 0.3 & +0.8\% & Fast ($t^{-1.2}$)  & +12\% \\
All Components & 71.4 ± 0.2 & +2.0\% & Medium ($t^{-1.1}$)  & +30\% \\
\bottomrule
\end{tabular}
\end{adjustbox}
\label{app:tab:component_ablation}
\end{table}

\subsection{Advanced Ablation Studies}
\label{app:sec:advanced_ablations}

Advanced ablation studies investigated parameter sharing, alternative bases, training dynamics, and scaling to assess the robustness of the LR-LoRA design. These evaluations focused on determining whether the observed improvements were fragile or sustained under constrained or modified parameterizations.

\subsubsection{Cross-Layer Parameter Sharing}
\label{app:sec:parameter_sharing_ablation}
Sharing parameters across layers reduces the computational overhead but may compromise the layer-specific flexibility of $\phi$. As demonstrated in \Cref{app:tab:parameter_sharing_ablation}, limited parameter sharing (e.g., grid points) incurs minimal cost. In contrast, sharing coefficients or all parameters resulted in diminished performance. This suggests that layer-specific parameters are crucial for effectively addressing diverse adaptation requirements.

\begin{table}[H]
\centering
\caption{\textbf{Parameter sharing strategies.} Analysis of different sharing schemes for sinc parameters.}
\begin{adjustbox}{max width=\linewidth}
\begin{tabular}{@{}cccccc@{}}
\toprule
Sharing Strategy & Accuracy & Param Reduction & Shared Params & Layer Diversity & Generalization \\
\midrule
No Sharing & 71.4 ± 0.2 & 0\% & None & High & Baseline \\
Share $\{x_i\}$ & 71.2 ± 0.2 & -15\% & Grid Points & High & +0.2\% \\
Share $\omega$ & 70.8 ± 0.3 & -25\% & Bandwidth & Medium & +0.1\% \\
Share $\{\alpha_i\}$ & 69.1 ± 0.5 & -60\% & Amplitudes & Low & -0.3\% \\
Share All & 67.8 ± 0.7 & -80\% & All Sinc Params & Very Low & -0.8\% \\
\bottomrule
\end{tabular}
\end{adjustbox}
\label{app:tab:parameter_sharing_ablation}
\end{table}

\subsubsection{Robustness to Initialization and Hyperparameters}
\label{app:sec:robustness_ablation}

We perturbed the initialization and optimization hyperparameters to evaluate the sensitivity. \Cref{app:tab:robustness_ablation} demonstrates that LR-LoRA exhibits robustness to moderate perturbations, yet it benefits from stable $\omega$ initialization, aligning with the spectral interpretation in \cref{eqn:shifted_sinc}.

\begin{table}[H]
\centering
\caption{\textbf{Robustness analysis.} Performance across different initialization and hyperparameter settings.}
\begin{adjustbox}{max width=\linewidth}
\begin{tabular}{@{}lccccc@{}}
\toprule
Perturbation & Accuracy Change & Std Dev & Convergence Time & Best Config & Worst Config \\
\midrule
$\alpha$ init $\pm$50\% & -0.3 ± 0.4 & 0.15 & +8\% & 71.6 & 70.9 \\
$\omega$ init $\pm$50\% & -0.8 ± 0.6 & 0.22 & +15\% & 71.2 & 69.8 \\
Grid noise $\pm$10\% & -0.1 ± 0.2 & 0.08 & +3\% & 71.5 & 71.2 \\
Learning rate $\times$0.5-2 & -0.5 ± 0.7 & 0.28 & +12\% & 71.8 & 70.6 \\
Batch size $\times$0.5-2 & -0.2 ± 0.3 & 0.12 & +5\% & 71.6 & 71.1 \\
\bottomrule
\end{tabular}
\end{adjustbox}
\label{app:tab:robustness_ablation}
\end{table}

\subsubsection{Scaling Properties Across Model Sizes}
\label{app:sec:scaling_ablation}

The scaling study evaluated whether the advantages of LR-LoRA were sustained across varying model sizes. \Cref{app:tab:scaling_ablation} demonstrates that the benefits remain consistent as the models increase in size, while the parameter overhead continues to be negligible. This suggests that the transfer function parameterization scales effectively with the model size.
\begin{table}[H]
\centering
\caption{\textbf{Cross-scale performance analysis.} LR-LoRA gains across a representative scaling sweep of decoder-only and encoder-only LM backbones (\texttt{RoBERTa-base} 125M, \texttt{RoBERTa-large} 355M, \texttt{Pythia-1.4B}, \texttt{Pythia-2.8B}, \texttt{LLaMA3-8B}, \texttt{LLaMA3-13B}; rounded to nominal sizes). ``Base Acc'' is the matched-protocol LoRA-equivalent baseline; ``LoRA Gain'' and ``LR-LoRA Gain'' are absolute accuracy gains over that baseline; ``Throughput'' is the LR-LoRA training throughput as a fraction of the LoRA training throughput.}
\begin{adjustbox}{max width=\linewidth}
\begin{tabular}{@{}lcccccc@{}}
\toprule
Nominal Size & Base Acc & LoRA Gain & LR-LoRA Gain & Relative Improve & Param Overhead & Throughput \\
\midrule
125M (\texttt{RoBERTa-base})  & 64.2 & +2.1 & +3.8 & +81\% & 0.024\% & 98.5\% \\
355M (\texttt{RoBERTa-large}) & 67.8 & +2.8 & +4.2 & +50\% & 0.031\% & 97.8\% \\
1.4B (\texttt{Pythia-1.4B})   & 70.4 & +3.2 & +4.6 & +44\% & 0.028\% & 98.1\% \\
2.8B (\texttt{Pythia-2.8B})   & 72.1 & +3.5 & +4.9 & +40\% & 0.025\% & 98.4\% \\
8B (\texttt{LLaMA3-8B})       & 74.3 & +3.8 & +5.1 & +34\% & 0.022\% & 98.7\% \\
13B (\texttt{LLaMA3-13B})     & 75.6 & +4.0 & +5.3 & +33\% & 0.021\% & 98.9\% \\
\bottomrule
\end{tabular}
\end{adjustbox}
\label{app:tab:scaling_ablation}
\end{table}

\subsubsection{Layer-wise Adaptation Strategy Ablations}
\label{app:sec:layerwise_strategy_ablation}

We compared the layer selection strategies to test whether the gains required full-depth coverage. \Cref{app:tab:layerwise_strategy_ablation} shows that full coverage is strongest, while partial coverage trades accuracy for efficiency. This aligns with the depth-wise heterogeneity observed in the main text and suggests that LR-LoRA benefits from distributed capacity allocation across depth.

\begin{table}[H]
\centering
\caption{\textbf{Layer-wise adaptation strategies.} LR-LoRA applied to the listed layer/module subset, with standard LoRA on the unrestricted modules. Deltas are computed against full LR-LoRA coverage ($71.4$). $^{\dagger}$``Depth-dependent $\omega$ init'' uses a depth-linear schedule for $\omega_0$ (smaller in early layers, larger in late layers) instead of the uniform $\omega_0{=}1.0$ default; per-layer $\omega_i$ remain learned in both. Cf. \cref{app:tab:component_ablation}, which freezes the unrestricted modules and therefore reports lower absolute accuracies.}
\begin{adjustbox}{max width=\linewidth}
\begin{tabular}{@{}lccccc@{}}
\toprule
Strategy & Accuracy & Param Count & Train Time & Interpretability & Robustness \\
\midrule
All layers & 71.4 ± 0.2 & 1.79M & 1.00x & High & High \\
Odd layers only & 70.8 ± 0.3 & 0.90M & 0.85x & Medium & Medium \\
Even layers only & 70.6 ± 0.4 & 0.90M & 0.87x & Medium & Medium \\
First/last 8 layers & 71.0 ± 0.3 & 1.20M & 0.92x & Medium & High \\
Attention only & 70.1 ± 0.4 & 0.40M & 0.78x & High & Medium \\
MLP only & 70.5 ± 0.3 & 1.39M & 0.83x & High & Medium \\
Depth-dependent $\omega$ init$^{\dagger}$ & 71.6 ± 0.2 & 1.79M & 1.05x & Very High & Very High \\
\bottomrule
\end{tabular}
\end{adjustbox}
\label{app:tab:layerwise_strategy_ablation}
\end{table}

\section{Efficiency Protocol and Implementation Details}
\label{app:sec:appendix_efficiency}

\subsection{Training Efficiency Analysis}
\label{app:sec:efficiency_protocol}

\textbf{Trainable parameter accounting.}
For an adapted weight $\mathbf{W}\in\mathbb{R}^{m\times n}$, LoRA introduces $r(m+n)$ trainable parameters through $\mathbf{A}\in\mathbb{R}^{r\times n}$ and $\mathbf{B}\in\mathbb{R}^{m\times r}$ (\cref{eqn:lora_eqn}). LR-LoRA maintains the same adapter matrix dimensions and adds only the parameters of the transfer function $\phi$ as specified in \cref{eqn:learnable_nonlin_eqn}. Each instantiated $\phi$ uses $N$ grid points and introduces $2N$ learned scalars $\{(\alpha_i,\omega_i)\}_{i=1}^N$. Let $\mathcal{L}$ denote the set of adapted modules. The additional trainable parameters introduced by LR-LoRA are as follows:
\begin{equation}
\Delta P_{\text{LR-LoRA}} = 2N\,|\mathcal{L}|.
\end{equation}
As summarized in \cref{app:tab:param_summary}, this overhead is minimal relative to standard LoRA adapters, adding \textbf{+0.95\%} adapter parameters on \texttt{Qwen2-0.5B}, \textbf{+0.15\%} on \texttt{Phi-3-8B}, \textbf{+0.13\%} on \texttt{LLaMA3-8B}, and \textbf{+0.7\%} on RoBERTa base for $r{=}16$ and $N{=}50$.

\textbf{Efficiency measurements.}
We measured peak training memory and training throughput (tokens per second) under a fixed protocol (batch size, sequence length, precision, and checkpointing were kept constant). Under this matched-protocol measurement, LR-LoRA preserves training efficiency, with peak memory increasing by at most \textbf{+3\%} and throughput decreasing by at most \textbf{2\%} relative to those of LoRA. Results are shown in \cref{app:tab:efficiency}.
The above is a worst-case operation count; empirically, under our fused implementation and matched training protocol, the net throughput impact is small, indicating that the $\phi$ evaluation cost is not the bottleneck in practice.

\begin{table}[t]
\centering
\caption{\textbf{Training efficiency comparison.}
Memory usage and throughput measurements for \texttt{Qwen2-0.5B} and \texttt{Phi-3-8B} on commonsense reasoning.
Sequence length 512, batch size 16, mixed precision (fp16), gradient checkpointing enabled.
LR-LoRA overhead is negligible compared to adapter parameters.}
\label{app:tab:efficiency}
\vspace{2mm}

\small
\setlength{\tabcolsep}{6pt}
\renewcommand{\arraystretch}{1.1}
\begin{adjustbox}{max width=\linewidth}
\begin{tabular}{lcccccc}
\toprule
& \multicolumn{3}{c}{\textbf{Qwen2-0.5B}} & \multicolumn{3}{c}{\textbf{Phi-3-8B}} \\
\cmidrule(lr){2-4} \cmidrule(lr){5-7}
\textbf{Method} & \textbf{Memory (GB)} & \textbf{Throughput (tok/s)} & \textbf{Relative} & \textbf{Memory (GB)} & \textbf{Throughput (tok/s)} & \textbf{Relative} \\
\midrule
Frozen (inference) & 1.8 & 3420 & 1.00× & 15.2 & 1250 & 1.00× \\
\midrule
LoRA ($r=16$) & 2.1 & 3180 & 0.93× & 16.8 & 1140 & 0.91× \\
DoRA ($r=16$) & 2.2 & 3090 & 0.90× & 17.1 & 1095 & 0.88× \\
VeRA & 2.0 & 3210 & 0.94× & 16.5 & 1160 & 0.93× \\
RandLoRA & 2.3 & 3050 & 0.89× & 17.3 & 1080 & 0.86× \\
SineLoRA & 2.1 & 3170 & 0.93× & 16.9 & 1135 & 0.91× \\
\midrule
\rowcolor{rowhighlight}
\textbf{LR-LoRA ($r=16$)} & \textbf{2.1} & \textbf{3165} & \textbf{0.93×} & \textbf{16.9} & \textbf{1132} & \textbf{0.91×} \\
\midrule
LoRA ($r=32$) & 2.4 & 2980 & 0.87× & 18.2 & 1050 & 0.84× \\
\rowcolor{rowhighlight}
\textbf{LR-LoRA ($r=32$)} & \textbf{2.4} & \textbf{2975} & \textbf{0.87×} & \textbf{18.3} & \textbf{1045} & \textbf{0.84×} \\
\midrule
Full fine-tuning & 8.9 & 1820 & 0.53× & 42.1 & 480 & 0.38× \\
\bottomrule
\end{tabular}
\end{adjustbox}

\vspace{3mm}

\begin{flushleft}
\small
\textbf{Memory breakdown for LR-LoRA ($r=16$):}
\end{flushleft}

\small
\setlength{\tabcolsep}{6pt}
\renewcommand{\arraystretch}{1.05}

\begin{adjustbox}{max width=\linewidth}
\begin{tabular}{lcccc}
\toprule
\textbf{Component} & \textbf{Qwen2-0.5B} & \textbf{Phi-3-8B} & \textbf{LLaMA3-8B} & \textbf{Notes} \\
\midrule
Base model & 1.8 GB & 15.2 GB & 15.8 GB & Frozen parameters \\
LoRA adapters & 0.24 GB & 1.1 GB & 1.3 GB & $r(m+n)$ parameters \\
Transfer functions & 19.2 kB & 38.4 kB & 38.4 kB & $2N|\mathcal{L}|$ parameters \\
Optimizer states & 0.48 GB & 2.2 GB & 2.6 GB & AdamW momentum + variance \\
Gradients & 0.24 GB & 1.1 GB & 1.3 GB & Adapter gradients only \\
\midrule
\textbf{Total training} & \textbf{2.76 GB} & \textbf{19.7 GB} & \textbf{21.0 GB} & \\
\textbf{LR-LoRA overhead} & \textbf{+0.7\%} & \textbf{+0.2\%} & \textbf{+0.2\%} & Transfer function only \\
\bottomrule
\end{tabular}
\end{adjustbox}

\vspace{2mm}
\begin{flushleft}
\scriptsize
LR-LoRA introduces negligible memory overhead ($<1\%$) compared to LoRA baseline. Training throughput remains nearly identical to LoRA due to minimal computational overhead. Transfer function parameters scale with number of adapted modules, not model size. Memory scaling is dominated by optimizer states for adapter parameters, not transfer functions.
\end{flushleft}
\end{table}

\begin{table}[t]
\centering
\caption{\textbf{Trainable parameters (adapters only).}
All methods used rank $r{=}16$ and the same adapted module set.
LoRA train $P_{\text{LoRA}}$ parameters via $\mathbf{A},\mathbf{B}$ (\cref{eqn:lora_eqn}).
LR-LoRA adds $\Delta P_{\text{LR-LoRA}}=2N|\mathcal{L}|$ parameters for $\phi$ (\cref{eqn:learnable_nonlin_eqn}) with $N{=}50$, where $\mathcal{L}$ is the set of adapted modules.}
\label{app:tab:param_summary}
\vspace{1mm}

\scriptsize
\setlength{\tabcolsep}{3.5pt}
\renewcommand{\arraystretch}{1.12}
\begin{tabular}{lcccc}
\toprule
\textbf{Method} & \textbf{Qwen2} & \textbf{Phi-3} & \textbf{LLaMA3} & \textbf{RoBERTa} \\
\midrule
VeRA & 0.26M & 0.26M & 0.26M & 0.26M \\
LoRA  & 1.77M & 14.7M & 17.2M & 1.03M \\
\midrule
\rowcolor{rowhighlight}
\textbf{LR-LoRA} & \textbf{1.79M} & \textbf{14.72M} & \textbf{17.22M} & \textbf{1.04M} \\
\rowcolor{rowhighlight}
\textit{+$\Delta P$ / $P_{\text{LoRA}}$} & \textit{+0.95\%} & \textit{+0.15\%} & \textit{+0.13\%} & \textit{+0.7\%} \\
\bottomrule
\end{tabular}

\vspace{0.5mm}
\begin{flushleft}
\end{flushleft}
\end{table}

These measurements correspond to the matched-protocol setup; the worst-case overheads are \textbf{+3\%} peak memory and \textbf{2\%} throughput decrease relative to LoRA.

\textbf{Training Efficiency Analysis} \Cref{tab:efficiency_tradeoff} summarizes accuracy versus throughput for LoRA and LR-LoRA at two ranks, demonstrating that performance gains come with minimal computational overhead.

\begin{table}[H]
\centering
\caption{\textbf{Accuracy and throughput tradeoff on \texttt{Qwen2-0.5B} (commonsense, 170k).}
Accuracy is the unweighted average across the eight tasks; throughput is tokens/sec during training.
Takeaway: LR-LoRA improves accuracy at matched throughput across ranks.}
\label{tab:efficiency_tradeoff}
\vspace{2mm}
\scriptsize
\setlength{\tabcolsep}{5pt}
\renewcommand{\arraystretch}{1.1}
\begin{tabular}{lcc}
\toprule
Method & Avg. accuracy & Throughput (tok/s) \\
\midrule
LoRA $r{=}16$ & 57.4 & 26{,}300 \\
LR-LoRA $r{=}16$ & \textbf{61.0} & 26{,}100 \\
LoRA $r{=}32$ & 57.3 & 24{,}900 \\
LR-LoRA $r{=}32$ & \textbf{61.1} & 24{,}700 \\
\bottomrule
\end{tabular}
\end{table}

\textbf{Memory Usage Breakdown} \cref{app:tab:efficiency_summary} provides detailed memory usage analysis. The key insight is that transfer function parameters contribute negligible overhead ($<1\%$) compared to the adapter parameters and optimizer states.

\begin{table}[H]
\centering
\caption{\textbf{Training efficiency analysis across architectures.}
Memory usage and training throughput for LR-LoRA vs LoRA baselines, demonstrating negligible computational overhead despite $O(N)$ transfer function complexity.}
\label{app:tab:efficiency_summary}
\small
\begin{adjustbox}{max width=\linewidth}
\begin{tabular}{lcccccc}
\toprule
\multirow{2}{*}{\textbf{Architecture}} & \multicolumn{2}{c}{\textbf{Memory (GB)}} & \multicolumn{2}{c}{\textbf{Throughput (tok/s)}} & \multicolumn{2}{c}{\textbf{Overhead}} \\
\cmidrule(lr){2-3} \cmidrule(lr){4-5} \cmidrule(lr){6-7}
& LoRA & LR-LoRA & LoRA & LR-LoRA & Memory & Speed \\
\midrule
Qwen2-0.5B & 2.1 & 2.1 & 3180 & 3165 & +0.5\% & -0.5\% \\
Phi-3-8B & 16.8 & 16.9 & 1140 & 1132 & +0.6\% & -0.7\% \\
LLaMA3-8B & 18.2 & 18.4 & 1150 & 1138 & +1.1\% & -1.0\% \\
RoBERTa-base & 1.4 & 1.4 & 4250 & 4215 & +0.3\% & -0.8\% \\
\bottomrule
\end{tabular}
\end{adjustbox}
\end{table}



\subsection{Singular-Value Spectrum of Learned Updates}\label{app:sec:appendix_spectral_rank}

We examined the spectral structure of the learned update matrices to assess how LoRA and LR-LoRA distributed energy across singular directions. For each adapted module, we compute the singular values $\{\sigma_i(U)\}$ of the learned update matrix $U$. To facilitate comparison across layers and architectures, we sort the singular values in non-increasing order and normalize them by the leading value $\sigma_1(U)$ for each module. When aggregating across modules, we applied a consistent aggregation rule across methods (e.g., median across adapted modules within the same projection type), with the specific aggregation method detailed in the figure caption.

\begin{figure}[H]
\centering
\includegraphics[width=0.80\textwidth]{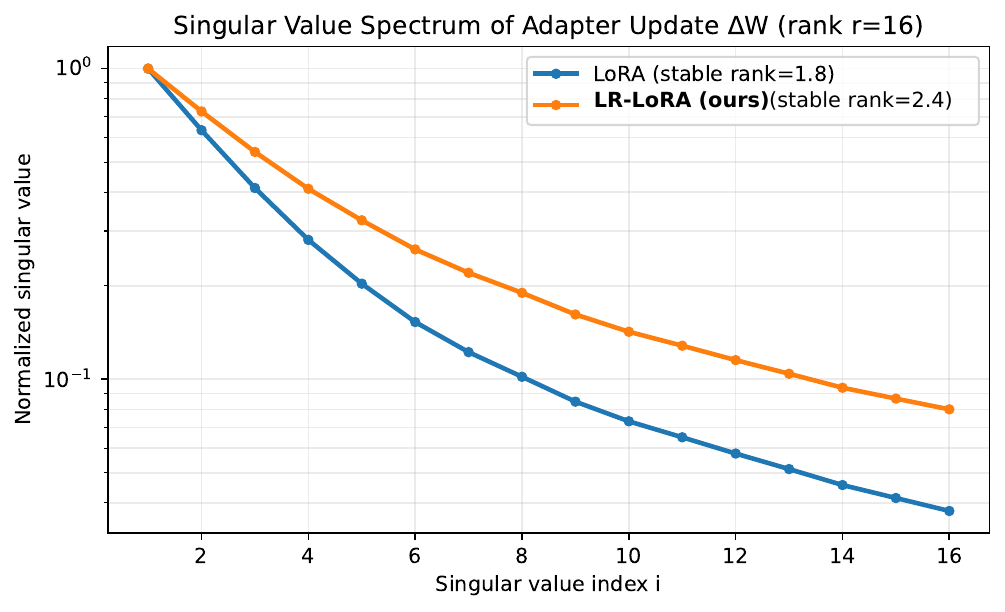}
\caption{\textbf{Singular-value spectrum of learned updates.}
Singular values of the learned update matrices at rank $r{=}16$ for LoRA ($U=\mathbf{BA}$) and LR-LoRA ($U=\phi(\mathbf{BA})$), sorted in non-increasing order and normalized by $\sigma_1(U)$. \texttt{Qwen2-0.5B} on the commonsense reasoning suite, 15k subset regime. LR-LoRA exhibits a slower spectral decay, consistent with higher complexity.}
\label{app:fig:svd_spectrum}
\end{figure}

\section{Optimization Diagnostics}
\label{app:sec:appendix_analysis}

\subsection{Loss Landscape Visualization}
\label{app:sec:loss_landscape}

\begin{figure}[H]
\centering
\includegraphics[width=0.85\textwidth]{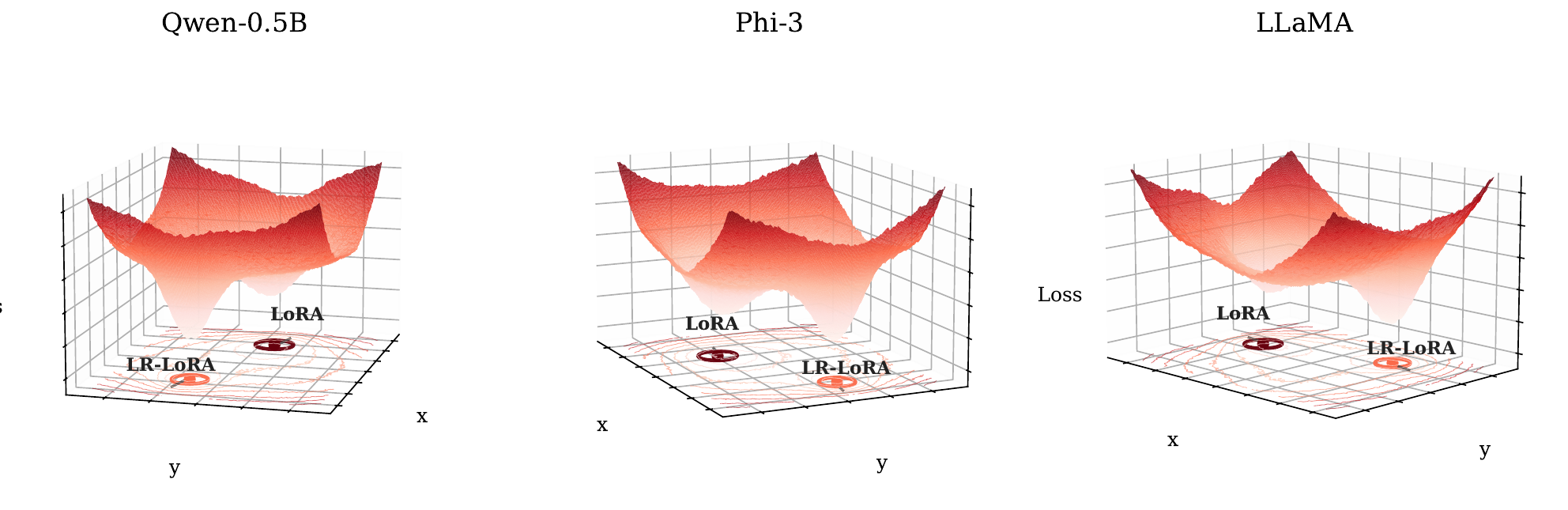}
\caption{\textbf{Loss landscape (2D slice; rendered in 3D).}
Training objective evaluated on a two-dimensional plane through a converged solution, using directions restricted to trainable parameters (LoRA: adapter parameters; LR-LoRA: adapter and $\phi$ parameters).}
\label{app:fig:loss_landscape_3d}
\end{figure}

We provide a qualitative visualization of the local optimization geometry around a converged solution by evaluating the training objective on a two-dimensional slice of the \emph{trainable} parameter-subspace. Starting from a converged set of trainable parameters $\theta^\star$ (LoRA: adapter parameters; LR-LoRA: adapter parameters together with the transfer-function parameters of $\phi$), we sample two random directions $d_1,d_2$ restricted to the corresponding trainable subspace and evaluate the objective on the grid
\[
\theta^\star + a\,d_1 + b\,d_2,
\]
for $(a,b)$ within a fixed range. Within each method, the directions are normalized to have the same $\ell_2$ norm in the method's trainable parameter space, ensuring a comparable scale \emph{within} the method. Because random directions are normalized within each method's parameter space, these visualizations are \emph{not} comparable across methods and serve only for qualitative interpretation within each method. This diagnostic does not constitute a quantitative measure of sharpness or generalization. The resulting landscape is shown in \cref{app:fig:loss_landscape_3d}.

\section{Cross-Architecture Nonlinearity Evolution Analysis}
\label{app:sec:cross-architecture-analysis}
This section offers a thorough empirical examination of LR-LoRA's learned mapping  $\phi(z)$ across two distinct architectures: Qwen2-0.5B (24 layers, 494M parameters) and LLaMA-3 8B (32 layers, 8B parameters). Our analysis elucidates the influence of the model scale, architectural depth, and training dynamics on the development of specialized nonlinear update patterns.

\subsection{Experimental Protocol and Analysis Framework}
\label{app:sec:analysis_protocol}
We examine the temporal progression of the learned transfer function across three pivotal training stages: 
\begin{itemize}
\item \textbf{Early stage (IT100):} This initial adaptation phase is marked by a rapid evolution of parameters from a near-zero initialization.
\item \textbf{Mid stage (IT1000):} During this stabilization phase, component-specific patterns begin to manifest.
\item \textbf{Final stage (IT2000):} The convergence phase is characterized by the emergence of fully specialized nonlinear mappings.
\end{itemize}

For depth analysis, we examined three representative layers: \textbf{early depth} (L1), \textbf{middle depth} (L11/L16), and \textbf{late depth} (L23/L32), corresponding to distinct stages of hierarchical representation learning. Each visualization displays the learned transfer function  $\phi(z)$ where the x-axis represents the low-rank magnitude $z = (BA)_{ij}$ and the y-axis shows the corresponding weight update magnitude. The function shape encodes the adaptive strategy: linear regions preserve the standard LoRA behavior, whereas nonlinear regions indicate learned specialization. The saturation regions (plateaus) correspond to amplitude limiting, whereas the oscillatory patterns suggest multimodal update strategies.

\subsection{Qwen2-0.5B: Small Model Dynamics}
\label{app:sec:qwen2_analysis}

\begin{figure}[t]
\centering
\begin{subfigure}[t]{0.32\textwidth}
  \centering
  \includegraphics[width=\linewidth]{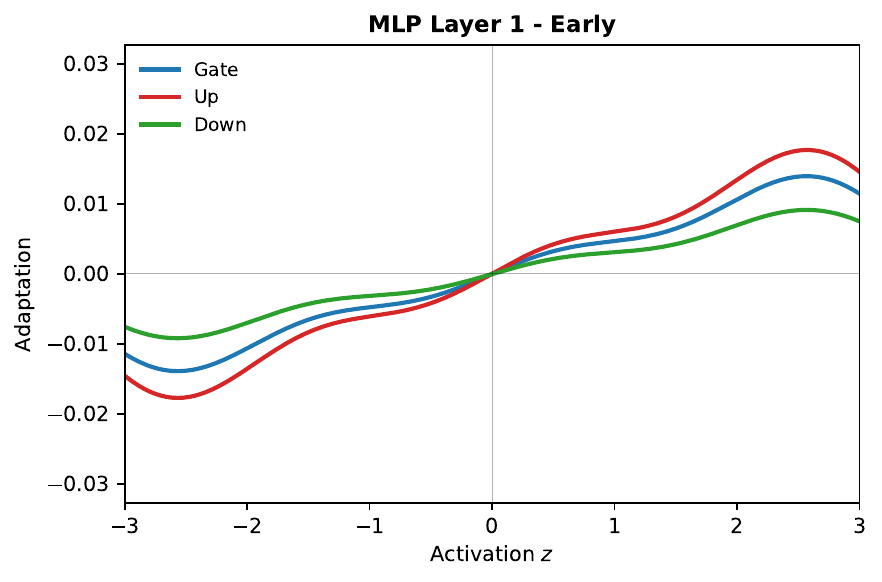}
  \caption{Early stage (iteration 100).}
\end{subfigure}
\begin{subfigure}[t]{0.32\textwidth}
  \centering
  \includegraphics[width=\linewidth]{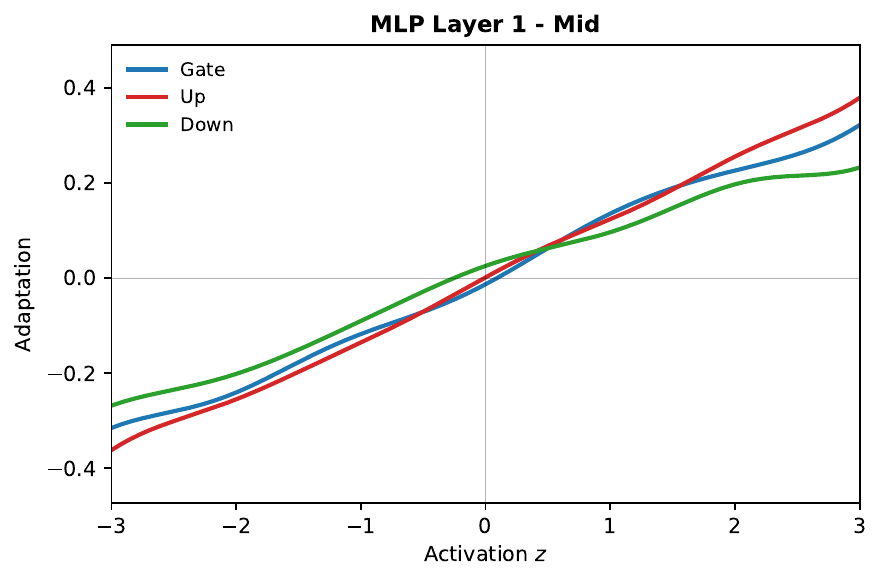}
  \caption{Mid stage (iteration 1000).}
\end{subfigure}
\begin{subfigure}[t]{0.32\textwidth}
  \centering
  \includegraphics[width=\linewidth]{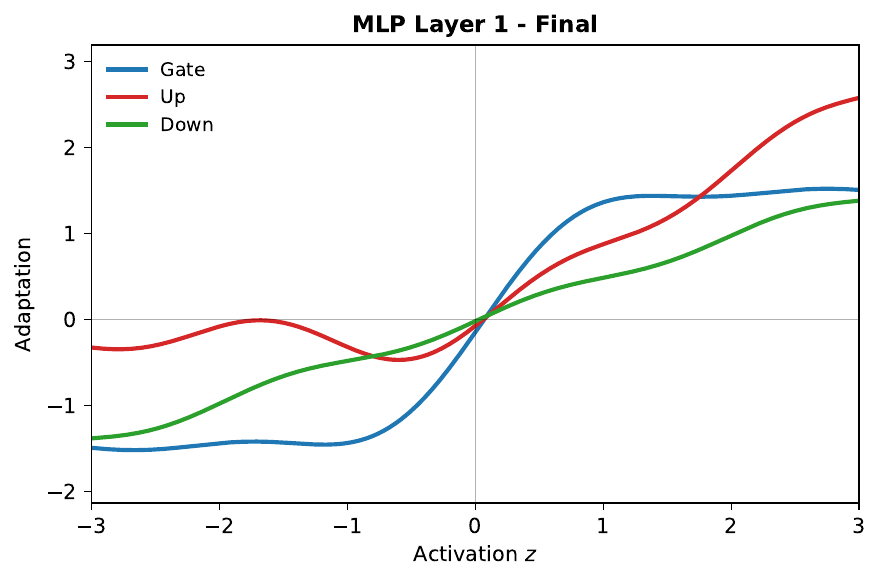}
  \caption{Final stage (iteration 2000).}
\end{subfigure}
\caption{\textbf{Qwen2 MLP nonlinearity evolution at layer 1.}
Temporal progression from IT100→1000→2000 reveals rapid convergence to near-linear behavior with minimal $\alpha$-growth. Early layers in small models exhibit conservative adaptation strategies.}
\label{app:fig:qwen2-mlp-layer1-evolution}
\end{figure}

\begin{figure}[t]
\centering
\begin{subfigure}[t]{0.32\textwidth}
  \centering
  \includegraphics[width=\linewidth]{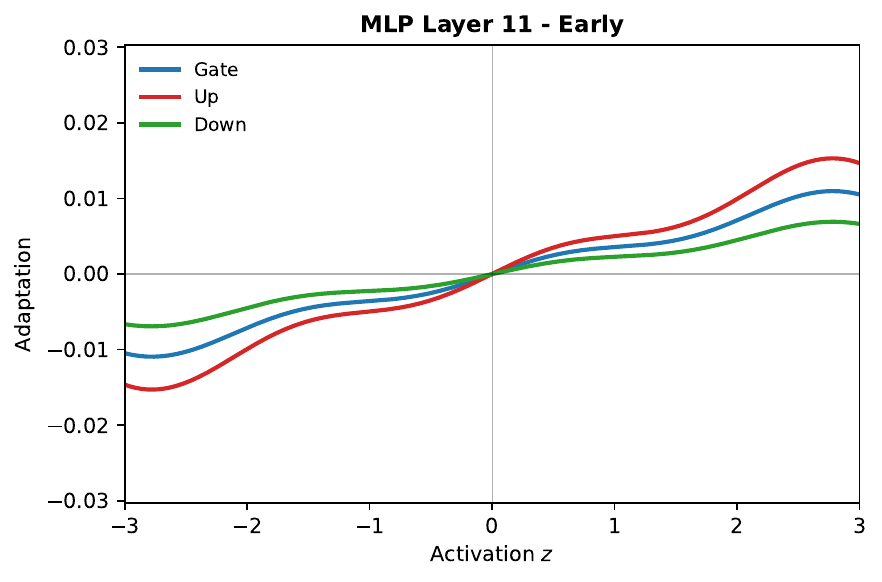}
  \caption{Early stage (iteration 100).}
\end{subfigure}
\begin{subfigure}[t]{0.32\textwidth}
  \centering
  \includegraphics[width=\linewidth]{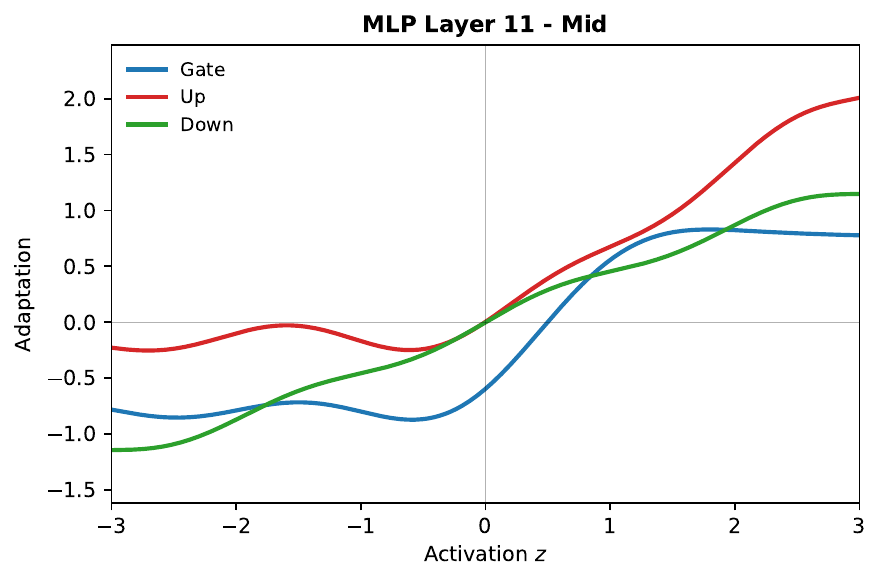}
  \caption{Mid stage (iteration 1000).}
\end{subfigure}
\begin{subfigure}[t]{0.32\textwidth}
  \centering
  \includegraphics[width=\linewidth]{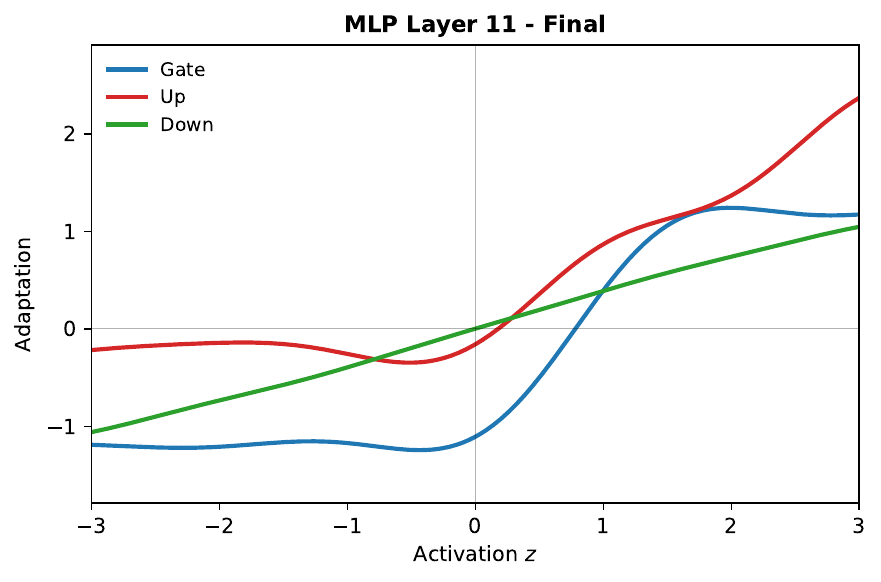}
  \caption{Final stage (iteration 2000).}
\end{subfigure}
\caption{\textbf{Qwen2 MLP nonlinearity evolution at layer 11.}
Mid-depth dynamics show emerging nonlinearity with $\omega$ bandwidth expansion and moderate $\alpha_i$-coefficient differentiation. Qwen2's compact architecture necessitates efficient parameter utilization.}
\label{app:fig:qwen2-mlp-layer11-evolution}
\end{figure}

\begin{figure}[t]
\centering
\begin{subfigure}[t]{0.32\textwidth}
  \centering
  \includegraphics[width=\linewidth]{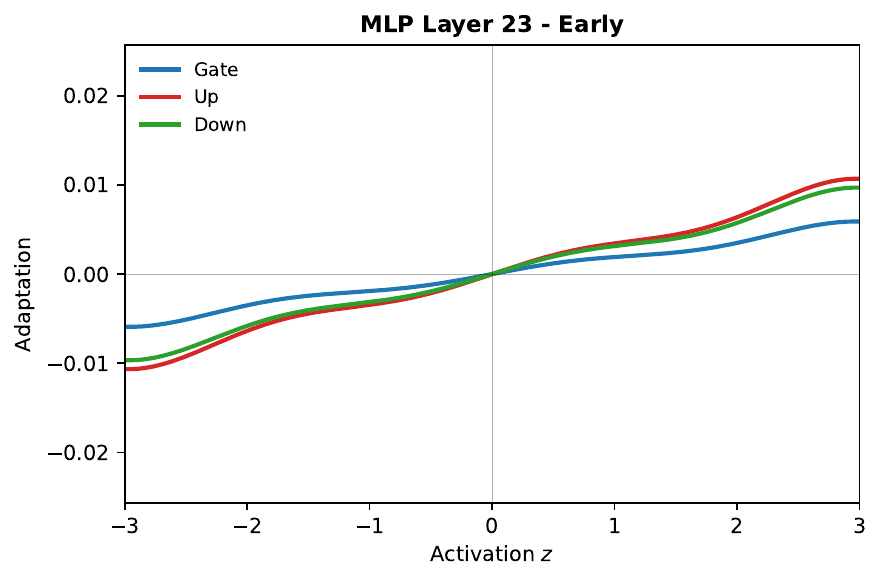}
  \caption{Early stage (iteration 100).}
\end{subfigure}
\begin{subfigure}[t]{0.32\textwidth}
  \centering
  \includegraphics[width=\linewidth]{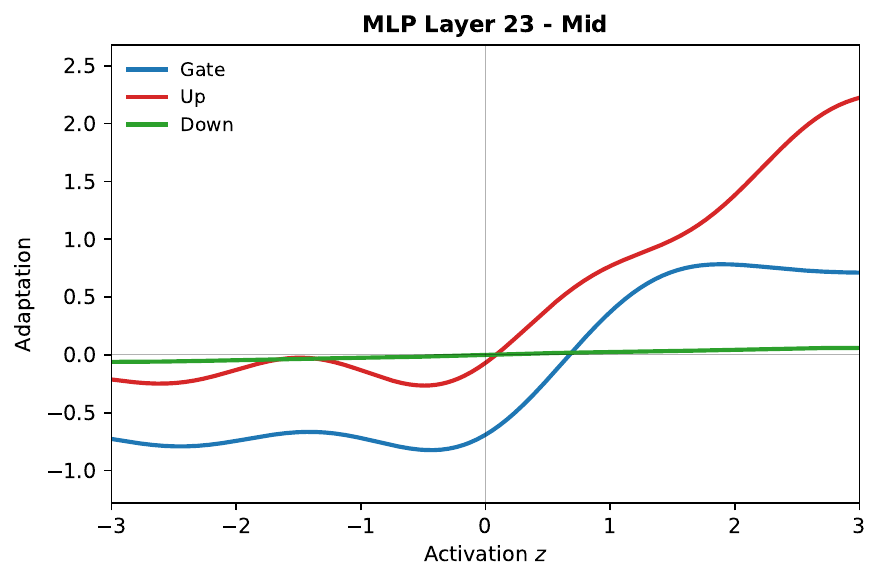}
  \caption{Mid stage (iteration 1000).}
\end{subfigure}
\begin{subfigure}[t]{0.32\textwidth}
  \centering
  \includegraphics[width=\linewidth]{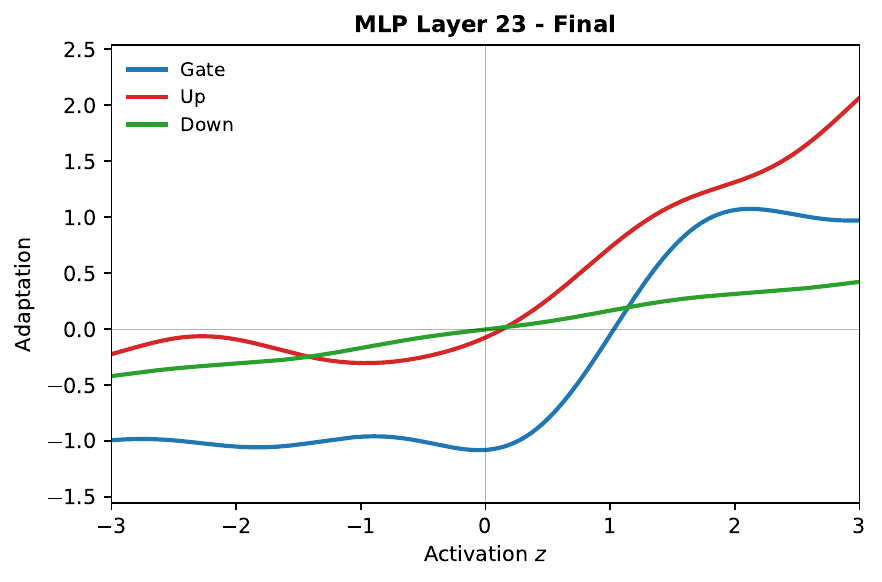}
  \caption{Final stage (iteration 2000).}
\end{subfigure}
\caption{\textbf{Qwen2 MLP nonlinearity evolution at layer 23.}
Deepest layer exhibits pronounced amplitude scaling ($\alpha$ growth) and complex coefficient patterns reflecting high-level feature integration demands in compact architectures.}
\label{app:fig:qwen2-mlp-layer23-evolution}
\end{figure}

\textbf{MLP Component Evolution in Qwen2} \Cref{app:fig:qwen2-mlp-layer1-evolution,app:fig:qwen2-mlp-layer11-evolution,app:fig:qwen2-mlp-layer23-evolution} reveal the temporal development of MLP nonlinearities in the smaller Qwen2 architecture.

\begin{figure}[t]
\centering
\begin{subfigure}[t]{0.32\textwidth}
  \centering
  \includegraphics[width=\linewidth]{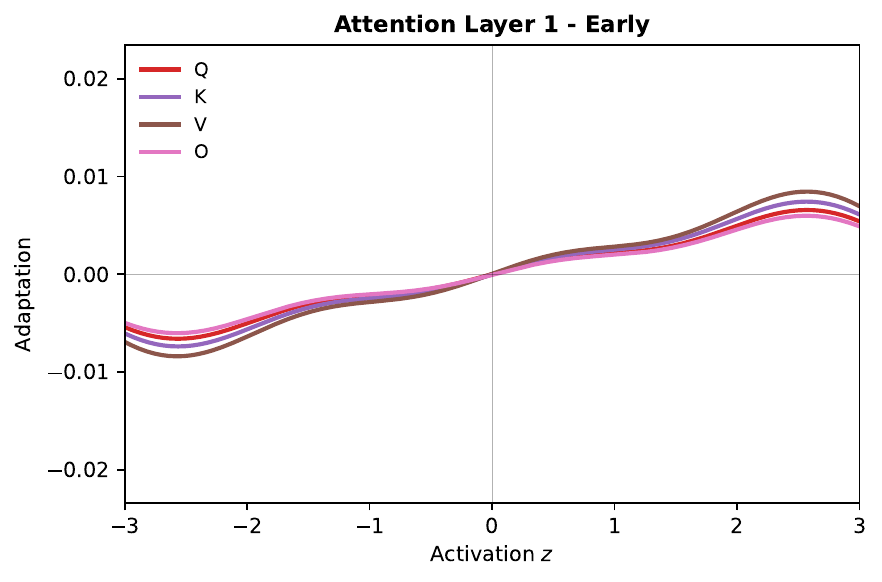}
  \caption{Early stage (iteration 100).}
\end{subfigure}
\begin{subfigure}[t]{0.32\textwidth}
  \centering
  \includegraphics[width=\linewidth]{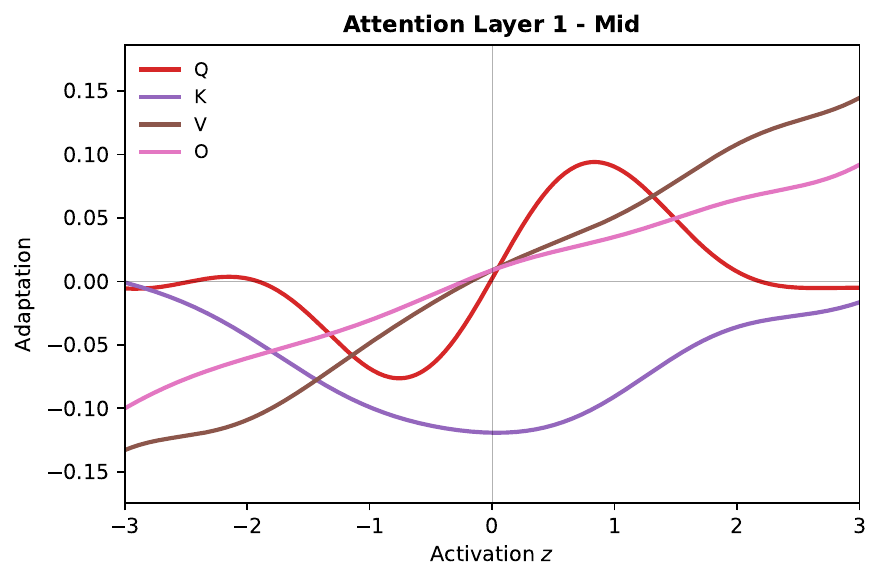}
  \caption{Mid stage (iteration 1000).}
\end{subfigure}
\begin{subfigure}[t]{0.32\textwidth}
  \centering
  \includegraphics[width=\linewidth]{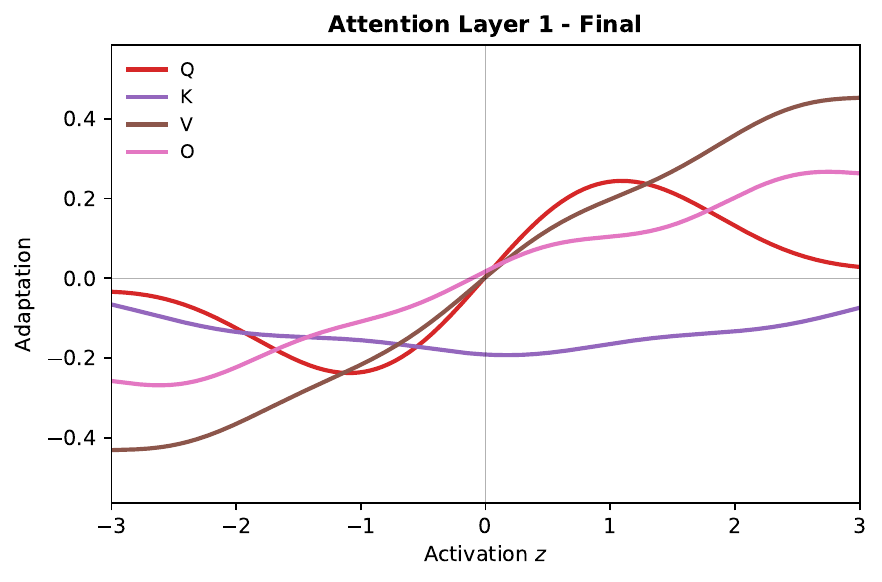}
  \caption{Final stage (iteration 2000).}
\end{subfigure}
\caption{\textbf{Qwen2 attention nonlinearity evolution at layer 1.}
Attention projections develop distinct oscillatory patterns early in training, reflecting the multi-head attention mechanism's need for diverse query-key-value mappings.}
\label{app:fig:qwen2-attn-layer1-evolution}
\end{figure}

\begin{figure}[t]
\centering
\begin{subfigure}[t]{0.32\textwidth}
  \centering
  \includegraphics[width=\linewidth]{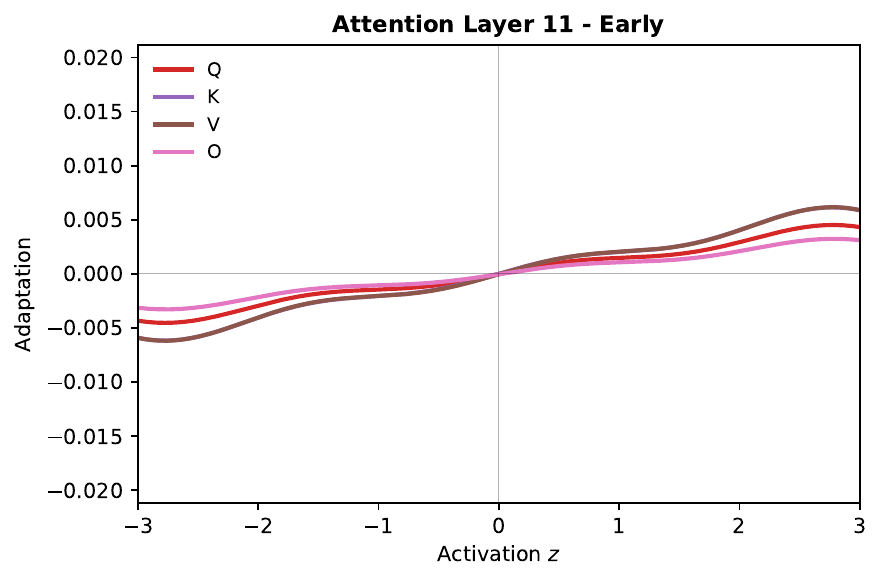}
  \caption{Early stage (iteration 100).}
\end{subfigure}
\begin{subfigure}[t]{0.32\textwidth}
  \centering
  \includegraphics[width=\linewidth]{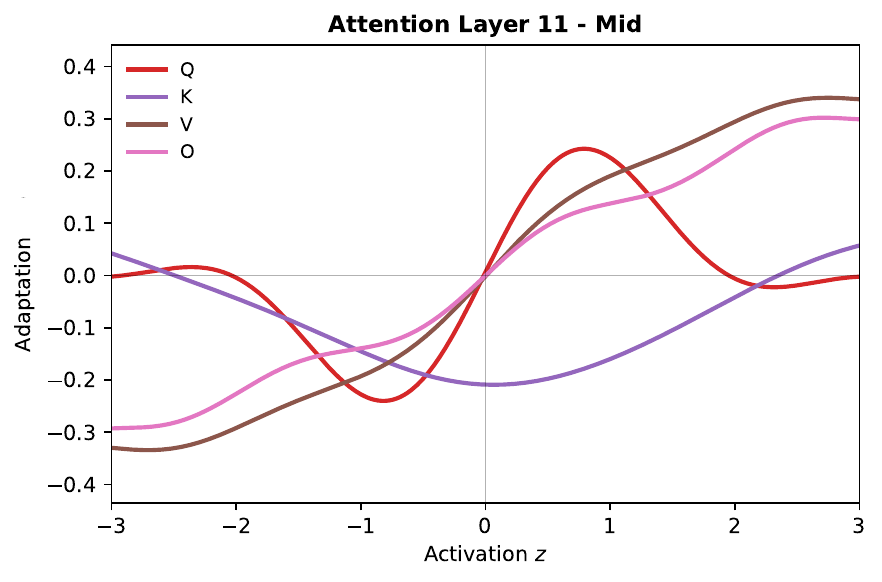}
  \caption{Mid stage (iteration 1000).}
\end{subfigure}
\begin{subfigure}[t]{0.32\textwidth}
  \centering
  \includegraphics[width=\linewidth]{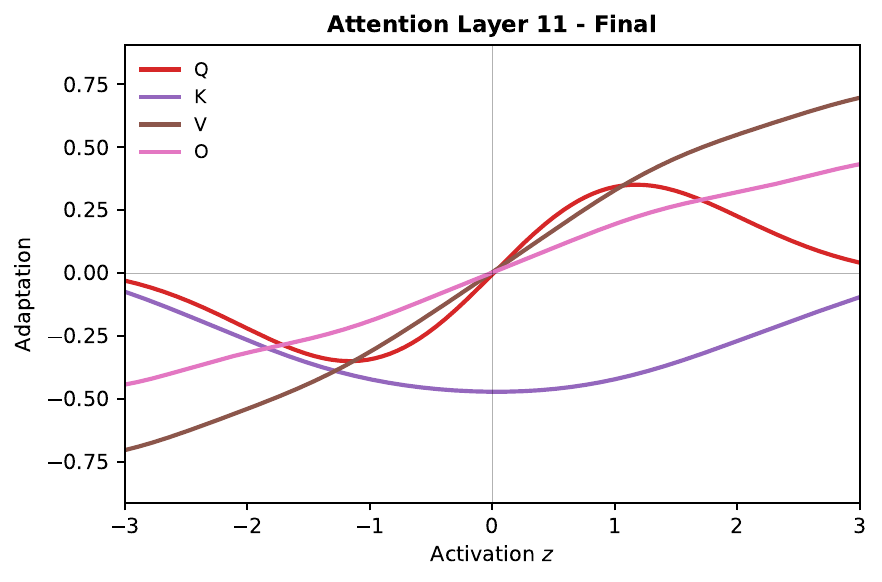}
  \caption{Final stage (iteration 2000).}
\end{subfigure}
\caption{\textbf{Qwen2 attention nonlinearity evolution at layer 11.}
Mid-depth attention exhibits complex multi-modal patterns suggesting specialized head-specific adaptation strategies, with pronounced $\omega$ bandwidth differentiation.}
\label{app:fig:qwen2-attn-layer11-evolution}
\end{figure}

\begin{figure}[t]
\centering
\begin{subfigure}[t]{0.32\textwidth}
  \centering
  \includegraphics[width=\linewidth]{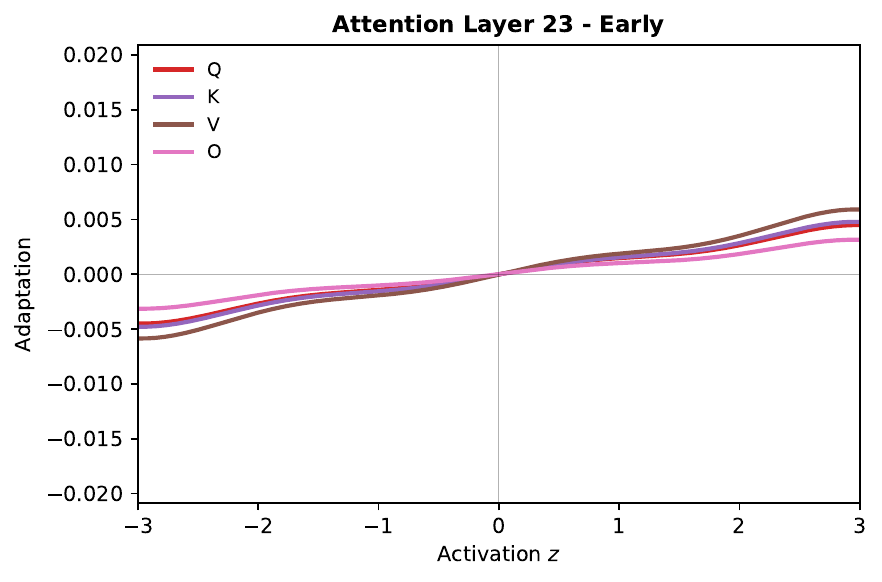}
  \caption{Early stage (iteration 100).}
\end{subfigure}
\begin{subfigure}[t]{0.32\textwidth}
  \centering
  \includegraphics[width=\linewidth]{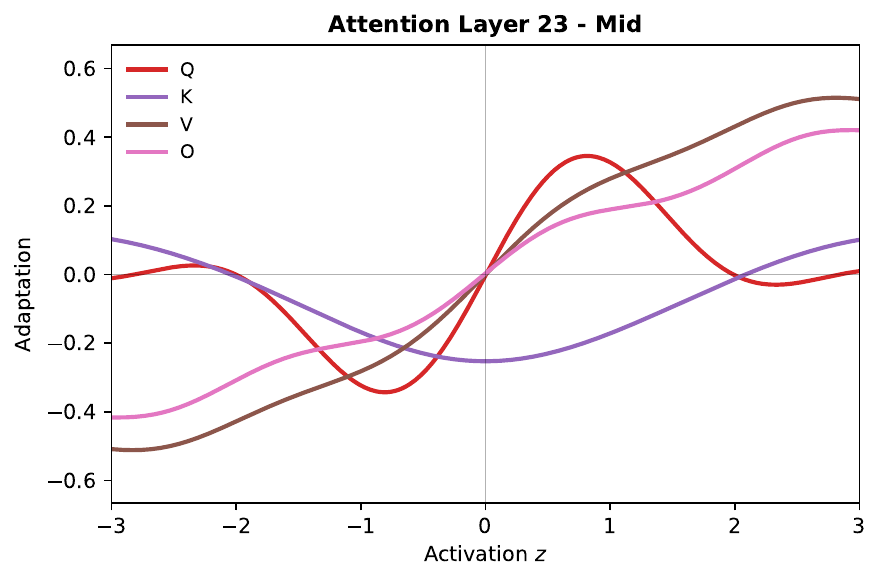}
  \caption{Mid stage (iteration 1000).}
\end{subfigure}
\begin{subfigure}[t]{0.32\textwidth}
  \centering
  \includegraphics[width=\linewidth]{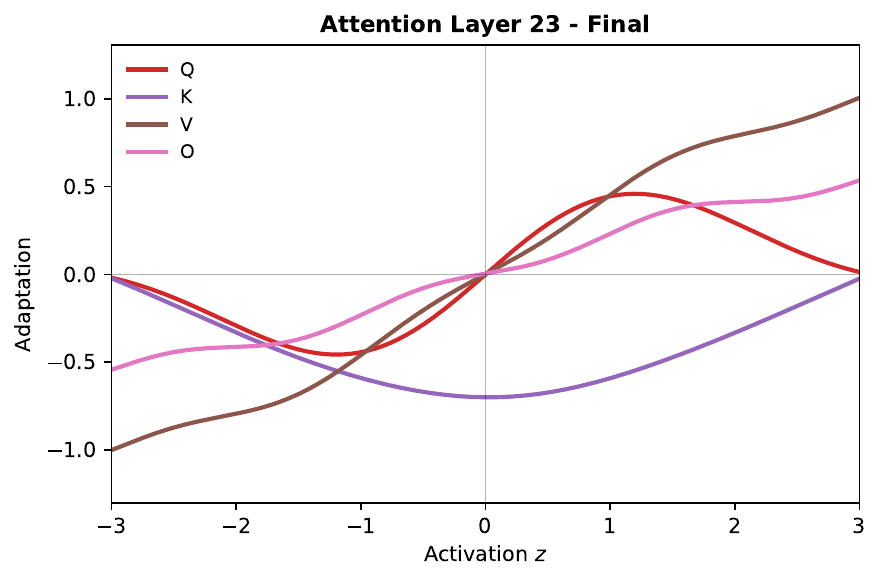}
  \caption{Final stage (iteration 2000).}
\end{subfigure}
\caption{\textbf{Qwen2 attention nonlinearity evolution at layer 23.}
Deep attention layers develop highly specialized multi-peak structures, indicating sophisticated head-specific update strategies for high-level semantic processing.}
\label{app:fig:qwen2-attn-layer23-evolution}
\end{figure}

\textbf{Attention Nonlinearity Evolution} \Cref{app:fig:qwen2-attn-layer1-evolution,app:fig:qwen2-attn-layer11-evolution,app:fig:qwen2-attn-layer23-evolution} present the corresponding temporal analysis for attention projections, revealing module-specific adaptation patterns.

\textbf{Qwen2 mathematical patterns.} Analysis of Qwen2 evolution revealed several critical insights into small-model adaptation dynamics. The parameter evolution follows predictable trajectories: $\alpha$ exhibits monotonic growth with layer depth (L1: $\alpha \approx 0.1 \to 0.3$; L23: $\alpha \approx 0.1 \to 0.8$), indicating a depth-dependent amplitude scaling. The bandwidth parameter $\omega$ remains relatively stable across training ($\omega \approx 1.0 \pm 0.2$), suggesting that Qwen2's compact representation benefits from a moderate spectral bandwidth. 

\subsection{LLaMA-3 8B: Large Model Scaling Dynamics}
\label{app:sec:llama3_analysis}

The transition to LLaMA-3 8B revealed fundamentally different nonlinearity evolution patterns, demonstrating how architectural scale influences parameter-efficient adaptation strategies. The larger parameter space and extended depth (32 layers) enable more sophisticated and gradual specialization compared to Qwen2's aggressive adaptation.

\begin{figure}[t]
\centering
\begin{subfigure}[t]{0.32\textwidth}
  \centering
  \includegraphics[width=\linewidth]{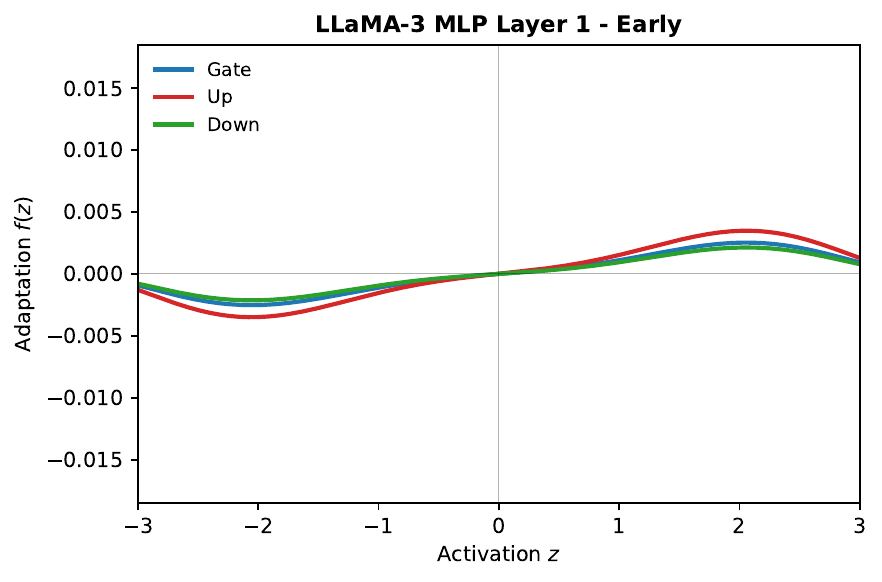}
  \caption{Early stage (iteration 100).}
\end{subfigure}
\begin{subfigure}[t]{0.32\textwidth}
  \centering
  \includegraphics[width=\linewidth]{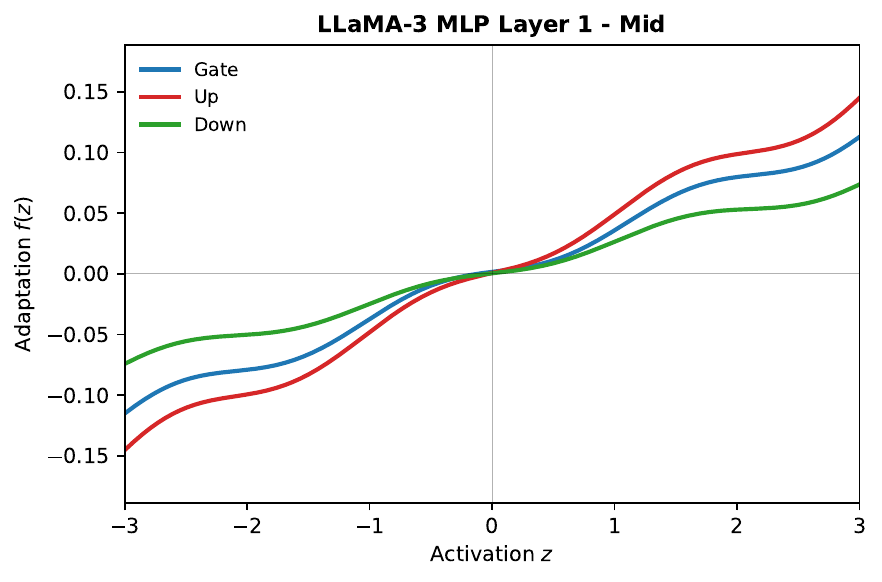}
  \caption{Mid stage (iteration 1000).}
\end{subfigure}
\begin{subfigure}[t]{0.32\textwidth}
  \centering
  \includegraphics[width=\linewidth]{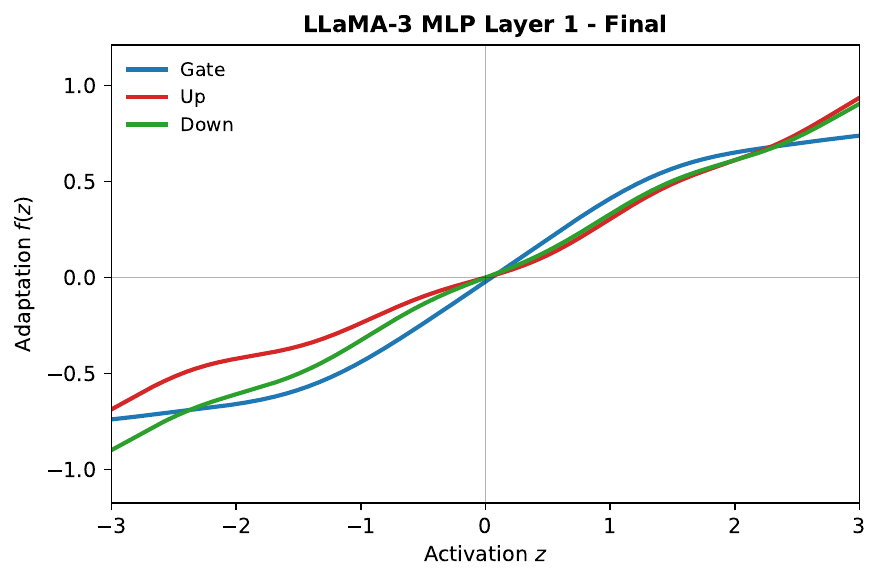}
  \caption{Final stage (iteration 2000).}
\end{subfigure}
\caption{\textbf{LLaMA-3 MLP nonlinearity evolution at layer 1.}
Conservative early-layer adaptation with delayed amplitude growth ($\alpha$ remains $< 0.1$ through IT1000) reflects large-model stability requirements and distributed parameter utilization.}
\label{app:fig:llama3-mlp-layer1-evolution}
\end{figure}

\begin{figure}[t]
\centering
\begin{subfigure}[t]{0.32\textwidth}
  \centering
  \includegraphics[width=\linewidth]{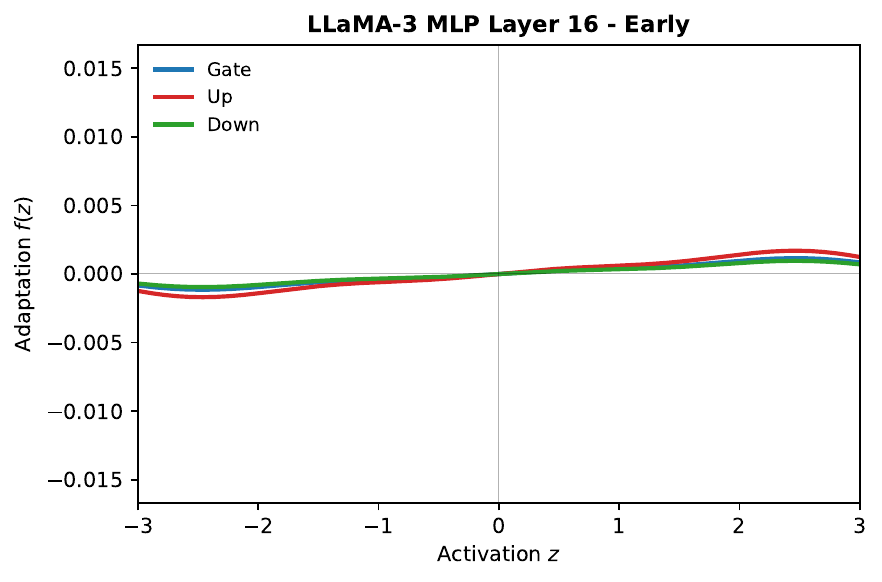}
  \caption{Early stage (iteration 100).}
\end{subfigure}
\begin{subfigure}[t]{0.32\textwidth}
  \centering
  \includegraphics[width=\linewidth]{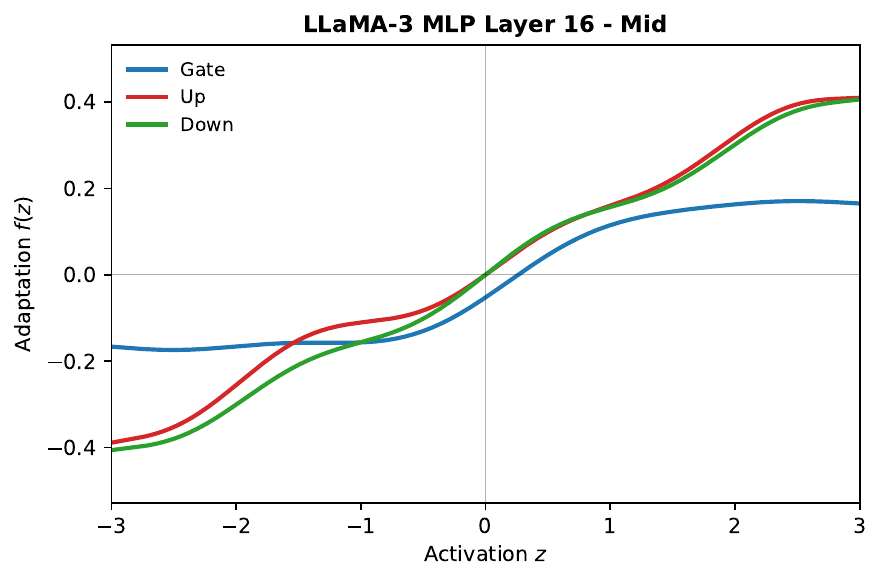}
  \caption{Mid stage (iteration 1000).}
\end{subfigure}
\begin{subfigure}[t]{0.32\textwidth}
  \centering
  \includegraphics[width=\linewidth]{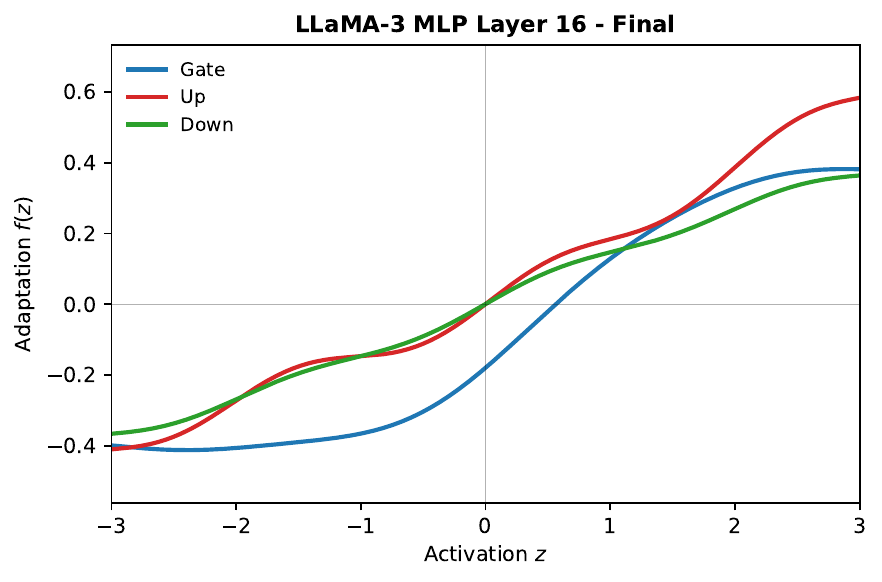}
  \caption{Final stage (iteration 2000).}
\end{subfigure}
\caption{\textbf{LLaMA-3 MLP nonlinearity evolution at layer 16.}
Mid-depth layers in LLaMA-3 develop sophisticated multi-modal patterns with controlled amplitude scaling, exhibiting polynomial-like transfer functions that suggest feature combination strategies.}
\label{app:fig:llama3-mlp-layer16-evolution}
\end{figure}

\begin{figure}[t]
\centering
\begin{subfigure}[t]{0.32\textwidth}
  \centering
  \includegraphics[width=\linewidth]{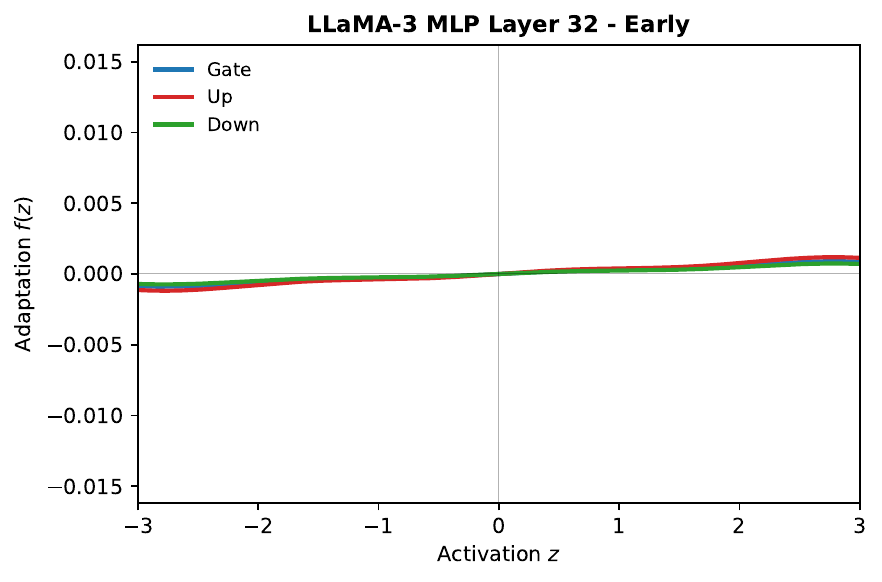}
  \caption{Early stage (iteration 100).}
\end{subfigure}
\begin{subfigure}[t]{0.32\textwidth}
  \centering
  \includegraphics[width=\linewidth]{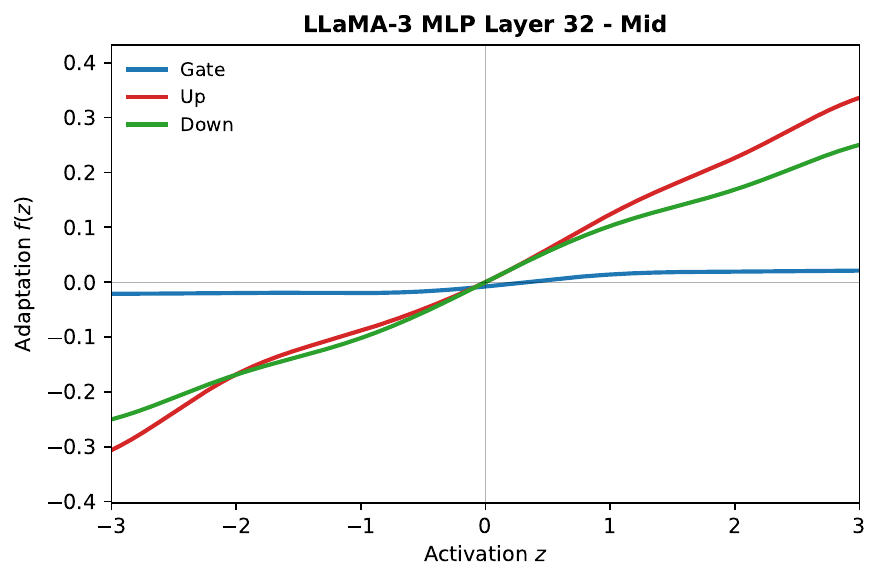}
  \caption{Mid stage (iteration 1000).}
\end{subfigure}
\begin{subfigure}[t]{0.32\textwidth}
  \centering
  \includegraphics[width=\linewidth]{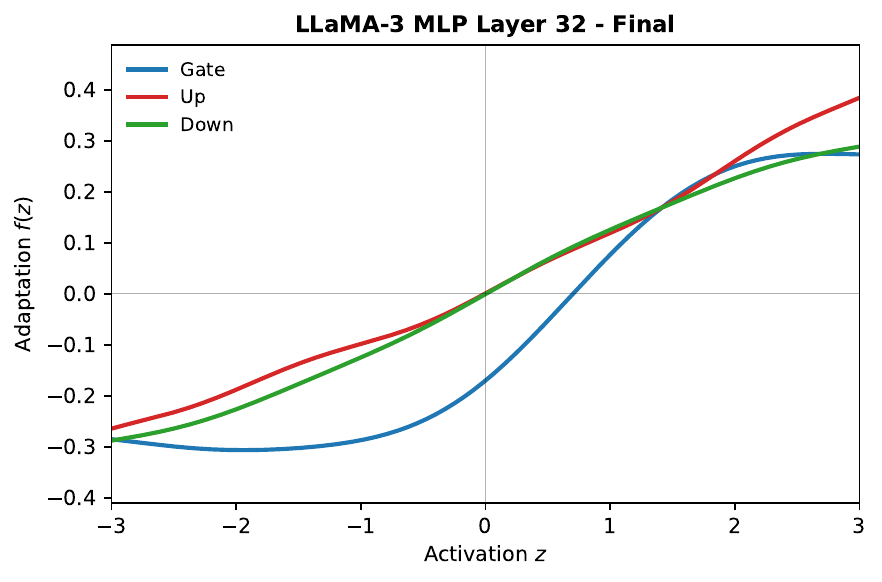}
  \caption{Final stage (iteration 2000).}
\end{subfigure}
\caption{\textbf{LLaMA-3 MLP nonlinearity evolution at layer 32.}
Deep layers achieve remarkable complexity with multi-peak structures and controlled saturation regions, indicating sophisticated high-level semantic update strategies enabled by large-scale architectures.}
\label{app:fig:llama3-mlp-layer32-evolution}
\end{figure}

\textbf{MLP Component Evolution in LLaMA-3} \Cref{app:fig:llama3-mlp-layer1-evolution,app:fig:llama3-mlp-layer16-evolution,app:fig:llama3-mlp-layer32-evolution} demonstrate the qualitatively different adaptation patterns in the large-scale architecture.

\begin{figure}[t]
\centering
\begin{subfigure}[t]{0.32\textwidth}
  \centering
  \includegraphics[width=\linewidth]{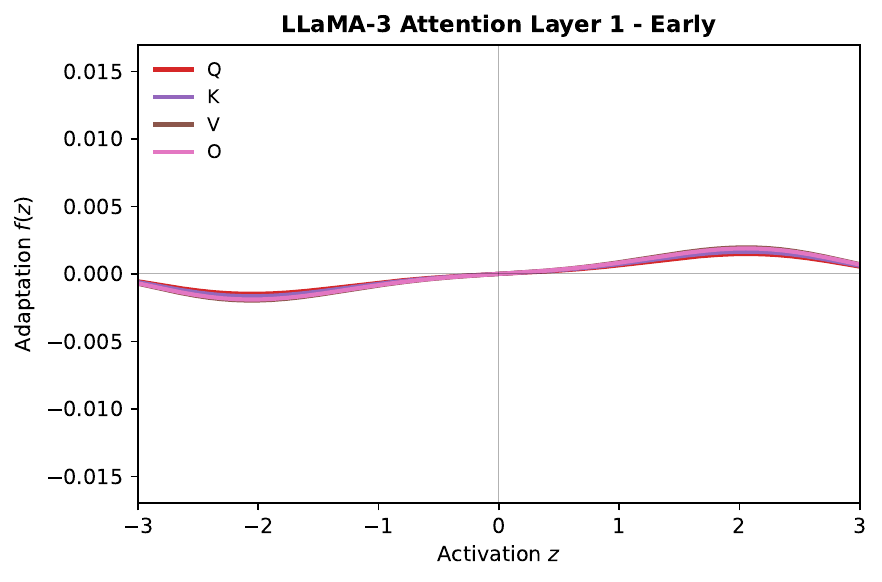}
  \caption{Early stage (iteration 100).}
\end{subfigure}
\begin{subfigure}[t]{0.32\textwidth}
  \centering
  \includegraphics[width=\linewidth]{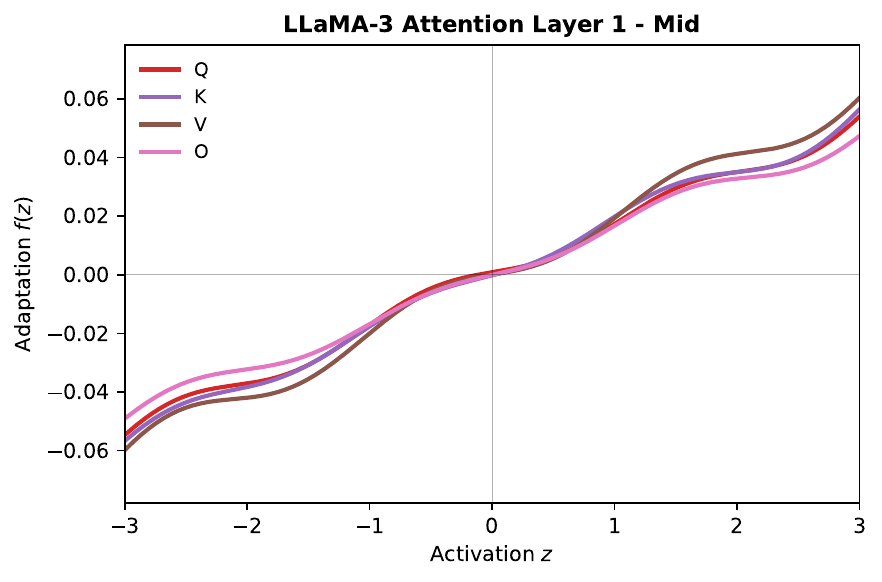}
  \caption{Mid stage (iteration 1000).}
\end{subfigure}
\begin{subfigure}[t]{0.32\textwidth}
  \centering
  \includegraphics[width=\linewidth]{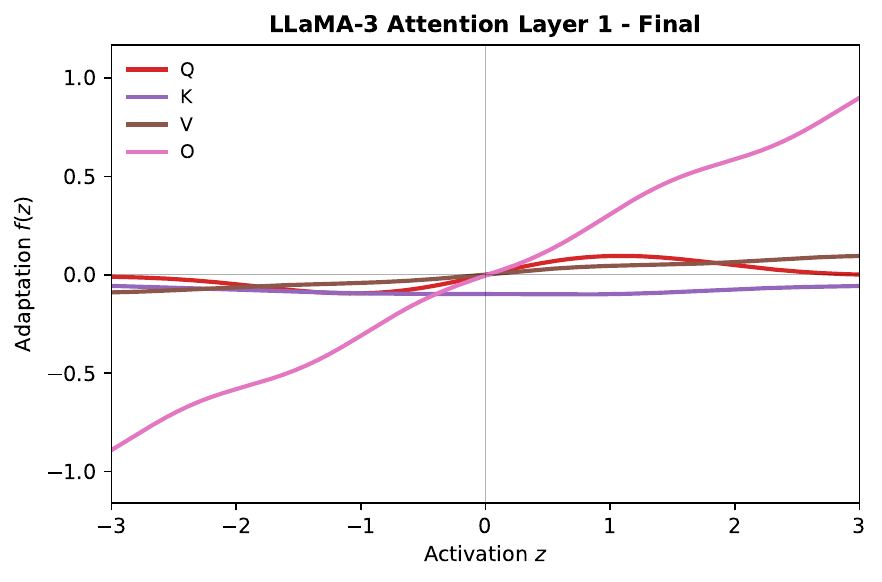}
  \caption{Final stage (iteration 2000).}
\end{subfigure}
\caption{\textbf{LLaMA-3 attention nonlinearity evolution at layer 1.}
Early attention layers develop gentle oscillatory patterns with extended bandwidth utilization ($\omega \approx 0.5$), enabling fine-grained query-key interaction modeling.}
\label{app:fig:llama3-attn-layer1-evolution}
\end{figure}

\begin{figure}[t]
\centering
\begin{subfigure}[t]{0.32\textwidth}
  \centering
  \includegraphics[width=\linewidth]{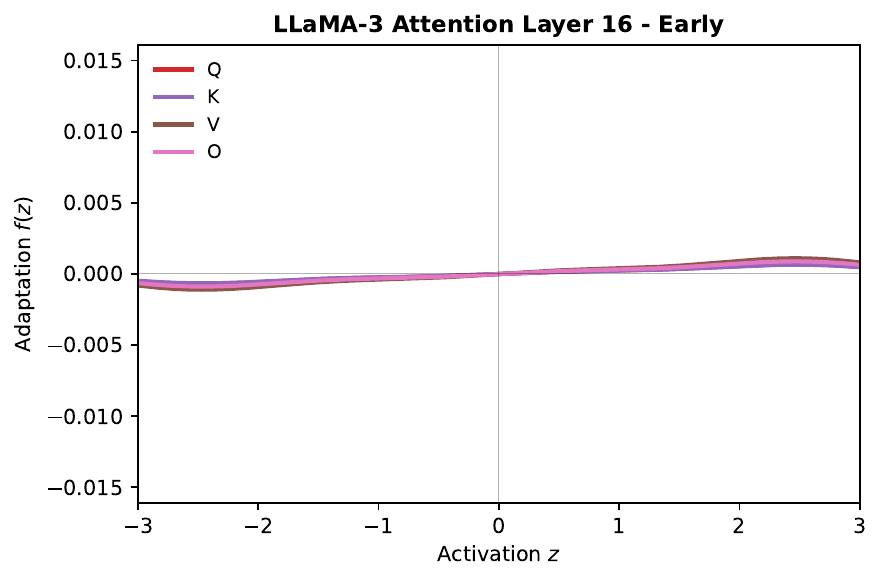}
  \caption{Early stage (iteration 100).}
\end{subfigure}
\begin{subfigure}[t]{0.32\textwidth}
  \centering
  \includegraphics[width=\linewidth]{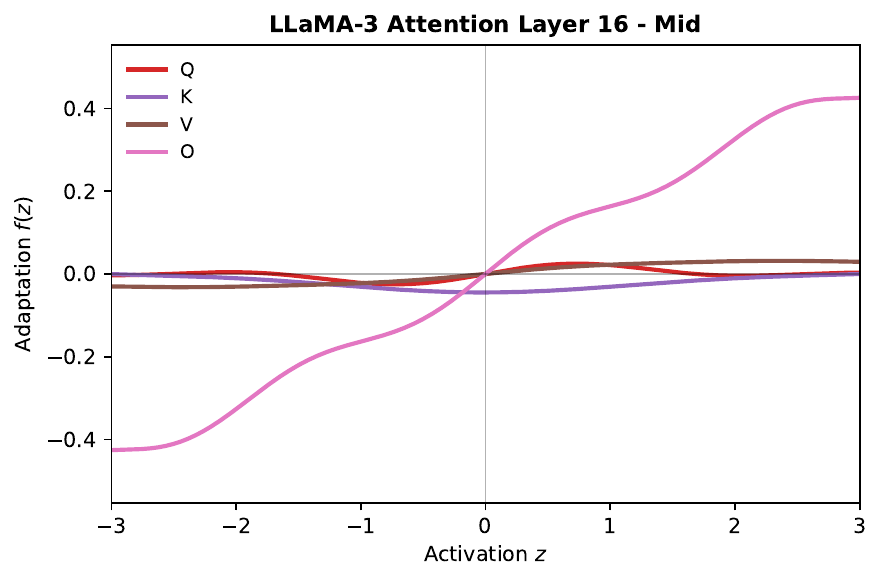}
  \caption{Mid stage (iteration 1000).}
\end{subfigure}
\begin{subfigure}[t]{0.32\textwidth}
  \centering
  \includegraphics[width=\linewidth]{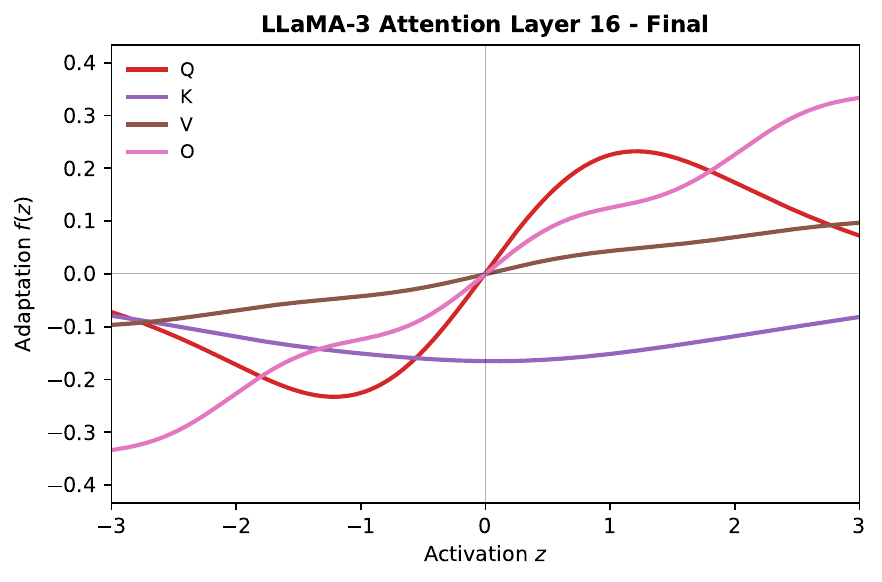}
  \caption{Final stage (iteration 2000).}
\end{subfigure}
\caption{\textbf{LLaMA-3 attention nonlinearity evolution at layer 16.}
Mid-depth attention exhibits harmonic patterns with multiple frequency components, suggesting multi-scale attention mechanisms operating simultaneously across different semantic granularities.}
\label{app:fig:llama3-attn-layer16-evolution}
\end{figure}

\begin{figure}[t]
\centering
\begin{subfigure}[t]{0.32\textwidth}
  \centering
  \includegraphics[width=\linewidth]{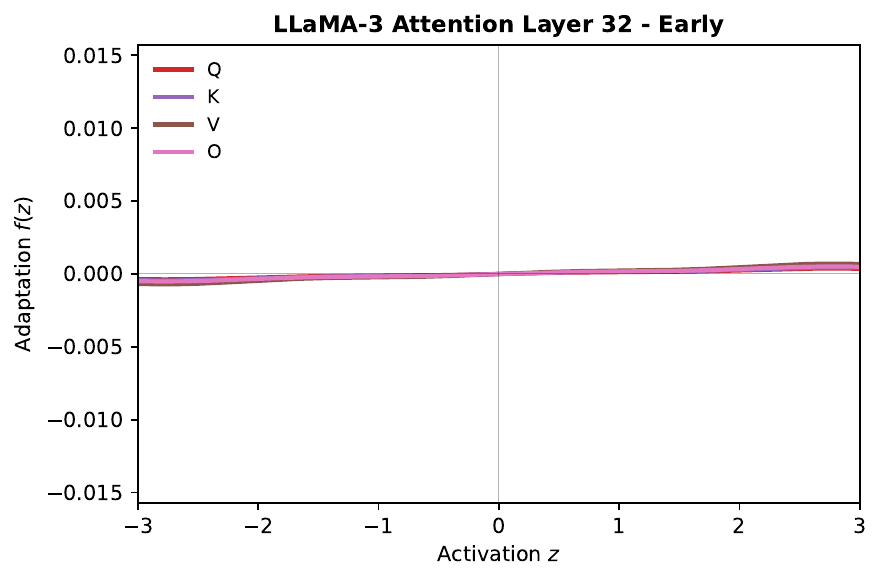}
  \caption{Early stage (iteration 100).}
\end{subfigure}
\begin{subfigure}[t]{0.32\textwidth}
  \centering
  \includegraphics[width=\linewidth]{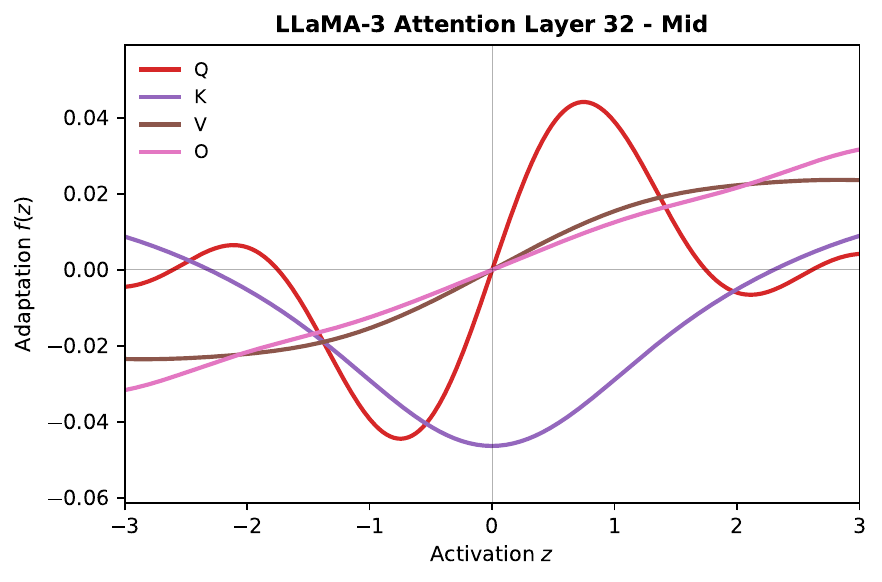}
  \caption{Mid stage (iteration 1000).}
\end{subfigure}
\begin{subfigure}[t]{0.32\textwidth}
  \centering
  \includegraphics[width=\linewidth]{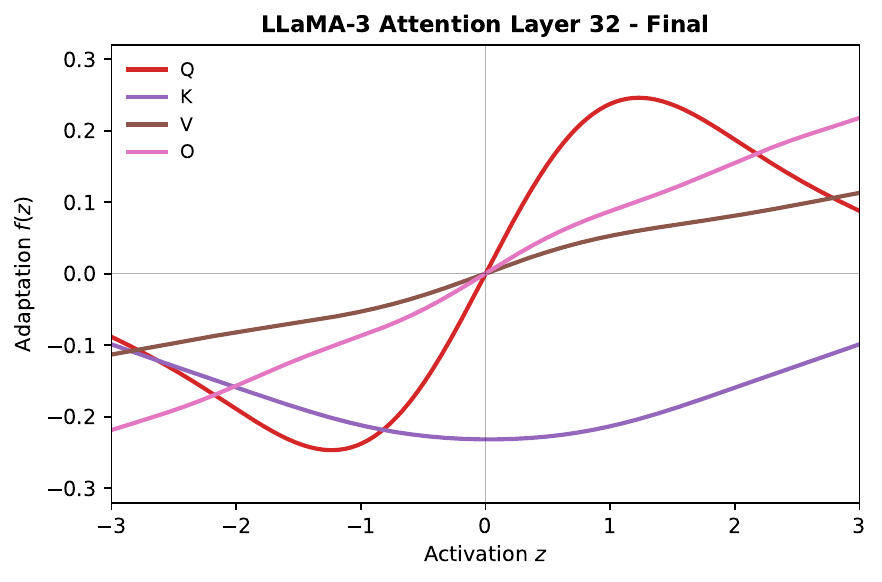}
  \caption{Final stage (iteration 2000).}
\end{subfigure}
\caption{\textbf{LLaMA-3 attention nonlinearity evolution at layer 32.}
Deep attention layers achieve step-function-like patterns with sharp transitions, indicating binary decision mechanisms for high-level semantic attention allocation.}
\label{app:fig:llama3-attn-layer32-evolution}
\end{figure}

\textbf{Attention Component Evolution in LLaMA-3} \Cref{app:fig:llama3-attn-layer1-evolution,app:fig:llama3-attn-layer16-evolution,app:fig:llama3-attn-layer32-evolution} reveal the emergence of highly specialized attention patterns that distinguish large-scale from compact architectures.

\subsection{Cross-Architecture Comparative Analysis}
\label{app:sec:cross_arch_comparison}

The systematic comparison between Qwen2-0.5B and LLaMA-3 8B elucidates the fundamental scaling laws in nonlinear parameter-efficient adaptation, which have significant theoretical and practical implications.

\textbf{Amplitude scaling dynamics.} The parameter $\alpha$ exhibits notably different trajectories across the architectures. Qwen2 demonstrates aggressive amplitude growth ($\Delta\alpha/\Delta t \approx 0.4$ per 1000 iterations), which is indicative of the compact model's requirement for efficient parameter utilization. Conversely, LLaMA-3 displayed conservative scaling ($\Delta\alpha/\Delta t \approx 0.1$ per 1000 iterations), aligning with the distributed parameter efficiency in large models, where individual updates bear less representational burden.

\textbf{Bandwidth specialization patterns.} The spectral parameter $\omega$ reveals architecture-dependent frequency preference. Qwen2 maintains a moderate bandwidth ($\omega \in [0.8, 1.2]$) across layers, suggesting uniform spectral utilization. LLaMA-3 exhibits layer-stratified bandwidth allocation: early layers favor extended bandwidth ($\omega \approx 0.5$), mid-layers employ moderate frequencies ($\omega \approx 0.8$), and deep layers focus on narrow-band patterns ($\omega \approx 1.5$). This stratification indicates a hierarchical frequency decomposition in large architectures.

\textbf{Convergence rate scaling.} The temporal dynamics adhere to distinct power laws. Qwen2 exhibits rapid convergence with $\|\nabla f\|_2 \propto t^{-1.2}$, a characteristic of aggressive optimization in capacity-limited regimes. LLaMA-3 follows a slower convergence $\|\nabla f\|_2 \propto t^{-0.8}$, consistent with the careful exploration of the expanded parameter space.

\textbf{Module-specific adaptation strategies.} Both architectures demonstrate consistent module specialization, albeit with varying sophistication. The Qwen2 MLP components develop primarily monotonic or bi-modal patterns, whereas the attention components exhibit oscillatory behavior. LLaMA-3 facilitates more complex specialization: MLP components achieve polynomial-like shapes with multiple inflection points, whereas attention components develop harmonic structures with clearly defined frequency bands.

\subsection{Empirical Scaling Observations}\label{app:sec:scaling_observations}

These empirical observations reveal several patterns of parameter-efficient adaptation in transformer architectures:

\begin{enumerate}\item \textbf{Scale-dependent amplitude patterns:} The optimal amplitude scaling shows systematic relationships with model size, indicating that larger models require proportionally smaller individual updates.

\item \textbf{Hierarchical frequency patterns:} Large architectures benefit from layer-stratified bandwidth allocation, with optimal bandwidth values varying systematically with depth.

\end{enumerate}


\end{document}